\documentclass[10pt,twocolumn,letterpaper]{article}

\usepackage[pagenumbers]{cvpr} 

\usepackage{algorithm}
\usepackage{algorithmic}
\usepackage{array}
\usepackage{ragged2e}

\usepackage{multirow}
\usepackage{colortbl}
\usepackage{booktabs}
\usepackage{xcolor}
\usepackage{makecell}
\usepackage{bibentry}
\usepackage{bm}
\usepackage{amsmath, amssymb, mathtools}

\usepackage{titletoc}
\usepackage{cuted} 

\usepackage[accsupp]{axessibility}  

\definecolor{cvprblue}{rgb}{0.21,0.49,0.74}
\definecolor{Gain}{RGB}{210,240,210}
\definecolor{Loss}{RGB}{255,220,220}
\definecolor{Neutral}{RGB}{240,240,240}
\definecolor{HeaderGray}{gray}{0.93}
\definecolor{CleanGreen}{RGB}{215,235,215}
\definecolor{DeepRed}{RGB}{144,30,30}
\definecolor{DefenseOrange}{RGB}{200,90,0}
\usepackage[pagebackref,breaklinks,colorlinks,allcolors=cvprblue]{hyperref}

\def\paperID{41711} 
\def\confName{CVPR}
\def\confYear{2026}

\title{\textit{Out of Sight, Out of Track}: Adversarial Attacks on Propagation-based Multi-Object Trackers via Query State Manipulation}

\author{Halima Bouzidi\textsuperscript{*} \;Haoyu Liu \;Yonatan Achamyeleh \;Praneetsai Iddamsetty \;Mohammad Al Faruque \\
University of California, Irvine, CA, USA \\
{\tt\small \{hbouzidi, hliu32, yachamye, iddamsep, alfaruqu\}@uci.edu}
}

\begin{document}
\maketitle
\renewcommand{\thefootnote}{\fnsymbol{footnote}}
\footnotetext[1]{Corresponding Author.}

\begin{abstract}
Recent Tracking-by-Query-Propagation (TBP) methods have advanced Multi-Object Tracking (MOT) by enabling end-to-end (E2E) pipelines with long-range temporal modeling. However, this reliance on query propagation introduces unexplored architectural vulnerabilities to adversarial attacks. We present FADE, a novel attack framework designed to exploit these specific vulnerabilities. FADE employs two attack strategies targeting core TBP mechanisms: (i) Temporal Query Flooding: Generates spurious temporally consistent track queries to exhaust the tracker's limited query budget, forcing it to terminate valid tracks. (ii) Temporal Memory Corruption: Directly attacks the query updater's memory by severing temporal links via state de-correlation and erasing the learned feature identity of matched tracks. Furthermore, we introduce a differentiable pipeline to optimize these attacks for physical-world realizability by leveraging simulations of advanced perception sensor spoofing. Experiments on MOT17 and MOT20 benchmarks demonstrate that FADE is highly effective against state-of-the-art TBP trackers, causing significant identity switches and track terminations.
\end{abstract}    
\section{Introduction}

Multi-Object Tracking (MOT) is a fundamental capability for systems that perceive and interact with dynamic environments~\cite{bewley2016simple} and a core component in applications like trajectory planning and situational awareness~\cite{ravindran2020multi}. The traditional \textit{Tracking-by-Detection} (TBD) paradigm works by first detecting objects and then using a separate association algorithm, such as Kalman filters~\cite{braso2020learning} or appearance matching~\cite{bertinetto2016fully}, to keep track of them. However, these disjoint pipeline stages often struggle in complex scenarios with heavy occlusions or high object similarity~\cite{song2024sftrack, wang2024smiletrack, zhang2022bytetrack}.
\begin{figure}[ht]
\centering
\includegraphics[width=0.4\textwidth]{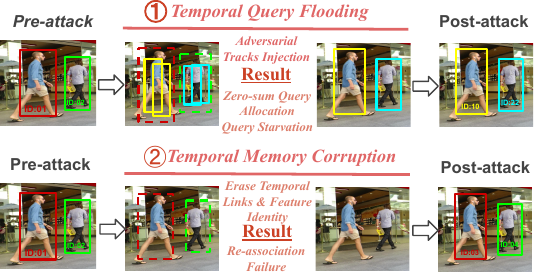}
\vspace{-0.2cm}
\caption{Overview of the Proposed FADE Attack Strategies.}
\label{fig:teaser}
\vspace{-0.7cm}
\end{figure}

Addressing these issues, the \textit{Tracking-by-Propagation} (TBP) paradigm emerges, framing MOT as an End-to-End (E2E) autoregressive task~\cite{sun2020transtrack,zeng2022motr,zhang2023motrv2}. TBP methods propagate \textit{track queries}, which are persistent representations of objects' motion and appearance, over time. 
TBP trackers evolve around three mechanisms: (\textit{i}) query propagation, (\textit{ii}) temporal memory representation, and (\textit{iii}) training dynamics. MOTR~\cite{zeng2022motr} eliminates the need for association heuristics~\cite{sahbani2016kalman}. MOTRv2~\cite{zhang2023motrv2} improves the detection capability with external detectors~\cite{ge2021yolox}. MeMOTR~\cite{gao2023memotr} enhances temporal modeling with a built-in long-term memory bank. Samba~\cite{segu2024samba} advances memory modeling with State-Space Models (SSM). The recent CO-MOT~\cite{yan2023bridging} introduces a new label assignment strategy to balance gradient contributions between persistent track queries and new detection queries.

However, MOT models are still vulnerable to adversarial attacks, which can degrade their tracking reliability. Adversarial attacks against TBD trackers exploit vulnerabilities across three axes categorized as: \textit{(i) Detection evasion attacks} that minimize detection confidence (False Negative~\cite{lu2017adversarial}) or maximize false detections (Daedalus~\cite{wang2021daedalus}). \textit{(ii) Association disruption attacks}, such as Hijack~\cite{jia2020fooling} and F\&F~\cite{zhou2023f}, target temporal predictions by maximizing the Mahalanobis distance between perturbed detections and Kalman filter predictions, causing gating rejections and identity switches. \textit{(iii) Feature manipulation attacks}, as introduced in BankTweak~\cite{shin2024banktweak}, directly perturb appearance embeddings to manipulate cosine similarity in the feature extractor, disrupting the object-to-track matching process.

Prior adversarial attacks on MOT are mainly designed for TBD architectures and can not effectively exploit vulnerabilities in emerging TBP trackers. Detection evasion attacks~\cite{lu2017adversarial,wang2021daedalus} targeting hard NMS thresholds are ineffective against learned suppression in E2E trackers, where duplicate handling occurs through differentiable attention rather than fixed Intersection-over-Union (IoU) rules. Association disruption attacks~\cite{jia2020fooling,zhou2023f} that manipulate Kalman filter motion predictions become inapplicable when temporal dynamics are learned through query propagation. Feature manipulation attacks~\cite{shin2024banktweak} designed for separate appearance banks can not transfer to integrated architectures where detection and re-identification share unified query representations.
They are also impractical as they assume an attacker who can access the inference pipeline and directly inject adversarial noise into the extracted association features.

TBP trackers introduce novel structural vulnerabilities. First, the \textit{fixed track query budget} creates a zero-sum resource allocation problem: allocating a query to a spurious track reduces capacity for legitimate ones, and when the combined count of real and adversarial objects approaches this budget, real tracks can face query starvation. Second, the TBP temporal architecture creates a critical duality. On one hand, by aggregating multi-frame evidence, it implements temporal smoothing that attenuates isolated, single-frame perturbations. On the other hand, this mechanism creates temporal dependency chains. Query-based trackers propagate learned hidden states recurrently, where each query's current representation depends on its previous state, allowing state corruption to propagate forward across frames. Specifically, this temporal coupling is amplified by the tracker's core temporal memory mechanism, whether it is a \textit{propagated hidden state}~\cite{zeng2022motr, zhang2023motrv2, yan2023bridging}, a \textit{built-in long-term memory bank}~\cite{gao2023memotr}, or a \textit{state-space model}~\cite{segu2024samba}, since this component stores historical embeddings that inform matching, allowing corrupted memories to persist and contaminate future associations. In contrast, TBD trackers compute associations independently per frame, providing implicit reset mechanisms that contain perturbations.

Translating digital attacks into the physical world for MOT faces many challenges, as conventional patch-based attacks are often designed for single-object tracking (SOT) and fail to impact multiple targets simultaneously. 
Recently, remote camera sensor spoofing attacks have emerged as a powerful and stealthy physical vector by manipulating the camera sensor's physical components using: (\textit{i}) \textit{Acoustic signals}~\cite{ji2021poltergeist,zhu2023tpatch}, which vibrate a camera's stabilization sensors to induce motion blur, and (\textit{ii}) \textit{Electromagnetic interference}~\cite{liao2025your,liu2025magshadow}, which directly targets the sensor's electronics to disrupt analog-to-digital conversion and manipulate pixel values. These precise sensor-level manipulations can be leveraged to erase or alter object trajectories, directly undermining MOT trackers' reliance on both accurate detections and temporal consistency. However, prior work has only evaluated these vectors against single-frame detection tasks, leaving unexplored their implications for temporal tracking and identity association in MOT.

\noindent \textbf{Scientific Contributions.} We present the first comprehensive adversarial vulnerability analysis of the emerging TBP trackers. \textit{First}, we identify structural constraints unique to query-based architectures: (\textit{i}) the fixed query budget creates zero-sum resource allocation vulnerabilities, (\textit{ii}) recurrent hidden state propagation enables temporal error chains, and (\textit{iii}) built-in temporal memory mechanisms (long-term memory or SSMs) amplify attack persistence. 
\textit{Second}, we propose two novel attack strategies as depicted in Fig.\ref{fig:teaser}:
\begin{itemize}
    \item (1) \textit{Temporal Query Flooding}: Generates spurious, persistent tracks to exhaust the zero-sum query budget while simultaneously siphoning legitimate track identities via cost mimicry to ensure their persistence.
    \item (2) \textit{Temporal Memory Corruption}: Destroys legitimate tracks by severing temporal links via state de-correlation and erasing the feature identities of matched queries.
\end{itemize}
\textit{Third}, we propose an E2E differentiable pipeline for crafting both digital and simulated physical adversarial examples (AEs), leveraging differentiable simulations of perception sensor spoofing (e.g., acoustic~\cite{zhu2023tpatch} and electromagnetic~\cite{ren2025ghostshot,liu2025magshadow,liao2025your}) as attack vectors.
\textit{Finally}, we evaluate FADE on a diverse set of state-of-the-art TBP-based MOT trackers (MOTR~\cite{zeng2022motr}, MOTRv2~\cite{zhang2023motrv2}, MeMOTR~\cite{gao2023memotr}, CO-MOT~\cite{yan2023bridging}, and Samba~\cite{segu2024samba}) across challenging benchmarks (MOT17, MOT20). FADE achieves up to $\approx$30-point~HOTA drop and over \textbf{$10\times$} identity switches (IDSW). 

\section{Related Work}

\subsection{Multi-object Tracking (MOT).}
Multi-object tracking (MOT) aims to detect objects and assign consistent identities over time~\cite{bewley2016simple}. The dominant \textit{tracking-by-detection} (TBD) paradigm first detects objects, then links them using separate association modules, such as Kalman filters~\cite{welch1995introduction} or feature re-identification~\cite{braso2020learning, zhang2022bytetrack}. To overcome TBD's disjointed pipeline, the \textit{tracking-by-propagation} (TBP) paradigm emerged
\cite{meinhardt2022trackformer, zeng2022motr}. TBP trackers propagate learnable \textit{track queries} across frames, jointly learning detection and temporal association. Recent works have focused on evolving this architecture by integrating external detectors~\cite{zhang2023motrv2}, refining training dynamics~\cite{yan2023bridging}, and enhancing temporal robustness with memory banks~\cite{gao2023memotr} or State-Space Models (SSMs)~\cite{segu2024samba}. A key architectural shift from TBD's unbounded tracker instantiation is that TBP trackers rely on a fixed query budget and manage recurrent state propagation through persistent memory. These novel mechanisms, central to TBP's success, also introduce unique, unexplored adversarial attack surfaces.

\subsection{Adversarial Attacks on MOT}
Adversarial attacks on MOT~\cite{biggio2018wild, goodfellow2014explaining} have almost exclusively focused on the TBD paradigm. These attacks target three main stages: detection evasion~\cite{lu2017adversarial, wang2021daedalus}, association disruption by manipulating Kalman filter predictions~\cite{jia2020fooling, zhou2023f}, or feature manipulation of appearance embeddings~\cite{shin2024banktweak}. However, these methods are architecturally incompatible with TBP trackers, as their logic is coupled to TBD-specific components (e.g., Kalman filters, separate Re-ID banks) that do not exist in E2E query propagation models. Consequently, the unique vulnerabilities of TBP models (i.e. query budget, recurrent memory) remain unstudied. To our knowledge, FADE is the first to uncover TBP vulnerabilities by attacking core components within the TBP E2E architecture.

\subsection{Physical Adversarial Examples for MOT}
Patch-based adversarial examples, while studied for SOT~\cite{ding2021towards, long2024papmot}, are not suited for MOT. Their static nature, SOT-focused, and visual obtrusiveness make them ineffective at disrupting the complex, multi-object temporal dynamics~\cite{long2024papmot}. 
Physical sensor manipulation through acoustic signals or electromagnetic interference~\cite{wang2022survey} overcomes patch limitations by affecting all objects simultaneously, regardless of count, scale, or occlusion. Acoustic Adversarial Injection (AAI), as demonstrated in Poltergeist~\cite{ji2021poltergeist} and TPatch~\cite{zhu2023tpatch}, exploits camera stabilizer resonance to induce motion blur. Electromagnetic Adversarial Injection (EAI) directly disrupts the analog-to-digital conversion to corrupt raw image data~\cite{liao2025your, liu2025magshadow, jiang2023glitchhiker}. However, these attacks have only been evaluated on single-frame detection systems~\cite{zhu2023tpatch, liao2025your}, leaving unexplored their potential for multi-object tracking.

\section{FADE: Adversarial Attack Design}

\subsection{A Primer on TBP-based Trackers}
Consider a video sequence of $T$ frames, denoted as $\mathcal{V} = \{F_1, F_2, \dots, F_T\}$, where $F_t$ is the $t$-th frame. For each frame $F_t$, a vision encoder extracts an image feature map $X_t \in \mathbb{R}^{H \times W \times C}$, where $H, W$ are spatial dimensions and $C$ is feature depth. The TBP pipeline is auto-regressive, using results from $F_{t-1}$ to process $F_t$ as illustrated in Fig.~\ref{fig:framework}(b).

\subsubsection{Query Set Initialization at Frame $F_t$}
At any given frame $F_t$, the input to the decoder consists of two distinct sets of queries:

\noindent \textbf{(\textit{i}) Track Queries ($\mathcal{T}_{t-1}$):} A set of $M \leq B$ active track queries, $\mathcal{T}_{t-1} = \{q_1^{track}, \dots, q_M^{track}\}$, which are propagated from the frame $F_{t-1}$, and $B$ is a fixed query budget. Each $q_i^{track} \in \mathbb{R}^D$ is a $D$-dimensional embedding that encodes the history (motion, appearance) of a specific object $i$.

\noindent \textbf{(\textit{ii}) Detection Queries ($\mathcal{Q}_{det}$):} A set of $N$ newly initialized queries, $\mathcal{Q}_{det} = \{q_1^{det}, \dots, q_N^{det}\}$, which detect new objects entering frame $F_t$.
These two sets are concatenated to form the input query: $\mathcal{Q}_t^{in} = \mathcal{T}_{t-1} \cup \mathcal{Q}_{det}$ of size $M+N$.

\subsubsection{Deformable Transformer Decoder}
The decoder updates the entire query set $\mathcal{Q}_t^{in}$ by processing them jointly with the image features $X_t$. This joint processing is crucial, as it allows queries to perform self-attention to, for example, suppress duplicate detections (e.g., a 'track' query and a 'det' query locking onto the same object).
Let $\mathcal{Q}^{(l)}$ be the queries at layer $l$. The update process is:
$$ \mathcal{Q}^{(l+1)} = \text{TransformerLayer}(\mathcal{Q}^{(l)}, X_t) , 1 \leq l \leq L$$
After $L$ layers, the final output query set is $\mathcal{Q}_t^{out} = \mathcal{Q}^{(L)}$.

\subsubsection{Prediction and Track Query Update}
The output queries $\mathcal{Q}_t^{out} = \{q_i^{out}\}_{i=1}^{M+N}$ are fed into prediction heads to produce the final outputs for frame $t$.

\noindent \textbf{(\textit{i}) Prediction Head:} For each $q_i^{out}$, prediction heads output a bounding box $\hat{b}_i \in \mathbb{R}^4$ and a detection score $\hat{c}_i \in \mathbb{R}^K$.

\noindent \textbf{(\textit{ii}) Track Query Propagation and Update:} This is the core autoregressive component. The output queries $\mathcal{Q}_t^{out}$ that correspond to active tracks are processed by a Query Updater module (e.g., an MLP, SSM) to generate the \textit{next} set of track queries, $\mathcal{T}_t$. Some trackers~\cite{gao2023memotr, segu2024samba} integrate an explicit long-term memory bank $\mathcal{M}_t$. The update process then becomes a joint update of both the memory and the next set of queries, governed by the Query Updater: $$ (\mathcal{T}_t, \mathcal{M}_t) = \text{QueryUpdater}(\mathcal{Q}_t^{out}, \mathcal{M}_{t-1}) $$

\subsection{Adversarial Attacks Formulation}
FADE targets architectural vulnerabilities of TBP trackers: the zero-sum query budget and the recurrent temporal memory in the track query updater logic as depicted in Fig.~\ref{fig:framework}. We formulate two distinct adversarial attacks: (\textit{i}) \textit{Temporal Query Flooding} ($\mathcal{L}_{\text{TQF}}$) to exhaust the query budget for legitimate tracks while disrupting the query's internal states to prevent re-association via propagation. (\textit{ii}) \textit{Temporal Memory Corruption} ($\mathcal{L}_{\text{TMC}}$) to corrupt the track history in the query updater's built-in memory by minimizing pairwise similarity of frame-to-frame query-state embeddings.

\subsubsection{Temporal Query Flooding (TQF)}
The TQF attack exploits the fixed query budget by generating a surplus of high-confidence, temporally consistent spurious tracks. The objective is not to create false positives, but to force these spurious tracks to be integrated into the tracker's query state as persistent tracks, thereby creating a zero-sum allocation problem that makes legitimate tracks seem unreliable. Given the set of all track queries $\mathcal{T}_t$, the tracker's assignment logic partitions it into matched (associated) queries and unmatched (available) queries. The TQF attack targets the set of $K$ unmatched (available) queries, which we designate as the adversarial set $\mathcal{T}_{\text{adv}} \subset \mathcal{T}_t$. $\mathcal{L}_{\text{TQF}}$ is a composite loss with three components:

\begin{figure*}[t]
\centering
\includegraphics[width=.87\textwidth]{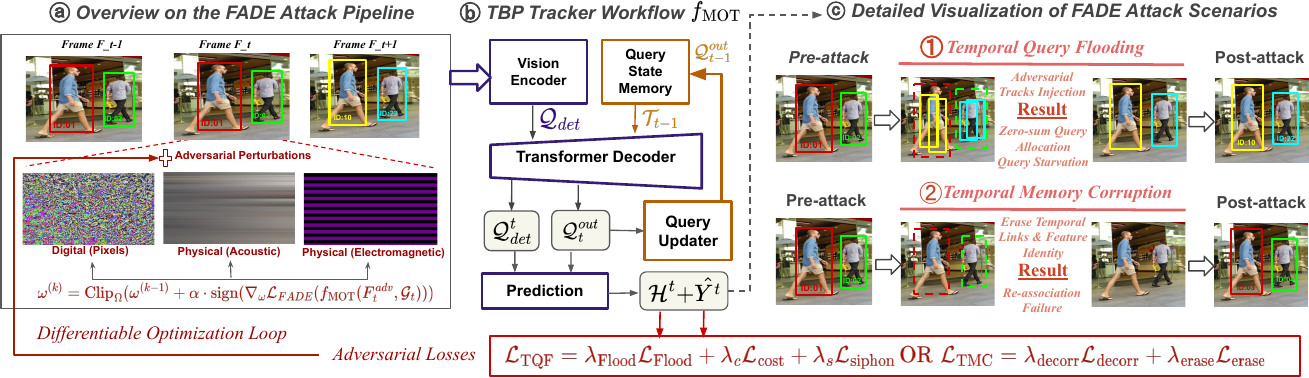}
\vspace{-0.2cm}
\caption{Overview of the FADE Attack Pipeline. \textbf{(a)} A PGD differentiable optimization loop crafts digital or physical perturbations $\omega$. \textbf{(b)} The TBP tracker ($f_{MOT}$) processes the perturbation, and its outputs/hidden-states ($\hat{Y}^t$ and $\mathcal{H}^t$) are used to calculate (bottom) $\mathcal{L}_{\text{FADE}} \in \{\mathcal{L}_{\text{TQF}} , \mathcal{L}_{\text{TMC}}\}$. \textbf{(c)} The visualized results show $\mathcal{L}_{\text{TQF}}$ causing query starvation via spurious track injection and $\mathcal{L}_{\text{TMC}}$ causing re-association failure via temporal memory corruption. The loss gradient is fed back to the optimization loop (a), closing the FADE pipeline.}
\label{fig:framework}
\vspace{-0.4cm}
\end{figure*}

\noindent \textbf{1. Query Flooding ($\mathcal{L}_{\text{Flood}}$).}
First, to exhaust the query budget, the attack maximizes the perceived objectness for all queries $q_i \in \mathcal{T}_{\text{adv}}$. Let $z_i \in \mathbb{R}^C$ be the output logits for query $q_i$. The objective is to minimize the squared error against a perfect confidence score of 1.0:
\begin{equation}
\mathcal{L}_{\text{Flood}} = \mathbb{E}_{q_i \in \mathcal{T}_{\text{adv}}} \left[ (1.0 - \max_c \sigma(z_i)_c)^2 \right]
\end{equation}
Maximizing this objective alone creates naive, single-frame false positives. To ensure they are persistent, the following query state manipulation components are required.

\noindent \textbf{2. Temporal State Amplification.}
This mechanism is designed to make the adversarial queries in $\mathcal{T}_{\text{adv}}$ \textit{reliable} to the query updater and indistinguishable from valid tracks.

\noindent \textbf{(\textit{i}) Cost Mimicry ($\mathcal{L}_{\text{Cost}}$):} To deceive the bipartite matching algorithm, the adversarial queries must present a low matching cost relative to ground-truth objects $\mathcal{G}_t = \{g_j\}$ (or tracker predictions). The attack minimizes the \textit{best} possible match for each ground-truth object from the adversarial set:
\vspace{-0.4cm}
\begin{equation}
\mathcal{L}_{\text{Cost}} = - \mathbb{E}_{g_j \in \mathcal{G}_t} \left[ \log \sum_{q_i \in \mathcal{T}_{\text{adv}}} \exp(-C(q_i, g_j)) \right]
\end{equation}
This encourages the tracker to misassociate a real object's identity with one of the new adversarial queries.

\noindent \textbf{(\textit{ii}) Identity Siphoning ($\mathcal{L}_{\text{Siphon}}$):} To achieve temporal persistence, the attack encourages the hidden state of a \textit{current} adversarial query, $h_i^t \in \mathcal{H}_{\text{adv}}^t$, to become indistinguishable from the hidden states of \textit{previous legitimate tracks}, $\mathcal{H}_{\text{anchor}}^{t-1}$.
\vspace{-0.4cm}
\begin{equation}
\mathcal{L}_{\text{Siphon}} = - \mathbb{E}_{h_i^t \in \mathcal{H}_{\text{adv}}^t, h_j^{t-1} \in \mathcal{H}_{\text{anchor}}^{t-1}} \left[ \text{cosine}(h_i^t, h_j^{t-1}) \right]
\end{equation}
This optimization makes the adversarial query appear to be the re-emergence of a known, tracked object, effectively \textit{siphoning} a legitimate track's identity and history. The full $\mathcal{L}_{\text{TQF}}$ attack is a weighted sum of these components:
\begin{equation}
\mathcal{L}_{\text{TQF}} = \lambda_{\text{Flood}} \mathcal{L}_{\text{Flood}} + \lambda_c \mathcal{L}_{\text{Cost}} + \lambda_s \mathcal{L}_{\text{Siphon}}
\end{equation}

\subsubsection{Temporal Memory Corruption (TMC)}
In contrast to TQF's infiltration strategy, the TMC attack is a direct \textit{forgetting} strategy. It does not create new tracks; rather, it seeks to \textit{destroy} existing legitimate tracks by corrupting the built-in temporal memory in the query updater and the hidden states $h_i \in \mathbb{R}^D$ of \textit{matched} tracks. This attack comprises two objectives:

\noindent \textbf{(\textit{i}) Temporal Decorrelation ($\mathcal{L}_{\text{Decorr}}$):} To sever temporal linkages, this objective \textit{minimizes} the frame-to-frame similarity between the current hidden states $\mathcal{H}^t$ and the previous states $\mathcal{H}^{t-1}$, preventing the query updater from re-associating current tracks with their own history.
\begin{equation}
\mathcal{L}_{\text{Decorr}} = \mathbb{E} \left[cosine(norm(\mathcal{H}^t), norm(\mathcal{H}^{t-1})) \right]
\end{equation}

\noindent \textbf{(\textit{ii}) Track Erasure ($\mathcal{L}_{\text{Erase}}$):} To destroy the identity of active, legitimate tracks, this objective targets the embeddings of all currently \textit{matched} queries, $\mathcal{H}_{\text{matched}}^t$. The attack minimizes the magnitude (L2-norm squared) of these embeddings, forcing them to collapse toward zero.
\begin{equation}
\mathcal{L}_{\text{Erase}} = \mathbb{E}_{h_i \in \mathcal{H}_{\text{matched}}^t} \left[ \Vert h_i \Vert_2^2 \right]
\end{equation}
This erases the representative features of valid tracks, rendering them \textit{untrackable} in subsequent frames. The full TMC attack is a weighted sum of these losses:
\begin{equation}
\mathcal{L}_{\text{TMC}} = \lambda_{\text{Decorr}} \mathcal{L}_{\text{Decorr}} + \lambda_{\text{Erase}} \mathcal{L}_{\text{Erase}}
\end{equation}

\begin{figure*}[t]
\centering
\includegraphics[width=.9\textwidth]{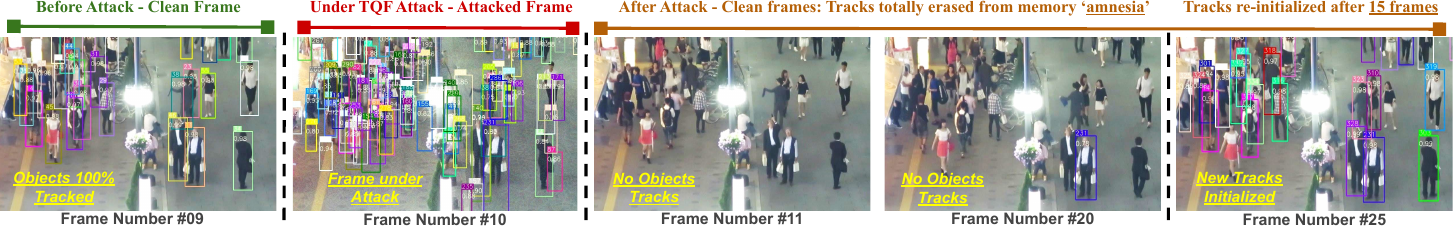}
\vspace{-0.2cm}
\caption{Visualization of the TQF attack effectiveness against MeMOTR~\cite{gao2023memotr} on a video sequence from the MOT17 benchmark.}
\label{fig:attack_vis}
\vspace{-0.4cm}
\end{figure*}

\subsubsection{Digital-to-Physical Adversarial Examples}
We simulate physical attacks from recent spoofing methods on camera sensors using acoustic signals~\cite{zhu2023tpatch,ji2021poltergeist} and electromagnetic signals~\cite{liao2025your, liu2025magshadow, jiang2023glitchhiker}. Our approach optimizes the parameters of a differentiable physical attack simulation model. These parameters have been calibrated on real-world camera sensors and we employ the same range of parameters from previous works~\cite{zhu2023tpatch,liao2025your,liu2025magshadow, jiang2023glitchhiker}. We model the generation of an adversarial example $F_{t}^{adv}$ from the clean input $F_t$ through a differentiable function $\mathcal{P}_{physical}(\cdot; \theta)$, where $\theta$ represents the physical parameters (more details on the physical attacks are provided in the Appendix):
$$F_{t}^{adv} = \mathcal{P}_{physical}(F_t; \theta_{(AAI \parallel EAI)})$$

\noindent \textbf{(\textit{i}) Acoustic Adversarial Injection (AAI).}
The AAI model simulates the motion blur caused by acoustic signals through a differentiable function that models the effects of camera motion~\cite{ji2021poltergeist,zhu2023tpatch}.
Let $\theta_{AAI} = (x, y, \phi, \mathbb{D})$ be the set of physical parameters for the AAI attack. Here, $x$ and $y$ define the maximum amplitudes of horizontal and vertical camera oscillation, respectively. $\phi$ represents a phase offset applied to the sinusoidal motion, influencing the specific pattern of oscillation. 
The parameter $\mathbb{D}$ controls the number of discrete steps over which the motion blur is simulated.

\noindent \textbf{(\textit{ii}) Electromagnetic Adversarial Interference (EAI).}
The EAI model simulates the effects of electromagnetic interference (EMI) that corrupts raw image data in the camera, resulting in visual artifacts~\cite{liao2025your, liu2025magshadow, jiang2023glitchhiker}.
Let $\theta_{EAI} = (\mathbb{M}, \mathbb{N})$ be the set of physical parameters for the EAI attack. Here $\mathbb{M}^{H\times2}$ is a 2D tensor where each row defines a horizontal color stripe by its starting row and its width. The maximum number of stripes, $\mathbb{N}$, determines the intensity of the EAI. 
The corrupted raw sensor effects are blended into the original frame within masking regions defined by $\mathbb{M}$.

\subsubsection{Gradient-based Attack Optimization}
As shown in Fig.~\ref{fig:framework}, we formulate the optimization for both digital and physical-simulation attacks within a unified Projected Gradient Descent (PGD) framework. The objective is to find a set of differentiable parameters, $\omega \in \Omega$, that maximizes the chosen adversarial loss, $\mathcal{L}_{\text{adv}} \in \{\mathcal{L}_{\text{TQF}}, \mathcal{L}_{\text{TMC}}\}$.
Let $f_{\text{MOT}}$ be the TBP tracker and $\mathcal{G}_t$ be the ground-truth data. We provide more details on FADE's optimization in the Appendix. The general optimization problem is:
\begin{equation}
\max_{\omega \in \Omega} \mathcal{L}_{\text{adv}}(f_{\text{MOT}}(F_t^{adv}(\omega)), \mathcal{G}_t)
\end{equation}

\noindent \textbf{(\textit{i}) Digital Attack:} The parameters $\omega$ are the pixel-wise perturbations $\delta_{\text{dig}}$, and the constraint set $\Omega$ is an $L_p$-norm ball $\Delta$ (e.g., $\Vert \delta_{\text{dig}} \Vert_\infty \leq \epsilon$) and $F_t^{adv}(\delta_{\text{dig}}) = \text{Clip}_{[0,1]}(F_t + \delta_{\text{dig}})$.

\noindent \textbf{(\textit{ii}) Physical Attack:} The parameters $\omega$ are the AAI and EAI attack parameters $\theta_{(AAI \parallel EAI)}$. The constraint set $\Omega$ is their plausible range $\Theta$ calibrated on real-world sensors as detailed in~\cite{zhu2023tpatch,liao2025your,liu2025magshadow, jiang2023glitchhiker}, and $F_t^{adv}(\theta_{\text{phys}}) = \mathcal{P}(F_t, \theta_{\text{phys}})$ is the output of a differentiable sensor-spoofing simulation model $\mathcal{P}(\cdot)$. We solve this maximization using PGD. The update for the parameters $\omega^{(k)}$ at iteration $k$ is:
$$
\omega^{(k)} = \text{Clip}_{\Omega} (\omega^{(k-1)} + \alpha \cdot \text{sign}(\nabla_{\omega} \mathcal{L}_{\text{adv}}(f_{\text{MOT}}(F_t^{adv}, \mathcal{G}_t)))
$$

\section{Evaluation}

\subsection{Experimental Setup}

\noindent \textbf{1. Datasets and Trackers.} 
We conduct experiments on two MOT datasets for pedestrian detection: MOT17~\cite{sun2019deep}, MOT20~\cite{dendorfer2020mot20}. MOT17 features frequent occlusions and similar objects. MOT20 focuses on extremely dense crowd scenes with severe occlusions. We study various state-of-the-art TBP trackers, including MOTR~\cite{zeng2022motr}, MOTRv2~\cite{zhang2023motrv2}, MeMOTR~\cite{gao2023memotr}, Samba~\cite{segu2024samba}, and the recent CO-MOT~\cite{yan2023bridging}.

\noindent \textbf{2. Metrics.}
We assess overall tracking performance using the standard HOTA, DetA, and AssA metrics~\cite{luiten2021hota}, as well as IDF1, IDR, IDP, and the rate of Identity Switches (IDSW)~\cite{ristani2016performance}. HOTA is our primary metric as it balances both detection and long-term association quality. 

\noindent \textbf{3. Simulated Physical Attacks.}
We implement the AAI model from Poltergeist~\cite{ji2021poltergeist} and TPatch~\cite{zhu2023tpatch}, calibrated on camera sensors (e.g., Samsung S20, iPhone 7). Parameters $\theta_{\text{AAI}}$ define the attack effect (e.g., blur kernel intensity, direction). We fix the number of blur samples to $\mathbb{D}=10$. For EAI, we implement models from GlitchHiker~\cite{jiang2023glitchhiker,liao2025your}, which defines the color stripe artifacts caused by EMI, calibrated on camera sensors (e.g., 360 M320 Dashcam, Google Pixel One). Our attack parameters $\theta_{\text{EAI}}$ parameterize the visual artifacts themselves (e.g., the location, color, and intensity of the $\mathbb{N}$ color stripes). We fix $\mathbb{N} = 20$. Both simulation models, $\mathcal{P}_{\text{AAI}}$ and $\mathcal{P}_{\text{EAI}}$, are fully differentiable with respect to these parameters ($\theta_{\text{AAI}}$, $\theta_{\text{EAI}}$). To ensure physical plausibility, the optimization search space for these parameters ($\theta_{\text{AAI}}$, $\theta_{\text{EAI}}$) is constrained (Appendix). The parameters governing blur intensity/direction and stripe color/location are bounded to match the magnitude and characteristics of artifacts physically demonstrated in the previous works~\cite{ji2021poltergeist, zhu2023tpatch, jiang2023glitchhiker, liao2025your}. This ensures our pipeline does not find an arbitrary, unrealistic optimum, but rather optimizes for the \textit{most damaging plausible artifact} within a physically-grounded threat model.

\noindent \textbf{4. Attack Optimization Details.}
Similar PGD parameters have been used on all trackers. For digital attacks, we set $\epsilon$ to $8/255$ and $\alpha$ to $1/255$. For simulated physical attacks, we set $\alpha$ to $8/255$. We run digital and simulated physical attacks for $T=50$ and $T=100$ iterations. We apply each adversarial perturbation for $\Delta_{\text{attack}} = 1$~frame for digital attacks and $\Delta_\text{attack} = 3$~frames for simulated physical attacks. PGD optimization takes $\sim$3s (RTX 3090) per frame.

\begin{table}[t!]
\centering
\scriptsize
\setlength{\tabcolsep}{3.2pt}
\renewcommand{\arraystretch}{0.7}
\caption{Digital Attack Performance on MOT17.}  
\label{tab:results_mot17}
\vspace{-0.3cm}
\scalebox{.75}{
\begin{tabular}{llccccccc}
\toprule
\rowcolor{HeaderGray}
\multicolumn{9}{c}{
\textbf{MOT17 Dataset – Average scene density $\sim$21~objects/frame.}
}\\[-0.2em]
\midrule
\textbf{Tracker} & \textbf{Attacker} & \textbf{HOTA}$\downarrow$ & \textbf{DetA}$\downarrow$ & \textbf{AssA}$\downarrow$ & \textbf{IDF1}$\downarrow$ & \textbf{IDR}$\downarrow$ & \textbf{IDP}$\downarrow$ & \textbf{IDSW}$\uparrow$ \\
\midrule
\multirow{6}{*}{\textbf{MOTR}~\cite{zeng2022motr}}
 & {\cellcolor{CleanGreen}}Clean & {\cellcolor{CleanGreen}}58.63 & {\cellcolor{CleanGreen}}49.90 & {\cellcolor{CleanGreen}}70.38 & {\cellcolor{CleanGreen}}69.35 & {\cellcolor{CleanGreen}}68.48 & {\cellcolor{CleanGreen}}71.40 & {\cellcolor{CleanGreen}}7.23 \\
 & Daedalus~\cite{wang2021daedalus} & 45.85 & 51.26 & 42.19 & 50.48 & 46.37 & 55.88 & 8.74 \\
 & Hijacking~\cite{jia2020fooling} & 46.08 & 51.24 & 42.61 & 50.77 & 46.78 & 55.97 & 8.89 \\
 & F\&F~\cite{zhou2023f} & 46.59 & 51.10 & 43.71 & 51.55 & 47.44 & 56.92 & 8.71 \\
 & \textbf{$\text{FADE}_{\text{TMC}}$} & \textbf{\textcolor{DeepRed}{45.89}} & \textbf{\textcolor{DeepRed}{51.36}} & \textbf{\textcolor{DeepRed}{42.18}} & \textbf{\textcolor{DeepRed}{50.45}} & \textbf{\textcolor{DeepRed}{46.35}} & \textbf{\textcolor{DeepRed}{55.79}} & \textbf{\textcolor{DeepRed}{8.77}} \\
 & \textbf{$\text{FADE}_{\text{TQF}}$} & \textbf{\textcolor{DeepRed}{45.90}} & \textbf{\textcolor{DeepRed}{51.40}} & \textbf{\textcolor{DeepRed}{42.17}} & \textbf{\textcolor{DeepRed}{50.51}} & \textbf{\textcolor{DeepRed}{46.40}} & \textbf{\textcolor{DeepRed}{55.86}} & \textbf{\textcolor{DeepRed}{8.73}} \\
\midrule
\multirow{6}{*}{\textbf{MOTRv2}~\cite{zhang2023motrv2}}
 & {\cellcolor{CleanGreen}}Clean & {\cellcolor{CleanGreen}}59.96 & {\cellcolor{CleanGreen}}49.15 & {\cellcolor{CleanGreen}}74.71 & {\cellcolor{CleanGreen}}71.99 & {\cellcolor{CleanGreen}}60.89 & {\cellcolor{CleanGreen}}89.68 & {\cellcolor{CleanGreen}}1.75 \\
 & Daedalus~\cite{wang2021daedalus} & 39.21 & 31.55 & 49.97 & 48.46 & 35.15 & 79.75 & 6.13 \\
 & Hijacking~\cite{jia2020fooling} & 56.31 & 46.13 & 70.146 & 68.25 & 56.34 & 88.34 & 2.02 \\
 & F\&F~\cite{zhou2023f} & 55.60 & 45.52 & 69.27 & 67.70 & 55.91 & 87.42 & 2.67 \\
 & \textbf{$\text{FADE}_{\text{TMC}}$} & \textbf{\textcolor{DeepRed}{46.76}} & \textbf{\textcolor{DeepRed}{37.63}} & \textbf{\textcolor{DeepRed}{59.44}} & \textbf{\textcolor{DeepRed}{56.22}} & \textbf{\textcolor{DeepRed}{42.80}} & \textbf{\textcolor{DeepRed}{83.89}} & \textbf{\textcolor{DeepRed}{3.83}} \\
 & \textbf{$\text{FADE}_{\text{TQF}}$} & \textbf{\textcolor{DeepRed}{39.29}} & \textbf{\textcolor{DeepRed}{31.68}} & \textbf{\textcolor{DeepRed}{49.96}} & \textbf{\textcolor{DeepRed}{49.02}} & \textbf{\textcolor{DeepRed}{35.87}} & \textbf{\textcolor{DeepRed}{78.77}} & \textbf{\textcolor{DeepRed}{5.65}} \\
\midrule
\multirow{6}{*}{\textbf{MeMOTR}~\cite{gao2023memotr}}
 & {\cellcolor{CleanGreen}}Clean & {\cellcolor{CleanGreen}}67.35 & {\cellcolor{CleanGreen}}57.87 & {\cellcolor{CleanGreen}}79.60 & {\cellcolor{CleanGreen}}80.83 & {\cellcolor{CleanGreen}}70.78 & {\cellcolor{CleanGreen}}94.84 & {\cellcolor{CleanGreen}}0.81 \\
 & Daedalus~\cite{wang2021daedalus} & 42.41 & 35.24 & 51.94 & 52.46 & 37.96 & 86.34 & 4.09 \\
 & Hijacking~\cite{jia2020fooling} & 55.92 & 49.67 & 64.14 & 66.91 & 55.00 & 86.38 & 2.48 \\
 & F\&F~\cite{zhou2023f} & 54.53 & 46.67 & 64.79 & 66.77 & 53.53 & 89.76 & 2.08 \\
 & \textbf{$\text{FADE}_{\text{TMC}}$} & \textbf{\textcolor{DeepRed}{41.56}} & \textbf{\textcolor{DeepRed}{35.74}} & \textbf{\textcolor{DeepRed}{49.18}} & \textbf{\textcolor{DeepRed}{51.60}} & \textbf{\textcolor{DeepRed}{37.61}} & \textbf{\textcolor{DeepRed}{84.07}} & \textbf{\textcolor{DeepRed}{4.63}} \\
 & \textbf{$\text{FADE}_{\text{TQF}}$} & \textbf{\textcolor{DeepRed}{41.41}} & \textbf{\textcolor{DeepRed}{34.83}} & \textbf{\textcolor{DeepRed}{50.03}} & \textbf{\textcolor{DeepRed}{51.41}} & \textbf{\textcolor{DeepRed}{37.07}} & \textbf{\textcolor{DeepRed}{85.38}} & \textbf{\textcolor{DeepRed}{4.31}} \\
\midrule
\multirow{6}{*}{\textbf{Samba}~\cite{segu2024samba}}
 & {\cellcolor{CleanGreen}}Clean & {\cellcolor{CleanGreen}}62.91 & {\cellcolor{CleanGreen}}50.58 & {\cellcolor{CleanGreen}}79.37 & {\cellcolor{CleanGreen}}73.67 & {\cellcolor{CleanGreen}}60.30 & {\cellcolor{CleanGreen}}95.93 & {\cellcolor{CleanGreen}}1.02 \\
 & Daedalus~\cite{wang2021daedalus} & 54.40 & 42.95 & 70.11 & 64.92 & 50.11 & 93.95 & 1.57 \\
 & Hijacking~\cite{jia2020fooling} & 57.40 & 46.66 & 71.77 & 68.34 & 54.33 & 93.60 & 1.45 \\
 & F\&F~\cite{zhou2023f} & 52.78 & 42.43 & 67.02 & 62.54 & 48.17 & 91.80 & 1.76 \\
 & \textbf{$\text{FADE}_{\text{TMC}}$} & \textbf{\textcolor{DeepRed}{48.04}} & \textbf{\textcolor{DeepRed}{45.19}} & \textbf{\textcolor{DeepRed}{51.93}} & \textbf{\textcolor{DeepRed}{56.01}} & \textbf{\textcolor{DeepRed}{44.01}} & \textbf{\textcolor{DeepRed}{78.74}} & \textbf{\textcolor{DeepRed}{3.63}} \\
 & \textbf{$\text{FADE}_{\text{TQF}}$} & \textbf{\textcolor{DeepRed}{45.53}} & \textbf{\textcolor{DeepRed}{32.71}} & \textbf{\textcolor{DeepRed}{64.37}} & \textbf{\textcolor{DeepRed}{52.71}} & \textbf{\textcolor{DeepRed}{37.62}} & \textbf{\textcolor{DeepRed}{91.33}} & \textbf{\textcolor{DeepRed}{2.24}} \\
\midrule
\multirow{6}{*}{\textbf{CO-MOT}~\cite{yan2023bridging}}
 & {\cellcolor{CleanGreen}}Clean & {\cellcolor{CleanGreen}}58.16 & {\cellcolor{CleanGreen}}46.22 & {\cellcolor{CleanGreen}}74.87 & {\cellcolor{CleanGreen}}69.87 & {\cellcolor{CleanGreen}}57.21 & {\cellcolor{CleanGreen}}91.97 & {\cellcolor{CleanGreen}}1.83 \\
 & Daedalus~\cite{wang2021daedalus} & 40.01 & 31.29 & 52.41 & 47.61 & 39.22 & 62.72 & 14.59 \\
 & Hijacking~\cite{jia2020fooling} & 51.68 & 40.63 & 67.20 & 62.95 & 50.40 & 86.02 & 3.93 \\
 & F\&F~\cite{zhou2023f} & 52.78 & 41.55 & 68.65 & 64.08 & 51.05 & 88.15 & 3.75 \\
 & \textbf{$\text{FADE}_{\text{TMC}}$} & \textbf{\textcolor{DeepRed}{41.73}} & \textbf{\textcolor{DeepRed}{31.82}} & \textbf{\textcolor{DeepRed}{55.89}} & \textbf{\textcolor{DeepRed}{50.34}} & \textbf{\textcolor{DeepRed}{39.23}} & \textbf{\textcolor{DeepRed}{72.34}} & \textbf{\textcolor{DeepRed}{10.94}} \\
 & \textbf{$\text{FADE}_{\text{TQF}}$} & \textbf{\textcolor{DeepRed}{37.26}} & \textbf{\textcolor{DeepRed}{27.43}} & \textbf{\textcolor{DeepRed}{51.93}} & \textbf{\textcolor{DeepRed}{44.84}} & \textbf{\textcolor{DeepRed}{32.88}} & \textbf{\textcolor{DeepRed}{74.66}} & \textbf{\textcolor{DeepRed}{9.50}} \\
\bottomrule
\end{tabular}
}
\vspace{-0.2cm}
\end{table}

\begin{table}[t!]
\centering
\scriptsize
\setlength{\tabcolsep}{3.2pt}
\renewcommand{\arraystretch}{0.7}
\caption{Digital Attack Performance on MOT20.}  
\label{tab:results_mot20}
\vspace{-0.3cm}
\scalebox{.75}{
\begin{tabular}{llccccccc}
\toprule
\rowcolor{HeaderGray}
\multicolumn{9}{c}{
\textbf{MOT20 Dataset – High Density: Avg. $\sim$150 objects/frame.}
}\\[-0.2em]
\midrule
\textbf{Tracker} & \textbf{Attacker} & \textbf{HOTA}$\downarrow$ & \textbf{DetA}$\downarrow$ & \textbf{AssA}$\downarrow$ & \textbf{IDF1}$\downarrow$ & \textbf{IDR}$\downarrow$ & \textbf{IDP}$\downarrow$ & \textbf{IDSW}$\uparrow$ \\
\midrule
\multirow{6}{*}{\textbf{MOTR}~\cite{zeng2022motr}}
 & {\cellcolor{CleanGreen}}Clean & {\cellcolor{CleanGreen}}55.06 & {\cellcolor{CleanGreen}}40.57 & {\cellcolor{CleanGreen}}75.89 & {\cellcolor{CleanGreen}}64.10 & {\cellcolor{CleanGreen}}57.65 & {\cellcolor{CleanGreen}}74.80 & {\cellcolor{CleanGreen}}5.14 \\
 & Daedalus~\cite{wang2021daedalus} & 42.07 & 41.54 & 43.63 & 49.30 & 45.22 & 54.37 & 8.78 \\
 & Hijacking~\cite{jia2020fooling} & 43.62 & 38.04 & 51.64 & 51.46 & 45.43 & 60.22 & 9.02 \\
 & F\&F~\cite{zhou2023f} & 41.36 & 31.84 & 55.04 & 47.94 & 39.30 & 64.88 & 7.72 \\
 & \textbf{$\text{FADE}_{\text{TMC}}$} & \textbf{\textcolor{DeepRed}{40.38}} & \textbf{\textcolor{DeepRed}{39.70}} & \textbf{\textcolor{DeepRed}{42.24}} & \textbf{\textcolor{DeepRed}{47.90}} & \textbf{\textcolor{DeepRed}{38.96}} & \textbf{\textcolor{DeepRed}{64.03}} & \textbf{\textcolor{DeepRed}{6.61}} \\
 & \textbf{$\text{FADE}_{\text{TQF}}$} & \textbf{\textcolor{DeepRed}{42.63}} & \textbf{\textcolor{DeepRed}{39.25}} & \textbf{\textcolor{DeepRed}{47.74}} & \textbf{\textcolor{DeepRed}{49.93}} & \textbf{\textcolor{DeepRed}{50.73}} & \textbf{\textcolor{DeepRed}{49.25}} & \textbf{\textcolor{DeepRed}{12.46}} \\
\midrule
\multirow{6}{*}{\textbf{MOTRv2}~\cite{zhang2023motrv2}}
 & {\cellcolor{CleanGreen}}Clean & {\cellcolor{CleanGreen}}59.56 & {\cellcolor{CleanGreen}}44.74 & {\cellcolor{CleanGreen}}80.71 & {\cellcolor{CleanGreen}}69.55 & {\cellcolor{CleanGreen}}54.90 & {\cellcolor{CleanGreen}}96.81 & {\cellcolor{CleanGreen}}0.73 \\
 & Daedalus~\cite{wang2021daedalus} & 29.25 & 24.15 & 36.82 & 38.44 & 25.16 & 82.96 & 4.90 \\
 & Hijacking~\cite{jia2020fooling} & 58.27 & 45.76 & 75.31 & 70.31 & 55.75 & 96.32 & 0.93 \\
 & F\&F~\cite{zhou2023f} & 58.85 & 45.25 & 77.80 & 70.05 & 55.39 & 96.85 & 0.73 \\
 & \textbf{$\text{FADE}_{\text{TMC}}$} & \textbf{\textcolor{DeepRed}{56.68}} & \textbf{\textcolor{DeepRed}{46.32}} & \textbf{\textcolor{DeepRed}{70.72}} & \textbf{\textcolor{DeepRed}{70.10}} & \textbf{\textcolor{DeepRed}{56.63}} & \textbf{\textcolor{DeepRed}{95.97}} & \textbf{\textcolor{DeepRed}{1.45}} \\
 & \textbf{$\text{FADE}_{\text{TQF}}$} & \textbf{\textcolor{DeepRed}{29.64}} & \textbf{\textcolor{DeepRed}{24.98}} & \textbf{\textcolor{DeepRed}{36.57}} & \textbf{\textcolor{DeepRed}{38.90}} & \textbf{\textcolor{DeepRed}{25.62}} & \textbf{\textcolor{DeepRed}{81.87}} & \textbf{\textcolor{DeepRed}{5.10}} \\
\midrule
\multirow{6}{*}{\textbf{MeMOTR}~\cite{gao2023memotr}}
 & {\cellcolor{CleanGreen}}Clean & {\cellcolor{CleanGreen}}69.61 & {\cellcolor{CleanGreen}}59.19 & {\cellcolor{CleanGreen}}83.17 & {\cellcolor{CleanGreen}}83.31 & {\cellcolor{CleanGreen}}72.84 & {\cellcolor{CleanGreen}}97.99 & {\cellcolor{CleanGreen}}0.46 \\
 & Daedalus~\cite{wang2021daedalus} & 62.51 & 53.13 & 74.68 & 77.08 & 64.57 & 96.32 & 1.03 \\
 & Hijacking~\cite{jia2020fooling} & 57.22 & 47.53 & 70.01 & 69.21 & 55.30 & 93.51 & 1.60 \\
 & F\&F~\cite{zhou2023f} & 44.54 & 39.20 & 51.61 & 54.78 & 40.82 & 83.55 & 3.17 \\
 & \textbf{$\text{FADE}_{\text{TMC}}$} & \textbf{\textcolor{DeepRed}{37.70}} & \textbf{\textcolor{DeepRed}{32.86}} & \textbf{\textcolor{DeepRed}{44.18}} & \textbf{\textcolor{DeepRed}{45.73}} & \textbf{\textcolor{DeepRed}{32.17}} & \textbf{\textcolor{DeepRed}{79.29}} & \textbf{\textcolor{DeepRed}{4.90}} \\
 & \textbf{$\text{FADE}_{\text{TQF}}$} & \textbf{\textcolor{DeepRed}{57.67}} & \textbf{\textcolor{DeepRed}{47.66}} & \textbf{\textcolor{DeepRed}{70.91}} & \textbf{\textcolor{DeepRed}{70.70}} & \textbf{\textcolor{DeepRed}{56.91}} & \textbf{\textcolor{DeepRed}{94.99}} & \textbf{\textcolor{DeepRed}{1.51}} \\
\midrule
\multirow{6}{*}{\textbf{Samba}~\cite{segu2024samba}}
 & {\cellcolor{CleanGreen}}Clean 
   & {\cellcolor{CleanGreen}}62.49 
   & {\cellcolor{CleanGreen}}47.83 
   & {\cellcolor{CleanGreen}}83.29 
   & {\cellcolor{CleanGreen}}72.47 
   & {\cellcolor{CleanGreen}}58.23 
   & {\cellcolor{CleanGreen}}98.27 
   & {\cellcolor{CleanGreen}}0.43 \\
 & Daedalus~\cite{wang2021daedalus} 
   & 59.86 & 47.25 & 76.96 & 71.93 & 57.21 & 97.66 & 0.76 \\
 & Hijacking~\cite{jia2020fooling} 
   & 54.70 & 38.76 & 78.98 & 62.17 & 46.39 & 98.09 & 0.46 \\
 & F\&F~\cite{zhou2023f} 
   & 60.08 & 47.74 & 76.70 & 72.53 & 57.86 & 97.82 & 0.85 \\
 & \textbf{$\text{FADE}_{\text{TMC}}$} 
   & \textbf{\textcolor{DeepRed}{54.91}} 
   & \textbf{\textcolor{DeepRed}{38.86}} 
   & \textbf{\textcolor{DeepRed}{79.77}} 
   & \textbf{\textcolor{DeepRed}{62.24}} 
   & \textbf{\textcolor{DeepRed}{46.65}} 
   & \textbf{\textcolor{DeepRed}{98.22}} 
   & \textbf{\textcolor{DeepRed}{0.47}} \\
 & \textbf{$\text{FADE}_{\text{TQF}}$} 
   & \textbf{\textcolor{DeepRed}{46.85}} 
   & \textbf{\textcolor{DeepRed}{12.40}} 
   & \textbf{\textcolor{DeepRed}{66.34}} 
   & \textbf{\textcolor{DeepRed}{22.70}} 
   & \textbf{\textcolor{DeepRed}{13.78}} 
   & \textbf{\textcolor{DeepRed}{95.19}} 
   & \textbf{\textcolor{DeepRed}{1.19}} \\
\midrule
\multirow{6}{*}{\textbf{CO-MOT}~\cite{yan2023bridging}}
 & {\cellcolor{CleanGreen}}Clean & {\cellcolor{CleanGreen}}64.31 & {\cellcolor{CleanGreen}}52.30 & {\cellcolor{CleanGreen}}80.35 & {\cellcolor{CleanGreen}}78.01 & {\cellcolor{CleanGreen}}65.43 & {\cellcolor{CleanGreen}}97.78 & {\cellcolor{CleanGreen}}0.45 \\
 & Daedalus~\cite{wang2021daedalus} & 33.49 & 26.94 & 43.58 & 41.08 & 27.68 & 84.65 & 4.86 \\
 & Hijacking~\cite{jia2020fooling} & 57.70 & 45.46 & 74.41 & 71.03 & 56.32 & 97.28 & 0.78 \\
 & F\&F~\cite{zhou2023f} & 53.80 & 41.05 & 71.71 & 66.15 & 50.76 & 97.10 & 1.15 \\
 & \textbf{$\text{FADE}_{\text{TMC}}$} & \textbf{\textcolor{DeepRed}{49.28}} & \textbf{\textcolor{DeepRed}{37.02}} & \textbf{\textcolor{DeepRed}{66.96}} & \textbf{\textcolor{DeepRed}{60.46}} & \textbf{\textcolor{DeepRed}{45.00}} & \textbf{\textcolor{DeepRed}{95.82}} & \textbf{\textcolor{DeepRed}{1.68}} \\
 & \textbf{$\text{FADE}_{\text{TQF}}$} & \textbf{\textcolor{DeepRed}{33.37}} & \textbf{\textcolor{DeepRed}{25.91}} & \textbf{\textcolor{DeepRed}{44.64}} & \textbf{\textcolor{DeepRed}{40.90}} & \textbf{\textcolor{DeepRed}{27.16}} & \textbf{\textcolor{DeepRed}{85.69}} & \textbf{\textcolor{DeepRed}{4.86}} \\
\bottomrule
\end{tabular}
}
\vspace{-0.4cm}
\end{table}

\subsection{Quantitative Evaluation of Digital Attacks.}
\label{sec:digital_results}
The quantitative results across MOT17 (Table~\ref{tab:results_mot17}) and MOT20 (Table~\ref{tab:results_mot20}) demonstrate the effectiveness of FADE's attacks. Each attack degrades its intended architectural target.
First, the TMC strategy, designed to induce \textit{forgetting} via state de-correlation ($\mathcal{L}_{\text{Decorr}}$) and feature erasure ($\mathcal{L}_{\text{Erase}}$), proves exceptionally powerful against memory-augmented trackers, specifically MeMOTR~\cite{gao2023memotr} and Samba~\cite{segu2024samba}. On MOT20, $\text{FADE}_{\text{TMC}}$ is the most effective attack against MeMOTR, causing its HOTA to decrease by $31.5$~points and spiking its IDSW by over $10\times$. Similarly, on MOT17, $\text{FADE}_{\text{TMC}}$ is the second most effective attack against Samba decreasing HOTA by  $14.78$~points and increasing IDSW by over $2.3\times$.
This confirms that the attack successfully forces re-association failures by directly corrupting the tracker's temporal memory.
The TQF strategy, which targets the zero-sum query budget, induces the most severe collapse on trackers without a robust temporal memory mechanism. 
On MOT17, $\text{FADE}_{\text{TQF}}$ is the most effective attack against CO-MOT with a $20.9$~point HOTA drop and over $5.2\times$ increase in IDSW. 
This degradation stems from a simultaneous collapse in detection and association, confirming that $\text{FADE}_{\text{TQF}}$ starves legitimate tracks of query resources.

\subsection{Comparison with SOTA Attacks.}
\label{sec:sota_comparison}
The comparison in Tables~\ref{tab:results_mot17} and \ref{tab:results_mot20} indicates that previous attacks designed for TBD are not effective for TBP trackers. This is evident on the high-density MOT20.
Association-focused attacks like Hijack~\cite{jia2020fooling} and F\&F~\cite{zhou2023f} are rendered ineffective against MOTRv2, with HOTA remaining unchanged ($59.56$ to $58.27$). 
This contrasts with $\text{FADE}_{\text{TQF}}$'s $30$~points HOTA degradation ($59.56$ to $29.64$), proving that targeting hand-crafted motion models is less effective against TBP's learned query propagation. Furthermore, FADE's strategies are demonstrably superior to naive flooding attacks like Daedalus~\cite{wang2021daedalus}. 
This is evident on advanced and memory-augmented trackers like MeMOTR (MOT20), Samba (MOT17), and CO-MOT (MOT17). 
On MeMOTR, Daedalus has only a marginal effect, proving that advanced TBP trackers can filter non-persistent noise. 
In contrast, $\text{FADE}_{\text{TMC}}$, which is engineered to corrupt memory mechanisms, achieves a $24$~points stronger degradation.
FADE's main contribution is its ability to defeat the temporal robustness that renders simple flooders (Daedalus) ineffective, proving that exploitation of TBP's \textit{unique} vulnerabilities is important for adversarial intervention.

\subsection{Qualitative Analysis of Digital Attacks.}
We visualize the qualitative impact of our attack in Fig.~\ref{fig:attack_vis}. This figure illustrates the persistent effect of a single-frame $\text{FADE}_{\text{TQF}}$ attack on MeMOTR.
In the benign state (Frame \#01), MeMOTR exhibits stable tracking of all pedestrians. At Frame \#10, the $\text{FADE}_{\text{TQF}}$ attack is applied, injecting spurious and high-confidence queries. The immediate result in Frame \#11 is not just query starvation, but a significant state propagation failure. The attack successfully corrupts the tracker's query state memory, causing it to drop \textit{all} tracks, both legitimate and adversarial. More importantly, this failure is non-transient. The tracker enters a persistent \textit{forgetting} period, failing to recover any tracks for many subsequent frames (as seen in Frame \#20), despite the frames returning to a clean state. The tracker only begins reinitializing  \textit{newly initialized tracks} at Frame \#25, a full 15 frames after the attack concluded. This demonstrates that FADE does not merely cause a temporary glitch; it induces a profound and lasting corruption of the TBP temporal logic, forcing a complete system reset.

\subsection{Quantitative Evaluation of Physical Attacks}
\label{sec:physical_results}
We validate our differentiable pipeline by optimizing $\text{FADE}_{\text{TMC}}$ strategy using simulated physical attacks, referred to as TMC\_AAI and TMC\_EAI, in Tables~\ref{tab:phy_results_mot17} and \ref{tab:phy_results_mot20}.

\noindent \textbf{1. Physical Perturbations Disrupt Association.}
The results prove that optimizing physical-world effects (blur, noise) can be translated into a sophisticated temporal-layer attack. The attack's impact is not merely on detection, but on association. This is evident on the high-density MOT20 dataset. 
On MOTR, the AAI attack causes a catastrophic $36$~points collapse in AssA, leading to a $19$~points drop in HOTA. Similarly, on MOT17, MOTR and Samba suffer the largest HOTA degradations ($13.5$ and $13.9$~points). 

\noindent \textbf{2. Architectural Filtering in Robust Trackers.}
While all trackers are degraded, architectures with stronger, externally-trained detection backbones (e.g., MOTRv2 and CO-MOT) exhibit some resilience. On MOT20, these trackers see a more modest $7$--$9$~points drop in HOTA. This suggests their robust detection heads can filter the low-level visual artifacts from AAI/EAI. 
However, this filtering mechanism is not perfect. Even on these robust trackers, the attack consistently causes a significant $8$--$9$~points drop in AssA. This proves that even the small amount of optimized noise that leaks past the detector is still highly effective at poisoning the downstream temporal query states.

\begin{table}[t!]
\centering
\scriptsize
\setlength{\tabcolsep}{3.2pt}
\renewcommand{\arraystretch}{0.7}
\caption{Simulated Physical Attack Performance on MOT17.}  
\label{tab:phy_results_mot17}
\vspace{-0.3cm}
\scalebox{.75}{
\begin{tabular}{llccccccc}
\toprule
\rowcolor{HeaderGray}
\multicolumn{9}{c}{
\textbf{MOT17 Dataset – Average scene density $\sim$21~objects/frame.}
}\\[-0.2em]
\midrule
\textbf{Tracker} & \textbf{Attack\_Vector} & \textbf{HOTA}$\downarrow$ & \textbf{DetA}$\downarrow$ & \textbf{AssA}$\downarrow$ & \textbf{IDF1}$\downarrow$ & \textbf{IDR}$\downarrow$ & \textbf{IDP}$\downarrow$ & \textbf{IDSW}$\uparrow$ \\
\midrule
\multirow{3}{*}{\textbf{MOTR} \cite{zeng2022motr}}
 & {\cellcolor{CleanGreen}}Clean & {\cellcolor{CleanGreen}}58.63 & {\cellcolor{CleanGreen}}49.90 & {\cellcolor{CleanGreen}}70.38 & {\cellcolor{CleanGreen}}69.35 & {\cellcolor{CleanGreen}}68.48 & {\cellcolor{CleanGreen}}71.40 & {\cellcolor{CleanGreen}}7.23 \\
 & TMC\_AAI & \textbf{\textcolor{DeepRed}{45.09}} & \textbf{\textcolor{DeepRed}{47.37}} & \textbf{\textcolor{DeepRed}{44.21}} & \textbf{\textcolor{DeepRed}{51.57}} & \textbf{\textcolor{DeepRed}{46.44}} & \textbf{\textcolor{DeepRed}{58.43}} & \textbf{\textcolor{DeepRed}{8.64}} \\
 & TMC\_EAI & \textbf{\textcolor{DeepRed}{42.81}} & \textbf{\textcolor{DeepRed}{47.87}} & \textbf{\textcolor{DeepRed}{39.37}} & \textbf{\textcolor{DeepRed}{47.76}} & \textbf{\textcolor{DeepRed}{42.77}} & \textbf{\textcolor{DeepRed}{54.55}} & \textbf{\textcolor{DeepRed}{9.22}} \\
\midrule
\multirow{3}{*}{\textbf{MOTRv2} \cite{zhang2023motrv2}}
 & {\cellcolor{CleanGreen}}Clean & {\cellcolor{CleanGreen}}59.96 & {\cellcolor{CleanGreen}}49.15 & {\cellcolor{CleanGreen}}74.71 & {\cellcolor{CleanGreen}}71.99 & {\cellcolor{CleanGreen}}60.89 & {\cellcolor{CleanGreen}}89.68 & {\cellcolor{CleanGreen}}1.75 \\
 & TMC\_AAI & \textbf{\textcolor{DeepRed}{53.66}} & \textbf{\textcolor{DeepRed}{44.12}} & \textbf{\textcolor{DeepRed}{66.57}} & \textbf{\textcolor{DeepRed}{66.61}} & \textbf{\textcolor{DeepRed}{53.83}} & \textbf{\textcolor{DeepRed}{88.96}} & \textbf{\textcolor{DeepRed}{2.00}} \\
 & TMC\_EAI & \textbf{\textcolor{DeepRed}{53.52}} & \textbf{\textcolor{DeepRed}{43.90}} & \textbf{\textcolor{DeepRed}{66.56}} & \textbf{\textcolor{DeepRed}{66.50}} & \textbf{\textcolor{DeepRed}{53.66}} & \textbf{\textcolor{DeepRed}{88.95}} & \textbf{\textcolor{DeepRed}{2.03}} \\
\midrule
\multirow{3}{*}{\textbf{MeMOTR} \cite{gao2023memotr}}
 & {\cellcolor{CleanGreen}}Clean & {\cellcolor{CleanGreen}}67.35 & {\cellcolor{CleanGreen}}57.87 & {\cellcolor{CleanGreen}}79.60 & {\cellcolor{CleanGreen}}80.83 & {\cellcolor{CleanGreen}}70.78 & {\cellcolor{CleanGreen}}94.84 & {\cellcolor{CleanGreen}}0.81 \\
 & TMC\_AAI & \textbf{\textcolor{DeepRed}{58.42}} & \textbf{\textcolor{DeepRed}{49.87}} & \textbf{\textcolor{DeepRed}{69.50}} & \textbf{\textcolor{DeepRed}{73.20}} & \textbf{\textcolor{DeepRed}{60.18}} & \textbf{\textcolor{DeepRed}{93.97}} & \textbf{\textcolor{DeepRed}{1.25}} \\
 & TMC\_EAI & \textbf{\textcolor{DeepRed}{58.84}} & \textbf{\textcolor{DeepRed}{50.13}} & \textbf{\textcolor{DeepRed}{70.13}} & \textbf{\textcolor{DeepRed}{73.54}} & \textbf{\textcolor{DeepRed}{60.59}} & \textbf{\textcolor{DeepRed}{94.22}} & \textbf{\textcolor{DeepRed}{1.02}} \\
\midrule
\multirow{3}{*}{\textbf{Samba} \cite{segu2024samba}}
 & {\cellcolor{CleanGreen}}Clean & {\cellcolor{CleanGreen}}62.91 & {\cellcolor{CleanGreen}}50.58 & {\cellcolor{CleanGreen}}79.37 & {\cellcolor{CleanGreen}}73.67 & {\cellcolor{CleanGreen}}60.30 & {\cellcolor{CleanGreen}}95.93 & {\cellcolor{CleanGreen}}1.02 \\
 & TMC\_AAI & \textbf{\textcolor{DeepRed}{48.97}} & \textbf{\textcolor{DeepRed}{37.87}} & \textbf{\textcolor{DeepRed}{64.14}} & \textbf{\textcolor{DeepRed}{59.45}} & \textbf{\textcolor{DeepRed}{43.72}} & \textbf{\textcolor{DeepRed}{93.17}} & \textbf{\textcolor{DeepRed}{1.86}} \\
 & TMC\_EAI & \textbf{\textcolor{DeepRed}{56.44}} & \textbf{\textcolor{DeepRed}{45.53}} & \textbf{\textcolor{DeepRed}{71.01}} & \textbf{\textcolor{DeepRed}{68.61}} & \textbf{\textcolor{DeepRed}{53.97}} & \textbf{\textcolor{DeepRed}{95.47}} & \textbf{\textcolor{DeepRed}{1.27}} \\
\midrule
\multirow{3}{*}{\textbf{CO-MOT} \cite{yan2023bridging}}
 & {\cellcolor{CleanGreen}}Clean & {\cellcolor{CleanGreen}}58.16 & {\cellcolor{CleanGreen}}46.22 & {\cellcolor{CleanGreen}}74.87 & {\cellcolor{CleanGreen}}69.87 & {\cellcolor{CleanGreen}}57.21 & {\cellcolor{CleanGreen}}91.97 & {\cellcolor{CleanGreen}}1.83 \\
 & TMC\_AAI & \textbf{\textcolor{DeepRed}{52.29}} & \textbf{\textcolor{DeepRed}{41.72}} & \textbf{\textcolor{DeepRed}{67.14}} & \textbf{\textcolor{DeepRed}{64.98}} & \textbf{\textcolor{DeepRed}{51.58}} & \textbf{\textcolor{DeepRed}{90.44}} & \textbf{\textcolor{DeepRed}{2.14}} \\
 & TMC\_EAI & \textbf{\textcolor{DeepRed}{51.89}} & \textbf{\textcolor{DeepRed}{41.44}} & \textbf{\textcolor{DeepRed}{66.51}} & \textbf{\textcolor{DeepRed}{64.82}} & \textbf{\textcolor{DeepRed}{51.25}} & \textbf{\textcolor{DeepRed}{91.12}} & \textbf{\textcolor{DeepRed}{2.40}} \\
\bottomrule
\end{tabular}
}
\vspace{-0.2cm}
\end{table}

\begin{table}[t!]
\centering
\scriptsize
\setlength{\tabcolsep}{3.2pt}
\renewcommand{\arraystretch}{0.7}
\caption{Simulated Physical Attack Performance on MOT20.}  
\label{tab:phy_results_mot20}
\vspace{-0.3cm}
\scalebox{.75}{
\begin{tabular}{llccccccc}
\toprule
\rowcolor{HeaderGray}
\multicolumn{9}{c}{
\textbf{MOT20 Dataset – High Density: Avg. $\sim$150 objects/frame.}
}\\[-0.2em]
\midrule
\textbf{Tracker} & \textbf{Attack\_Vector} & \textbf{HOTA}$\downarrow$ & \textbf{DetA}$\downarrow$ & \textbf{AssA}$\downarrow$ & \textbf{IDF1}$\downarrow$ & \textbf{IDR}$\downarrow$ & \textbf{IDP}$\downarrow$ & \textbf{IDSW}$\uparrow$ \\
\midrule
\multirow{3}{*}{\textbf{MOTR} \cite{zeng2022motr}}
 & {\cellcolor{CleanGreen}}Clean & {\cellcolor{CleanGreen}}55.06 & {\cellcolor{CleanGreen}}40.57 & {\cellcolor{CleanGreen}}75.89 & {\cellcolor{CleanGreen}}64.10 & {\cellcolor{CleanGreen}}57.65 & {\cellcolor{CleanGreen}}74.80 & {\cellcolor{CleanGreen}}5.14 \\
 & TMC\_AAI & \textbf{\textcolor{DeepRed}{36.17}} & \textbf{\textcolor{DeepRed}{34.06}} & \textbf{\textcolor{DeepRed}{39.72}} & \textbf{\textcolor{DeepRed}{43.08}} & \textbf{\textcolor{DeepRed}{31.83}} & \textbf{\textcolor{DeepRed}{69.95}} & \textbf{\textcolor{DeepRed}{5.64}} \\
 & TMC\_EAI & \textbf{\textcolor{DeepRed}{41.78}} & \textbf{\textcolor{DeepRed}{35.99}} & \textbf{\textcolor{DeepRed}{51.65}} & \textbf{\textcolor{DeepRed}{48.65}} & \textbf{\textcolor{DeepRed}{37.36}} & \textbf{\textcolor{DeepRed}{71.02}} & \textbf{\textcolor{DeepRed}{5.70}} \\
\midrule
\multirow{3}{*}{\textbf{MOTRv2} \cite{zhang2023motrv2}}
 & {\cellcolor{CleanGreen}}Clean & {\cellcolor{CleanGreen}}59.56 & {\cellcolor{CleanGreen}}44.74 & {\cellcolor{CleanGreen}}80.71 & {\cellcolor{CleanGreen}}69.55 & {\cellcolor{CleanGreen}}54.90 & {\cellcolor{CleanGreen}}96.81 & {\cellcolor{CleanGreen}}0.73 \\
 & TMC\_AAI & \textbf{\textcolor{DeepRed}{54.44}} & \textbf{\textcolor{DeepRed}{41.32}} & \textbf{\textcolor{DeepRed}{72.97}} & \textbf{\textcolor{DeepRed}{66.04}} & \textbf{\textcolor{DeepRed}{50.52}} & \textbf{\textcolor{DeepRed}{97.12}} & \textbf{\textcolor{DeepRed}{0.74}} \\
 & TMC\_EAI & \textbf{\textcolor{DeepRed}{53.23}} & \textbf{\textcolor{DeepRed}{40.25}} & \textbf{\textcolor{DeepRed}{71.65}} & \textbf{\textcolor{DeepRed}{64.48}} & \textbf{\textcolor{DeepRed}{48.93}} & \textbf{\textcolor{DeepRed}{96.47}} & \textbf{\textcolor{DeepRed}{0.84}} \\
\midrule
\multirow{3}{*}{\textbf{MeMOTR} \cite{gao2023memotr}}
 & {\cellcolor{CleanGreen}}Clean & {\cellcolor{CleanGreen}}69.61 & {\cellcolor{CleanGreen}}59.19 & {\cellcolor{CleanGreen}}83.17 & {\cellcolor{CleanGreen}}83.31 & {\cellcolor{CleanGreen}}72.84 & {\cellcolor{CleanGreen}}97.99 & {\cellcolor{CleanGreen}}0.46 \\
 & TMC\_AAI & \textbf{\textcolor{DeepRed}{62.30}} & \textbf{\textcolor{DeepRed}{52.83}} & \textbf{\textcolor{DeepRed}{74.56}} & \textbf{\textcolor{DeepRed}{77.64}} & \textbf{\textcolor{DeepRed}{64.60}} & \textbf{\textcolor{DeepRed}{97.81}} & \textbf{\textcolor{DeepRed}{0.58}} \\
 & TMC\_EAI & \textbf{\textcolor{DeepRed}{61.44}} & \textbf{\textcolor{DeepRed}{51.73}} & \textbf{\textcolor{DeepRed}{74.13}} & \textbf{\textcolor{DeepRed}{76.57}} & \textbf{\textcolor{DeepRed}{63.21}} & \textbf{\textcolor{DeepRed}{97.73}} & \textbf{\textcolor{DeepRed}{0.58}} \\
\midrule
\multirow{3}{*}{\textbf{Samba} \cite{segu2024samba}}
 & {\cellcolor{CleanGreen}}Clean & {\cellcolor{CleanGreen}}62.49 & {\cellcolor{CleanGreen}}47.83 & {\cellcolor{CleanGreen}}83.29 & {\cellcolor{CleanGreen}}72.47 & {\cellcolor{CleanGreen}}58.23 & {\cellcolor{CleanGreen}}98.27 & {\cellcolor{CleanGreen}}0.43 \\
 & TMC\_AAI & \textbf{\textcolor{DeepRed}{54.96}} & \textbf{\textcolor{DeepRed}{41.47}} & \textbf{\textcolor{DeepRed}{74.27}} & \textbf{\textcolor{DeepRed}{65.68}} & \textbf{\textcolor{DeepRed}{50.05}} & \textbf{\textcolor{DeepRed}{97.92}} & \textbf{\textcolor{DeepRed}{0.55}} \\
 & TMC\_EAI & \textbf{\textcolor{DeepRed}{55.46}} & \textbf{\textcolor{DeepRed}{41.98}} & \textbf{\textcolor{DeepRed}{74.82}} & \textbf{\textcolor{DeepRed}{66.42}} & \textbf{\textcolor{DeepRed}{50.88}} & \textbf{\textcolor{DeepRed}{98.25}} & \textbf{\textcolor{DeepRed}{0.53}} \\
\midrule
\multirow{3}{*}{\textbf{CO-MOT} \cite{yan2023bridging}}
 & {\cellcolor{CleanGreen}}Clean & {\cellcolor{CleanGreen}}64.31 & {\cellcolor{CleanGreen}}52.30 & {\cellcolor{CleanGreen}}80.35 & {\cellcolor{CleanGreen}}78.01 & {\cellcolor{CleanGreen}}65.43 & {\cellcolor{CleanGreen}}97.78 & {\cellcolor{CleanGreen}}0.45 \\
 & TMC\_AAI & \textbf{\textcolor{DeepRed}{57.70}} & \textbf{\textcolor{DeepRed}{46.78}} & \textbf{\textcolor{DeepRed}{72.37}} & \textbf{\textcolor{DeepRed}{72.55}} & \textbf{\textcolor{DeepRed}{58.15}} & \textbf{\textcolor{DeepRed}{97.79}} & \textbf{\textcolor{DeepRed}{0.55}} \\
 & TMC\_EAI & \textbf{\textcolor{DeepRed}{56.94}} & \textbf{\textcolor{DeepRed}{46.16}} & \textbf{\textcolor{DeepRed}{71.44}} & \textbf{\textcolor{DeepRed}{71.77}} & \textbf{\textcolor{DeepRed}{57.31}} & \textbf{\textcolor{DeepRed}{97.51}} & \textbf{\textcolor{DeepRed}{0.64}} \\
\bottomrule
\end{tabular}
}
\vspace{-0.4cm}
\end{table}

\noindent \textbf{3. Attack Effectiveness vs. Scene Density.}
Comparing MOT17 and MOT20, we observe that the physical attacks are \textit{more} effective in high-density scenarios. For MOTR, the HOTA degradation from AAI is significantly worse on MOT20 ($-18.9$ HOTA) than on MOT17 ($-13.5$ HOTA). This indicates that the chaos and frequent occlusions of dense scenes amplify the attack's ability to sever temporal links. High-density scenes naturally put more stress on the tracker's temporal memory; our attack exploits this by corrupting the memory at the exact moment it is most needed, leading to an overall performance collapse.

\subsection{Qualitative Analysis of Physical Attacks}
\begin{figure*}[t]
\centering
\includegraphics[width=.83\textwidth]{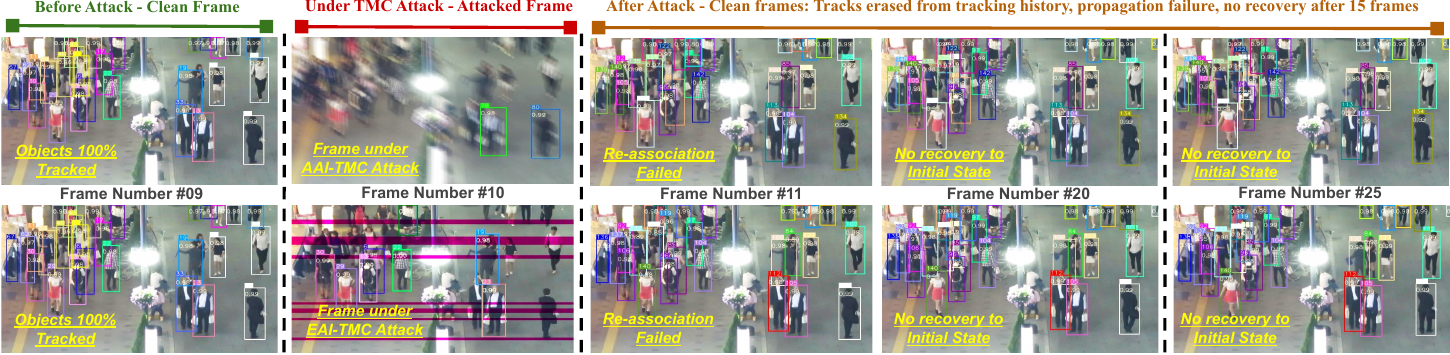}
\vspace{-0.2cm}
\caption{Qualitative comparison of the TMC attack with AAI (upper) and EAI (bottom) attack vectors on the MOT17 benchmark.}
\label{fig:phy_attack_vis}
\vspace{-0.3cm}
\end{figure*}

Fig.~\ref{fig:phy_attack_vis} visualizes the impact of our optimized physical attacks (AAI and EAI) on  MOTR under the $\text{FADE}_{\text{TMC}}$ attack. The visualization confirms that physically realizable artifacts can be optimized to trigger catastrophic temporal failures.
The top row of Fig.~\ref{fig:phy_attack_vis} illustrates the AAI-optimized $\text{FADE}_{\text{TMC}}$ attack. At Frame \#10, the differentiable pipeline generates a calibrated motion blur pattern. While some pedestrians remain visually distinguishable to the human eye, the specific blur kernel successfully severs the tracker's temporal association links. Immediately post-attack (Frame \#11), the tracker fails to re-associate the objects, dropping the original IDs. This forgetting is persistent; even 15 frames later (Frame \#25), the tracker fails to recover the initial identities, treating the pedestrians as entirely new objects or failing to track them altogether.
The bottom row in Fig.~\ref{fig:phy_attack_vis} depicts the EAI-optimized $\text{FADE}_{\text{TMC}}$ attack, which injects optimized color stripe artifacts. Similar to the AAI case, this seemingly low-level noise targets the query propagation logic. The injection at Frame \#10 causes an immediate decoupling of the track histories. The tracker enters a failure state where it cannot bridge the gap created by the single-frame perturbation, leading to long-term identity loss that persists through Frame \#25. 

\subsection{Ablation Study Analysis}
\label{sec:phys_ablation}
\subsubsection{Adversarial Loss Components}
Table~\ref{tab:ablation_loss} presents a detailed ablation study on the adversarial loss components, conducted on the entire MOT17 dataset.

\noindent \textbf{(\textit{i}) Analysis of TQF Loss Components.}
The ablation for $\mathcal{L}_{\text{TQF}}$ demonstrates a synergy between the baseline flooding attack ($\mathcal{L}_{\text{Flood}}$) and the state-manipulation components ($\mathcal{L}_{\text{Siphon}}$, $\mathcal{L}_{\text{Cost}}$).
On its own, $\mathcal{L}_{\text{Flood}}$ is effective, confirming that the zero-sum query budget is a fundamental flaw. For example, on CO-MOT, the naive flood drops HOTA from $58.16$ to $40.25$. However, the full $\mathcal{L}_{\text{TQF}}$ achieves a significantly stronger degradation ($37.26$ HOTA), proving that state manipulation is necessary for maximum impact.
The effectiveness of $\mathcal{L}_{\text{Siphon}}$ and $\mathcal{L}_{\text{Cost}}$ depends on the tracker's underlying logic. On Samba, the $\mathcal{L}_{\text{Cost}}$ component is important. Removing it (`w/o $\mathcal{L}_{\text{Cost}}$`) weakens the attack by $7.6$~HOTA points. Conversely, removing $\mathcal{L}_{\text{Siphon}}$ actually strengthens the attack slightly ($44.96$ HOTA), suggesting Samba's SSM-based memory successfully filters out identity-siphoned queries.
CO-MOT is vulnerable to $\mathcal{L}_{\text{Siphon}}$ but robust to $\mathcal{L}_{\text{Cost}}$.
This reveals that simple query updaters (CO-MOT) are fooled by identity confusion, while memory trackers (Samba) require cost mimicry to be deceived.

\noindent \textbf{(\textit{ii}) Analysis of TMC Loss Components.}
The ablation for $\mathcal{L}_{\text{TMC}}$ validates our \textit{forgetting} attack design. The strategy's power is derived from severing the temporal links.
The effect of $\mathcal{L}_{\text{Decorr}}$ (temporal decorrelation) is the single most critical component across all trackers. Removing it (`w/o $\mathcal{L}_{\text{Decorr}}$`) consistently weakens the attack, causing HOTA to rebound to near-clean baseline levels.
On Samba, removing $\mathcal{L}_{\text{Decorr}}$ causes a massive $13.6$-point jump in HOTA. 
On MeMOTR, it causes a $10.2$-point jump. This proves that severing the temporal link between query states is the main mechanism for inducing tracking failure.
$\mathcal{L}_{\text{Erase}}$ (feature erasure) acts as a complement to decorrelation. Its impact is measurable but less severe than $\mathcal{L}_{\text{Decorr}}$. For instance, on Samba, removing $\mathcal{L}_{\text{Erase}}$ weakens the attack by $4.3$~HOTA points. This confirms that while erasing feature content contributes to the failure, breaking the temporal association link is the prerequisite for catastrophic forgetting.

\vspace{-0.1cm}
\captionsetup{font=small,labelfont=bf}
\begin{table}[h]
\centering
\caption{Ablation Study on FADE Adversarial Losses.}
\label{tab:ablation_loss}
\vspace{-0.2cm}
\setlength{\tabcolsep}{3pt}
\renewcommand{\arraystretch}{.9}
\scalebox{.7}{
\begin{tabular}{lccc@{\hspace{4pt}}ccc@{\hspace{4pt}}ccc}
\toprule
\multirow{2}{*}{\textbf{Adv. Loss}} 
 & \multicolumn{3}{c}{\textbf{CO-MOT}~\cite{yan2023bridging}} 
 & \multicolumn{3}{c}{\textbf{MeMOTR}~\cite{gao2023memotr}} 
 & \multicolumn{3}{c}{\textbf{Samba}~\cite{segu2024samba}} \\
\cmidrule(lr){2-4} \cmidrule(lr){5-7} \cmidrule(lr){8-10}
 & HOTA & IDF1 & IDSW & HOTA & IDF1 & IDSW & HOTA & IDF1 & IDSW \\
\toprule
\rowcolor[rgb]{0.835,0.914,0.835} Clean & 58.16 & 69.87 & 1.83 & 67.35 & 80.83 & 0.81 & 62.91 & 73.67 & 1.02 \\
$\mathcal{L}_{\text{Flood}}$ & 40.25 & 48.51 & 14.45 & 41.04 & 50.99 & 4.47 & 51.50 & 61.85 & 1.91 \\
w/o $\mathcal{L}_{\text{Flood}}$ & 46.54 & 56.51 & 8.89 & 54.27 & 66.13 & 2.40 & 51.86 & 61.72 & 1.58 \\
w/o $\mathcal{L}_{\text{Cost}}$ & 36.44 & 43.99 & 9.77 & 42.53 & 52.67 & 4.10 & 53.14 & 64.02 & 1.98 \\
w/o $\mathcal{L}_{\text{Siphon}}$ & 39.00 & 46.44 & 15.28 & 41.88 & 51.94 & 4.44 & 44.96 & 51.87 & 2.28 \\
\rowcolor[rgb]{0.976,0.976,0.976} \textbf{\textcolor[rgb]{0.565,0.118,0.118}{Full $\mathcal{L}_{\text{TQF}}$}} & \textbf{\textcolor[rgb]{0.565,0.118,0.118}{37.26}} & \textbf{\textcolor[rgb]{0.565,0.118,0.118}{44.84}} & \textbf{\textcolor[rgb]{0.565,0.118,0.118}{9.50}} & \textbf{\textcolor[rgb]{0.565,0.118,0.118}{41.41}} & \textbf{\textcolor[rgb]{0.565,0.118,0.118}{51.41}} & \textbf{\textcolor[rgb]{0.565,0.118,0.118}{4.31}} & \textbf{\textcolor[rgb]{0.565,0.118,0.118}{45.53}} & \textbf{\textcolor[rgb]{0.565,0.118,0.118}{52.71}} & \textbf{\textcolor[rgb]{0.565,0.118,0.118}{2.24}} \\ 
\midrule
w/o $\mathcal{L}_{\text{Decorr}}$ & 53.27 & 64.36 & 3.41 & 51.79 & 58.53 & 4.24 & 61.63 & 73.59 & 1.11 \\
w/o $\mathcal{L}_{\text{Erase}}$ & 41.91 & 50.71 & 12.88 & 41.90 & 51.69 & 4.82 & 52.31 & 62.44 & 2.00 \\
\rowcolor[rgb]{0.976,0.976,0.976} \textbf{\textcolor[rgb]{0.565,0.118,0.118}{Full $\mathcal{L}_{\text{TMC}}$}} & \textbf{\textcolor[rgb]{0.565,0.118,0.118}{41.73}} & \textbf{\textcolor[rgb]{0.565,0.118,0.118}{50.34}} & \textbf{\textcolor[rgb]{0.565,0.118,0.118}{10.94}} & \textbf{\textcolor[rgb]{0.565,0.118,0.118}{41.56}} & \textbf{\textcolor[rgb]{0.565,0.118,0.118}{51.60}} & \textbf{\textcolor[rgb]{0.565,0.118,0.118}{4.63}} & \textbf{\textcolor[rgb]{0.565,0.118,0.118}{48.04}} & \textbf{\textcolor[rgb]{0.565,0.118,0.118}{56.01}} & \textbf{\textcolor[rgb]{0.565,0.118,0.118}{3.63}} \\
\bottomrule
\end{tabular}
}
\vspace{-0.2cm}
\end{table}

\vspace{-0.1cm}

\subsubsection{Physical Attacks Differentiable Optimization}
We analyze the convergence properties of our differentiable simulated physical attack pipelines, AAI and EAI, by evaluating attack strength against the number of PGD iterations ($T$). Results in Table~\ref{tab:ablation_optimization} for MOTR on samples from MOT17 reveal that while both physical simulations are fully differentiable, they have distinct optimization landscapes.

\noindent \textbf{(\textit{i}) AAI exhibits \textit{gradual} convergence.}
AAI attack's effectiveness depends on a sufficient number of optimization steps. At low iteration counts ($T=5$ and $T=10$), the attack is weak, with HOTA scores ($49.51$ and $49.62$) near the baseline. A significant performance \textit{drop} occurs between $T=10$ and $T=50$, where HOTA decreases by $6.6$~points. The attack continues to strengthen until $T=100$ ($39.85$ HOTA) before fully converging, as shown by the negligible change at $T=150$. This demonstrates that while the AAI simulation is highly differentiable, its parameter space is complex, requiring at least $T=100$ PGD iterations.

\noindent \textbf{(\textit{ii}) EAI demonstrates rapid convergence.}
The EAI attack model finds a powerful adversarial example almost immediately. The attack is already devastating at $T=5$ ($43.42$ HOTA), achieving a performance drop that the AAI attack only surpasses after $T=50$. Increasing the iteration count for EAI yields only marginal gains; the HOTA at T=150 ($41.92$) is only $1.5$~points lower than at $T=5$. This suggests the EAI attack vector has a \textit{flatter} or more accessible optimization landscape, where PGD converges quickly.

\vspace{-0.1cm}
\captionsetup{font=small,labelfont=bf}
\begin{table}[h]
\centering
\caption{Simulated Physical Attack Optimization Convergence. }
\label{tab:ablation_optimization}
\vspace{-0.2cm}
\setlength{\tabcolsep}{5pt}
\renewcommand{\arraystretch}{.95}
\scalebox{.7}{
\begin{tabular}{cccccccc} 
\toprule
\multirow{2}{*}{\begin{tabular}[c]{@{}c@{}}\textbf{Attack}\\\textbf{Vector}\end{tabular}} & \multirow{2}{*}{\textbf{Metrics }} & \multicolumn{6}{c}{\textbf{Number of PGD Iterations }} \\ 
\cline{3-8}
 &  & T=0 & T=5 & T=10 & T=50 & T=100 & \textbf{T=150} \\ 
\hline
\multirow{3}{*}{\begin{tabular}[c]{@{}c@{}}$\text{FADE}_{\text{TMC}}$\\AAI\end{tabular}} & HOTA & 52.03 & 49.51 & 49.62 & 43.02 & 39.85 & \textbf{\textcolor[rgb]{0.565,0.118,0.118}{39.81}} \\
 & IDF1 & 60.35 & 57.34 & 58.02 & 48.89 & 44.95 & \textbf{\textcolor[rgb]{0.565,0.118,0.118}{44.86}} \\
 & IDSW & 4.39 & 6.05 & 5.47 & 6.26 & 6.85 & \textbf{\textcolor[rgb]{0.565,0.118,0.118}{7.05}} \\ 
\hline
\multirow{3}{*}{\begin{tabular}[c]{@{}c@{}}$\text{FADE}_{\text{TMC}}$\\EAI\end{tabular}} & HOTA & 52.03 & 43.42 & 42.39 & 42.38 & 41.06 & \textbf{\textcolor[rgb]{0.565,0.118,0.118}{41.92}} \\
 & IDF1 & 60.35 & 49.23 & 48.01 & 47.62 & 46.40 & \textbf{\textcolor[rgb]{0.565,0.118,0.118}{47.47}} \\
 & IDSW & 4.39 & 6.24 & 6.99 & 6.11 & 6.40 & \textbf{\textcolor[rgb]{0.565,0.118,0.118}{7.26}} \\
\bottomrule
\end{tabular}
}
\vspace{-0.4cm}
\end{table}
\vspace{-0.1cm}

\section{Limitation}
FADE presents the first robustness study on TBP trackers, formalizing the threat models and failure physics required for future physical security benchmarks. However, a limitation of this work is the lack of closed-loop validation on real hardware; while our simulations are grounded in physically calibrated sensor profiles, real-world dynamics remain unexplored. Additionally, while we observe that adversarial attacks exhibit partial transferability among models sharing a common architectural lineage, achieving robust cross-architecture black-box transferability remains an open challenge. In the Appendix, we provide an empirical verification of these transferability properties and a defense robustness analysis, alongside suggestions for improvements.
\section{Conclusion}
We present FADE, the first adversarial framework designed to target the architectural constraints of emerging TBP trackers: the zero-sum query budget and recurrent state memory. We propose two attack strategies: Temporal Query Flooding (TQF) and Temporal Memory Corruption (TMC). Extensive evaluations demonstrate that FADE outperforms baseline attacks. Furthermore, we introduce a differentiable pipeline to map these attacks to physically grounded sensor-spoofing vectors. Our findings reveal that the TBP paradigm introduces structural vulnerabilities that must be secured to ensure reliable deployment.
{
    \small
    \bibliographystyle{ieeenat_fullname}
    \bibliography{main}

@String(ICME = {Int. Conf. Multimedia and Expo})

@String(ICIP = {IEEE Int. Conf. Image Process.})

@String(ICLR = {Int. Conf. Learn. Represent.})

@String(AAAI = {AAAI})

@String(ICME  =	{ICME})

@String(ICIP  = {ICIP})

@String(ICLR  = {ICLR})

@inproceedings{zeng2022motr,
  title={Motr: End-to-end multiple-object tracking with transformer},
  author={Zeng, Fangao and Dong, Bin and Zhang, Yuang and Wang, Tiancai and Zhang, Xiangyu and Wei, Yichen},
  booktitle={European conference on computer vision},
  pages={659--675},
  year={2022},
  organization={Springer}
}

@inproceedings{zhang2023motrv2,
  title={Motrv2: Bootstrapping end-to-end multi-object tracking by pretrained object detectors},
  author={Zhang, Yuang and Wang, Tiancai and Zhang, Xiangyu},
  booktitle={Proceedings of the IEEE/CVF conference on computer vision and pattern recognition},
  pages={22056--22065},
  year={2023}
}

@inproceedings{gao2023memotr,
  title={MeMOTR: Long-term memory-augmented transformer for multi-object tracking},
  author={Gao, Ruopeng and Wang, Limin},
  booktitle={Proceedings of the IEEE/CVF International Conference on Computer Vision},
  pages={9901--9910},
  year={2023}
}

@inproceedings{segu2024samba,
  title={Samba: Synchronized Set-of-Sequences Modeling for Multiple Object Tracking},
  author={Segu, Mattia and Piccinelli, Luigi and Li, Siyuan and Yang, Yung-Hsu and Schiele, Bernt and Van Gool, Luc},
  booktitle={The Thirteenth International Conference on Learning Representations},
  pages={30057--30070},
  year={2025},
  organization={ICLR}
}

@inproceedings{yan2023bridging,
  title={CO-MOT: Boosting end-to-end transformer-based multi-object tracking via coopetition label assignment and shadow sets},
  author={Luo, Weixin and Zhong, Yujie and Gan, Yiyang and Ma, Lin and others},
  booktitle={The Thirteenth International Conference on Learning Representations},
  year={2025}
}

@article{sun2020transtrack,
  title={Transtrack: Multiple object tracking with transformer},
  author={Sun, Peize and Cao, Jinkun and Jiang, Yi and Zhang, Rufeng and Xie, Enze and Yuan, Zehuan and Wang, Changhu and Luo, Ping},
  journal={arXiv preprint arXiv:2012.15460},
  year={2020}
}

@article{goodfellow2014explaining,
  title={Explaining and harnessing adversarial examples},
  author={Goodfellow, Ian J and Shlens, Jonathon and Szegedy, Christian},
  journal={arXiv preprint arXiv:1412.6572},
  year={2014}
}

@inproceedings{ding2021towards,
  title={Towards universal physical attacks on single object tracking},
  author={Ding, Li and Wang, Yongwei and Yuan, Kaiwen and Jiang, Minyang and Wang, Ping and Huang, Hua and Wang, Z Jane},
  booktitle={Proceedings of the AAAI Conference on Artificial Intelligence},
  volume={35},
  number={2},
  pages={1236--1245},
  year={2021}
}

@inproceedings{biggio2018wild,
  title={Wild patterns: Ten years after the rise of adversarial machine learning},
  author={Biggio, Battista and Roli, Fabio},
  booktitle={Proceedings of the 2018 ACM SIGSAC conference on computer and communications security},
  pages={2154--2156},
  year={2018}
}

@inproceedings{bewley2016simple,
  title={Simple online and realtime tracking},
  author={Bewley, Alex and Ge, Zongyuan and Ott, Lionel and Ramos, Fabio and Upcroft, Ben},
  booktitle={2016 IEEE international conference on image processing (ICIP)},
  pages={3464--3468},
  year={2016},
  organization={Ieee}
}

@inproceedings{braso2020learning,
  title={Learning a neural solver for multiple object tracking},
  author={Bras{\'o}, Guillem and Leal-Taix{\'e}, Laura},
  booktitle={Proceedings of the IEEE/CVF conference on computer vision and pattern recognition},
  pages={6247--6257},
  year={2020}
}

@inproceedings{sahbani2016kalman,
  title={Kalman filter and iterative-hungarian algorithm implementation for low complexity point tracking as part of fast multiple object tracking system},
  author={Sahbani, Bima and Adiprawita, Widyawardana},
  booktitle={2016 6th international conference on system engineering and technology (ICSET)},
  pages={109--115},
  year={2016},
  organization={IEEE}
}

@inproceedings{zhang2022bytetrack,
  title={Bytetrack: Multi-object tracking by associating every detection box},
  author={Zhang, Yifu and Sun, Peize and Jiang, Yi and Yu, Dongdong and Weng, Fucheng and Yuan, Zehuan and Luo, Ping and Liu, Wenyu and Wang, Xinggang},
  booktitle={European conference on computer vision},
  pages={1--21},
  year={2022},
  organization={Springer}
}

@inproceedings{meinhardt2022trackformer,
  title={Trackformer: Multi-object tracking with transformers},
  author={Meinhardt, Tim and Kirillov, Alexander and Leal-Taixe, Laura and Feichtenhofer, Christoph},
  booktitle={Proceedings of the IEEE/CVF conference on computer vision and pattern recognition},
  pages={8844--8854},
  year={2022}
}

@article{sun2019deep,
  title={Deep affinity network for multiple object tracking},
  author={Sun, ShiJie and Akhtar, Naveed and Song, HuanSheng and Mian, Ajmal and Shah, Mubarak},
  journal={IEEE transactions on pattern analysis and machine intelligence},
  volume={43},
  number={1},
  pages={104--119},
  year={2019},
  publisher={IEEE}
}

@article{dendorfer2020mot20,
  title={Mot20: A benchmark for multi object tracking in crowded scenes},
  author={Dendorfer, Patrick and Rezatofighi, Hamid and Milan, Anton and Shi, Javen and Cremers, Daniel and Reid, Ian and Roth, Stefan and Schindler, Konrad and Leal-Taix{\'e}, Laura},
  journal={arXiv preprint arXiv:2003.09003},
  year={2020}
}

@article{luiten2021hota,
  title={A higher order metric for evaluating multi-object tracking., 2021, 129},
  author={Luiten, J and Osep, A and Dendorfer, P and Torr, P and Geiger, A and Leal-Taix{\'e}, L and Hota, B Leibe},
  journal={DOI: https://doi. org/10.1007/s11263-020-01375-2. PMID: https://www. ncbi. nlm. nih. gov/pubmed/33642696},
  pages={548--578}
}

@inproceedings{ristani2016performance,
  title={Performance measures and a data set for multi-target, multi-camera tracking},
  author={Ristani, Ergys and Solera, Francesco and Zou, Roger and Cucchiara, Rita and Tomasi, Carlo},
  booktitle={European conference on computer vision},
  pages={17--35},
  year={2016},
  organization={Springer}
}

@article{ge2021yolox,
  title={Yolox: Exceeding yolo series in 2021},
  author={Ge, Zheng and Liu, Songtao and Wang, Feng and Li, Zeming and Sun, Jian},
  journal={arXiv preprint arXiv:2107.08430},
  year={2021}
}

@article{ravindran2020multi,
  title={Multi-object detection and tracking, based on DNN, for autonomous vehicles: A review},
  author={Ravindran, Ratheesh and Santora, Michael J and Jamali, Mohsin M},
  journal={IEEE Sensors Journal},
  volume={21},
  number={5},
  pages={5668--5677},
  year={2020},
  publisher={IEEE}
}

@inproceedings{song2024sftrack,
  title={SFTrack: A robust scale and motion adaptive algorithm for tracking small and fast moving objects},
  author={Song, Inpyo and Lee, Jangwon},
  booktitle={2024 IEEE/RSJ International Conference on Intelligent Robots and Systems (IROS)},
  pages={10870--10877},
  year={2024},
  organization={IEEE}
}

@inproceedings{wang2024smiletrack,
  title={Smiletrack: Similarity learning for occlusion-aware multiple object tracking},
  author={Wang, Yu-Hsiang and Hsieh, Jun-Wei and Chen, Ping-Yang and Chang, Ming-Ching and So, Hung-Hin and Li, Xin},
  booktitle={Proceedings of the AAAI conference on artificial intelligence},
  volume={38},
  number={6},
  pages={5740--5748},
  year={2024}
}

@article{welch1995introduction,
  title={An introduction to the Kalman filter},
  author={Welch, Greg and Bishop, Gary and others},
  year={1995},
  publisher={Chapel Hill, NC, USA}
}

@inproceedings{zhu2023tpatch,
  title={$\{$TPatch$\}$: A triggered physical adversarial patch},
  author={Zhu, Wenjun and Ji, Xiaoyu and Cheng, Yushi and Zhang, Shibo and Xu, Wenyuan},
  booktitle={32nd USENIX Security Symposium (USENIX Security 23)},
  pages={661--678},
  year={2023}
}

@inproceedings{ji2021poltergeist,
  title={Poltergeist: Acoustic adversarial machine learning against cameras and computer vision},
  author={Ji, Xiaoyu and Cheng, Yushi and Zhang, Yuepeng and Wang, Kai and Yan, Chen and Xu, Wenyuan and Fu, Kevin},
  booktitle={2021 IEEE symposium on security and privacy (SP)},
  pages={160--175},
  year={2021},
  organization={IEEE}
}

@inproceedings{jiang2023glitchhiker,
  title={$\{$GlitchHiker$\}$: Uncovering vulnerabilities of image signal transmission with $\{$IEMI$\}$},
  author={Jiang, Qinhong and Ji, Xiaoyu and Yan, Chen and Xie, Zhixin and Lou, Haina and Xu, Wenyuan},
  booktitle={32nd USENIX Security Symposium (USENIX Security 23)},
  pages={7249--7266},
  year={2023}
}

@inproceedings{zhang2024understanding,
  title={Understanding Impacts of Electromagnetic Signal Injection Attacks on Object Detection},
  author={Zhang, Youqian and Yang, Chunxi and Fu, Eugene Y and Jiang, Qinhong and Yan, Chen and Chau, Sze-Yiu and Ngai, Grace and Leong, Hong-Va and Luo, Xiapu and Xu, Wenyuan},
  booktitle={2024 IEEE International Conference on Multimedia and Expo (ICME)},
  pages={1--6},
  year={2024},
  organization={IEEE}
}

@inproceedings{ren2025ghostshot,
  title={GhostShot: Manipulating the Image of CCD Cameras with Electromagnetic Interference.},
  author={Ren, Yanze and Jiang, Qinhong and Yan, Chen and Ji, Xiaoyu and Xu, Wenyuan},
  booktitle={NDSS},
  pages={24--28},
  year={2025}
}

@article{liu2025magshadow,
  title={MagShadow: physical adversarial example attacks via electromagnetic injection},
  author={Liu, Ziwei and Lin, Feng and Ba, Zhongjie and Lu, Li and Ren, Kui},
  journal={IEEE Transactions on Dependable and Secure Computing},
  volume={22},
  number={4},
  pages={3307--3323},
  year={2025},
  publisher={IEEE}
}

@inproceedings{liao2025your,
  title={Is your autonomous vehicle safe? Understanding the threat of electromagnetic signal injection attacks on traffic scene perception},
  author={Liao, Wenhao and Yan, Sineng and Zhang, Youqian and Zhai, Xinwei and Wang, Yuanyuan and Fu, Eugene},
  booktitle={Proceedings of the AAAI Conference on Artificial Intelligence},
  volume={39},
  number={26},
  pages={27464--27472},
  year={2025}
}

@article{wang2022survey,
  title={A survey on physical adversarial attack in computer vision},
  author={Wang, Donghua and Yao, Wen and Jiang, Tingsong and Tang, Guijian and Chen, Xiaoqian},
  journal={arXiv preprint arXiv:2209.14262},
  year={2022}
}

@inproceedings{bertinetto2016fully,
  title={Fully-convolutional siamese networks for object tracking},
  author={Bertinetto, Luca and Valmadre, Jack and Henriques, Joao F and Vedaldi, Andrea and Torr, Philip HS},
  booktitle={European conference on computer vision},
  pages={850--865},
  year={2016},
  organization={Springer}
}

@inproceedings{shin2024banktweak,
  title={BankTweak: adversarial attack against multi-object trackers by manipulating feature banks},
  author={Shin, Woojin and Kang, Donghwa and Choi, Daejin and Kang, Brent Byunghoon and Lee, Jinkyu and Baek, Hyeongboo},
  booktitle={Proceedings of the Thirty-Fourth International Joint Conference on Artificial Intelligence},
  pages={1847--1855},
  year={2025}
}

@inproceedings{zhou2023f,
  title={F\&F attack: Adversarial attack against multiple object trackers by inducing false negatives and false positives},
  author={Zhou, Tao and Ye, Qi and Luo, Wenhan and Zhang, Kaihao and Shi, Zhiguo and Chen, Jiming},
  booktitle={Proceedings of the IEEE/CVF International Conference on Computer Vision},
  pages={4573--4583},
  year={2023}
}

@article{wang2021daedalus,
  title={Daedalus: Breaking nonmaximum suppression in object detection via adversarial examples},
  author={Wang, Derui and Li, Chaoran and Wen, Sheng and Han, Qing-Long and Nepal, Surya and Zhang, Xiangyu and Xiang, Yang},
  journal={IEEE Transactions on Cybernetics},
  volume={52},
  number={8},
  pages={7427--7440},
  year={2021},
  publisher={IEEE}
}

@inproceedings{jia2020fooling,
  title={Fooling Detection Alone is Not Enough: First Adversarial Attack against Multiple Object Tracking},
  author={Jia, Yunhan and Lu, Yantao and Shen, Junjie and Chen, Qi A and Zhong, Zhenyu and Wei, Tao},
  booktitle={International Conference on Learning Representations (ICLR)},
  year={2020}
}

@inproceedings{long2024papmot,
  title={Papmot: Exploring adversarial patch attack against multiple object tracking},
  author={Long, Jiahuan and Jiang, Tingsong and Yao, Wen and Jia, Shuai and Zhang, Weijia and Zhou, Weien and Ma, Chao and Chen, Xiaoqian},
  booktitle={European Conference on Computer Vision},
  pages={128--144},
  year={2024},
  organization={Springer}
}

@article{lu2017adversarial,
  title={Adversarial examples that fool detectors},
  author={Lu, Jiajun and Sibai, Hussein and Fabry, Evan},
  journal={arXiv preprint arXiv:1712.02494},
  year={2017}
}

@article{pang2024blinding,
  title={Blinding and blurring the multi-object tracker with adversarial perturbations},
  author={Pang, Haibo and Ma, Rongqi and Su, Jie and Liu, Chengming and Gao, Yufei and Jin, Qun},
  journal={Neural Networks},
  volume={176},
  pages={106331},
  year={2024},
  publisher={Elsevier}
}

@inproceedings{yao2024understanding,
  title={Understanding Model Ensemble in Transferable Adversarial Attack},
  author={Yao, Wei and Zhang, Zeliang and Tang, Huayi and Liu, Yong},
  booktitle={International Conference on Machine Learning},
  pages={71798--71833},
  year={2025},
  organization={PMLR}
}

@inproceedings{tang2024ensemble,
  title={Ensemble diversity facilitates adversarial transferability},
  author={Tang, Bowen and Wang, Zheng and Bin, Yi and Dou, Qi and Yang, Yang and Shen, Heng Tao},
  booktitle={Proceedings of the IEEE/CVF conference on computer vision and pattern recognition},
  pages={24377--24386},
  year={2024}
}
}

\maketitlesupplementary
\setcounter{page}{1}

\startcontents

\begin{strip}
    \vspace{-3em} 
    \centering
    \section*{Supplementary Material Overview}
    \vspace{-1em} 
    
    \begin{minipage}{.95\textwidth} 
        \printcontents{}{1}{\textbf{Contents}\vskip3pt\hrule\vskip5pt}
        \vskip3pt\hrule\vskip5pt
    \end{minipage}
    \vspace{1.5em} 
\end{strip}

\section{  Theories: Modeling TBP Vulnerabilities}
\label{sec:theory}
FADE is a \textit{white-box} attack framework specifically designed to exploit two core architectural assumptions in TBP trackers. We first model these assumptions mathematically to formalize their underlying vulnerabilities (Vuln).

\subsection{  Vuln 1: The Track Query Budget}
\label{sec:theory:budget}
At any frame $t$, a TBP tracker is given a candidate set of $M+N$ queries, $\mathcal{Q}_t^{in} = \mathcal{T}_{t-1} \cup \mathcal{Q}_{\text{det}}$. However, the tracker has a fixed, finite query budget $B$ for the \textit{next} frame's track queries, $\mathcal{T}_t$, where $|\mathcal{T}_t| \leq B$ and typically $M+N > B$.

The tracker must implicitly solve a constrained resource allocation problem: it must select the $B$ \textit{best} queries to propagate. We can model this as an optimization problem where the tracker seeks to maximize the total \textit{utility} (e.g., confidence or objectness score) of the propagated set:
\begin{equation}
\max_{\mathcal{T}_t \subseteq \mathcal{Q}_t^{out}} \sum_{q_i \in \mathcal{T}_t} s(q_i) \quad \text{subject to} \quad |\mathcal{T}_t| \leq B    
\label{eq:allocation}
\end{equation}

\noindent where $s(q_i)$ is the confidence score for the output query $q_i$.

\vspace{0.3cm}

\noindent \textbf{Attack Rationale ($\mathcal{L}_{\text{TQF}}$).}
The Temporal Query Flooding (TQF) attack exploits this zero-sum constraint. We introduce a set of $K$ adversarial queries $\mathcal{T}_{\text{adv}}$ and use an optimization objective $\mathcal{L}_{\text{TQF}}$ to force $s(q_j) \to 1$ for all $q_j \in \mathcal{T}_{\text{adv}}$. If the attacker can make its $K$ adversarial queries appear to have a higher utility $s(q_j)$ than $K$ legitimate track queries $q_i \in \mathcal{T}_t$, the tracker's optimization logic will be forced to discard the legitimate tracks to make room for the adversarial ones, hence the \textit{flooding} that leads to query starvation.

\subsection{  Vuln 2: The Auto-Regressive Query Update}
\label{sec:theory:dynamical}
The core of TBP tracking is its auto-regressive nature. We can model the temporal propagation of a single track's identity as a \textit{discrete-time dynamical system}.

Let $\mathbf{h}_i^t \in \mathbb{R}^D$ be the hidden state (the $D$-dimensional query embedding) for object $i$ at frame $t$. The Query Updater $\mathcal{F}$ acts as the state transition function, where $q_i^{out}$ is the output query from the decoder and $X_t$ is the image feature map:
\begin{equation}
(\mathbf{h}_i^t, \mathcal{M}_t) = \mathcal{F}(\mathbf{h}_i^{t-1}, \mathcal{M}_{t-1}, q_i^{out}, X_t)    
\label{eq:dyamic_system}
\end{equation}

A \textit{stable} track requires that the identity of the object is preserved across time steps. Mathematically, this implies that the hidden state $\mathbf{h}_i^t$ must remain \textit{close} to its predecessor $\mathbf{h}_i^{t-1}$ in the embedding manifold, representing the same object. We formalize the stability condition as follows:
\begin{equation}
\textit{Stability Condition:} \quad \text{sim}(\mathbf{h}_i^t, \mathbf{h}_i^{t-1}) \geq \tau_{\text{sim}}
\label{eq:dynamic_constraint}
\end{equation}
\noindent where $\text{sim}(\cdot, \cdot)$ is a similarity metric (e.g., cosine similarity) and $\tau_{\text{sim}}$ is a high threshold (e.g., $\approx 1.0$).

\vspace{0.3cm}

\noindent \textbf{Attack Rationale ($\mathcal{L}_{\text{TMC}}$).}
The Temporal Memory Corruption (TMC) attack is a direct \textit{destabilization attack} on the dynamical system of the auto-regressive query update logic. Instead of adding new queries, it seeks to violate the stability condition for existing, legitimate tracks.
\begin{enumerate}
    \item \textbf{$\mathcal{L}_{\text{Decorr}}$} is designed to \textit{directly minimize} $\text{sim}(\mathbf{h}_i^t, \mathbf{h}_i^{t-1})$, forcing a temporal disconnect.
    \item \textbf{$\mathcal{L}_{\text{Erase}}$} is a more powerful destructive attack, seeking to send the state to a null point, $\mathbf{h}_i^t \to \mathbf{0}$, from which no identity can be recovered.
\end{enumerate}

\section{Derivation of Temporal Query Flooding}
\label{sec:tqf}
$\mathcal{L}_{\text{TQF}}$ is a composite loss designed to solve the query budget exhaustion problem from Sec.~\ref{sec:theory:budget}.
$$
\mathcal{L}_{\text{TQF}} = \lambda_{\text{Flood}} \mathcal{L}_{\text{Flood}} + \lambda_c \mathcal{L}_{\text{Cost}} + \lambda_s \mathcal{L}_{\text{Siphon}}
$$

\subsection{  Component 1: Query Budget Exhaustion}
\label{sec:tqf:flood}
\textbf{Objective.} Force the tracker to assign maximal confidence to adversarial queries.
$$
\mathcal{L}_{\text{Flood}} = \mathbb{E}_{q_i \in \mathcal{T}_{\text{adv}}} \left[ (1.0 - \max_c \sigma(z_i)_c)^2 \right]
$$
\textbf{Justification.}
Let $\hat{c}_i = \max_c \sigma(z_i)_c$ be the tracker's predicted confidence score, $s(q_i)$, for query $q_i$. This loss function formulates the attack as a regression problem, minimizing the L2 (squared Euclidean) distance between the current confidence $\hat{c}_i$ and a target value of $1.0$.

While a cross-entropy loss could be used, the squared L2 loss is a \textit{hard-margin} objective that provides a strong gradient for any $\hat{c}_i < 1.0$. It aggressively punishes all adversarial queries that are not-perfectly-confident, ensuring the entire set $\mathcal{T}_{\text{adv}}$ presents itself as a block of high-utility \textit{distractors}, thereby maximizing its ability to displace legitimate tracks in the constrained allocation problem (Eq. \ref{eq:allocation}).

\subsection{  Component 2: Matching Deception}
\label{sec:tqf:cost}
\textbf{Objective.} Deceive the bipartite matching algorithm (e.g., hangarian matching) by making adversarial queries appear to be a low-cost match for real objects.
$$
\mathcal{L}_{\text{Cost}} = - \mathbb{E}_{g_j \in \mathcal{G}_t} \left[ \log \sum_{q_i \in \mathcal{T}_{\text{adv}}} \exp(-C(q_i, g_j)) \right]
$$
\textbf{Justification.}
This formulation is a direct application of the LogSumExp (LSE) method to create a differentiable approximation of the $\min$ function.
First, recall the LSE function, which is a smooth approximation of the $\max$ function:
$$
\text{LSE}(\mathbf{x}) = \log \sum_i \exp(x_i) \approx \max(\mathbf{x})
$$
We can derive a smooth $\min$ function by noting that $\min(\mathbf{x}) = - \max(-\mathbf{x})$.
\begin{align}
\min(\mathbf{x}) &= - \max(-\mathbf{x}) \\
&\approx - \text{LSE}(-\mathbf{x}) \\
&= - \log \sum_i \exp(-x_i)
\end{align}
Now, consider our $\mathcal{L}_{\text{Cost}}$ for a single ground-truth object $g_j$. Let $\mathbf{C}_j = \{C(q_i, g_j) | q_i \in \mathcal{T}_{\text{adv}}\}$ be the vector of matching costs between $g_j$ and all adversarial queries.
$$
\mathcal{L}_{\text{Cost}}(g_j) = - \log \sum_{q_i \in \mathcal{T}_{\text{adv}}} \exp(-C(q_i, g_j)) = - \text{LSE}(-\mathbf{C}_j)
$$
Therefore, $\mathcal{L}_{\text{Cost}}(g_j)$ is a differentiable approximation of $\min(\mathbf{C}_j)$.
$$
\mathcal{L}_{\text{Cost}}(g_j) \approx \min_{q_i \in \mathcal{T}_{\text{adv}}} C(q_i, g_j)
$$
By minimizing $\mathcal{L}_{\text{Cost}}$, our attack is \textit{minimizing the minimum matching cost}. This creates an adversarial query $q_i$ that appears to be an extremely reliable (low-cost) match for a real object $g_j$, fooling the tracker's association logic.

\subsection{  Component 3: Identity Siphoning}
\label{sec:tqf:siphon}
\textbf{Objective.} Make new adversarial queries appear to be the continuation of old, reliable, legitimate tracks.
$$
\mathcal{L}_{\text{Siphon}} = - \mathbb{E}_{h_i^t \in \mathcal{H}_{\text{adv}}^t, h_j^{t-1} \in \mathcal{H}_{\text{anchor}}^{t-1}} \left[ \text{cosine}(\mathbf{h}_i^t, \mathbf{h}_j^{t-1}) \right]
$$
\textbf{Justification.}
This loss is equivalent to \textit{maximizing} the cosine similarity between a new adversarial query state $\mathbf{h}_i^t$ and a historical, legitimate track state $\mathbf{h}_j^{t-1}$:
$$
\max_{\bm{\omega}} \mathbb{E} \left[ \frac{\mathbf{h}_i^t \cdot \mathbf{h}_j^{t-1}}{\|\mathbf{h}_i^t\|_2 \|\mathbf{h}_j^{t-1}\|_2} \right]
$$
\noindent \textbf{Geometric Interpretation (Cosine vs. L2 Distance).}
Here we justify the choice of cosine similarity over L2 distance. In TBP trackers, the \textit{identity} of an object is encoded in the \textit{direction} of its $D$-dimensional query vector $\mathbf{h}$. The \textit{magnitude} $\|\mathbf{h}\|$ often encodes non-identity features like confidence or visibility.
\begin{itemize}
    \item An L2 loss, $\mathcal{L}_{\text{L2}} = \|\mathbf{h}_i^t - \mathbf{h}_j^{t-1}\|_2^2$, would penalize differences in both direction and magnitude.
    \item Our chosen cosine loss, $\mathcal{L}_{\text{Siphon}}$, \textit{only} penalizes differences in direction.
\end{itemize}
By maximizing the cosine similarity, we are minimizing the angle $\theta$ between the two vectors, forcing $\mathbf{h}_i^t$ to align geometrically with $\mathbf{h}_j^{t-1}$ in the embedding manifold. The Query Updater $\mathcal{F}$ (from Sec.~\ref{sec:theory:dynamical}) interprets this directional alignment as a re-emergence of the \textit{same object}, thus \textit{siphoning} the legitimate track's identity.

\section{Derivation of Temporal Memory Corruption}
\label{sec:tmc}
$\mathcal{L}_{\text{TMC}}$ is the \textit{destabilization} attack from Sec.~\ref{sec:theory:dynamical}.
$$
\mathcal{L}_{\text{TMC}} = \lambda_{\text{Decorr}} \mathcal{L}_{\text{Decorr}} + \lambda_{\text{Erase}} \mathcal{L}_{\text{Erase}}
$$

\subsection{  Component 1: Temporal System Weakening}
\label{sec:tmc:decorr}
\textbf{Objective.} Sever the temporal link by violating the track stability condition in Eq. \ref{eq:dynamic_constraint}.
$$
\mathcal{L}_{\text{Decorr}} = \mathbb{E}_{i \in \text{matched}} \left[\text{cosine}(\mathbf{h}_i^t, \mathbf{h}_i^{t-1})\right]
$$
\textbf{Justification.}
As established in our dynamical system model (Eq. \ref{eq:dyamic_system}), track stability requires $\text{cosine}(\mathbf{h}_i^t, \mathbf{h}_i^{t-1}) \approx 1$.
The loss $\mathcal{L}_{\text{Decorr}}$ is a direct objective to \textit{minimize} this value. In non-negative feature spaces (common after ReLU nonlinearities), the effective minimum is 0.
$$
\min_{\bm{\omega}} \mathbb{E}_{i \in \text{matched}} \left[ \frac{\mathbf{h}_i^t \cdot \mathbf{h}_i^{t-1}}{\|\mathbf{a}\|_2 \|\mathbf{b}\|_2} \right] \quad \implies \quad \mathbf{h}_i^t \cdot \mathbf{h}_i^{t-1} \to 0
$$
Minimizing this loss forces the current state vector $\mathbf{h}_i^t$ to become \textit{orthogonal} to its predecessor $\mathbf{h}_i^{t-1}$ (i.e., $\mathbf{h}_i^t \perp \mathbf{h}_i^{t-1}$).
Geometrically, two orthogonal vectors share no directional information. This represents a break in the state's temporal linkage, guaranteeing a violation of the stability condition and causing the Query Updater to fail re-association.

\subsection{  Component 2: State Destruction at the Origin}
\label{sec:tmc:erase}
\textbf{Objective.} Destroy the track's identity by collapsing its feature embedding.
$$
\mathcal{L}_{\text{Erase}} = \mathbb{E}_{h_i \in \mathcal{H}_{\text{matched}}^t} \left[ \Vert \mathbf{h}_i \Vert_2^2 \right]
$$
\textbf{Justification.}
This loss minimizes the squared L2-norm (the \textit{energy}) of the feature vector $\mathbf{h}_i^t$.
\begin{itemize}
    \item \textbf{Information-Theoretic View:} The origin vector $\mathbf{0}$ is an information-theoretic \textit{null} state. It carries no information about identity, appearance, or motion. Minimizing this loss is an \textit{erasure} attack that drives the information content of the feature vector to zero.
    \item \textbf{Geometric View:} The unique global minimum of this convex loss function is $\mathbf{h}_i^t = \mathbf{0}$, the origin of the $D$-dimensional embedding space.
\end{itemize}
Forcing the state to collapse to the origin is mathematically destructive for the tracker, as all similarity metrics fail:
\begin{enumerate}
    \item \textbf{Dot Product Similarity:} $\text{dot}(\mathbf{h}_i^t, \mathbf{v}) = \text{dot}(\mathbf{0}, \mathbf{v}) = 0$ for any vector $\mathbf{v}$.
    \item \textbf{Cosine Similarity:} $\text{cosine}(\mathbf{h}_i^t, \mathbf{v}) = \frac{\mathbf{0} \cdot \mathbf{v}}{\|\mathbf{0}\|_2 \|\mathbf{v}\|_2}$, which is \textit{undefined} due to division by zero.
\end{enumerate}
This attack does not merely \textit{confuse} the tracker; it breaks the underlying mathematics of the similarity metrics it relies upon. It erases the track state, rendering future re-association impossible.

\vspace{-0.2cm}
\begin{figure}[ht]
\centering
\includegraphics[width=0.44\textwidth]{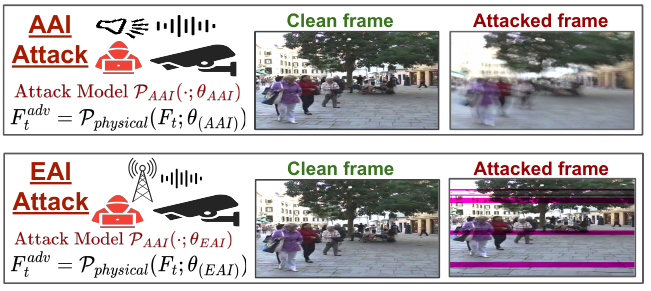}
\vspace{-0.3cm}
\caption{Physical Adversarial Attacks Scenarios and Models. 
}
\label{fig:attack_models}
\vspace{-0.5cm}
\end{figure}

\section{  Details on Simulated Physical Attacks}
\label{sec:aai_eai_details}
The following provides details on the simulated physical attack models, specifically covering their parameters, and visualizations showing how those parameters vary. We illustrate the attack scenarios and threat models in Fig.\ref{fig:attack_models}

\begin{figure}[h]
\centering
\includegraphics[width=.44\textwidth]{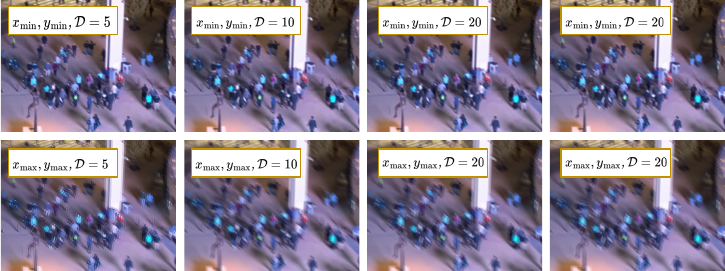}
\vspace{-0.2cm}
\caption{Illustration of the adversarial frame $F^{adv}_t$ from the $\mathcal{P}_{AAI}$ transformation. The top row when fixing $x, y, \phi$ at minimum values while varying hyperparameter $\mathcal{D}$ from 5 (low) to 30 (high). The top row when fixing $x, y, \phi$ at maximum values while varying hyperparameter $\mathcal{D}$ from 5 (low) to 30 (high). }
\label{fig:acoustic_transofrm}
\vspace{-0.3cm}
\end{figure}

\subsection{  Adversarial Acoustic Injection (AAI)}
\label{sec:supp_aai}
The AAI attack simulates the physical effect of sound waves on a camera sensor's mechanical stabilization systems. In modern cameras, Micro-Electro-Mechanical Systems (MEMS) stabilizers are highly susceptible to vibrations induced by carefully tuned acoustic signals, particularly within the ultrasonic frequency range~\cite{zhu2023tpatch, ji2021poltergeist}. This physical vibration causes a subtle, high-frequency motion of the camera sensor during the frame's exposure time, which visually manifests as motion blur in the captured image and can be exploited for adversarial intents~\cite{ji2021poltergeist,zhu2023tpatch}

To accurately simulate this physically-plausible phenomena in a differentiable manner, we model the camera's motion path as a continuous, sinusoidal oscillation. Our AAI simulation, represented by the function $\mathcal{P}_{AAI}(\cdot; \theta_{AAI})$, does not rely on a conventional convolution with a blur kernel. Instead, it utilizes a spatial transformation-based approach that directly models the effect of sensor movement and aggregates the resulting visual data.
The perturbed frame $F^{adv}_t$ is generated from a clean frame $F_t$ according to the following equation:
$$F^{adv}_t = \mathcal{P}_{AAI}(F_t; \theta_{AAI})$$
where the attack's learnable physical parameters are defined as $\theta_{AAI} = (x, y, \phi)$. The parameters $x$ and $y$ represent the maximum amplitude of the horizontal and vertical oscillations, respectively, while $\phi$ is a phase offset for the sinusoidal motion.

\subsubsection{Differentiable Approximation of Blur}
\label{sec:implementation}
In a physical AAI attack, the MEMS sensor resonance ($f_{\text{res}} \approx 26\text{kHz}$) is significantly higher than the camera framerate ($30\text{fps}$). This results in approximately $N_{cyc} \approx 866$ oscillation cycles per exposure duration $T_{\text{exp}}$.

\begin{figure}[h]
\centering
\includegraphics[width=.45\textwidth]{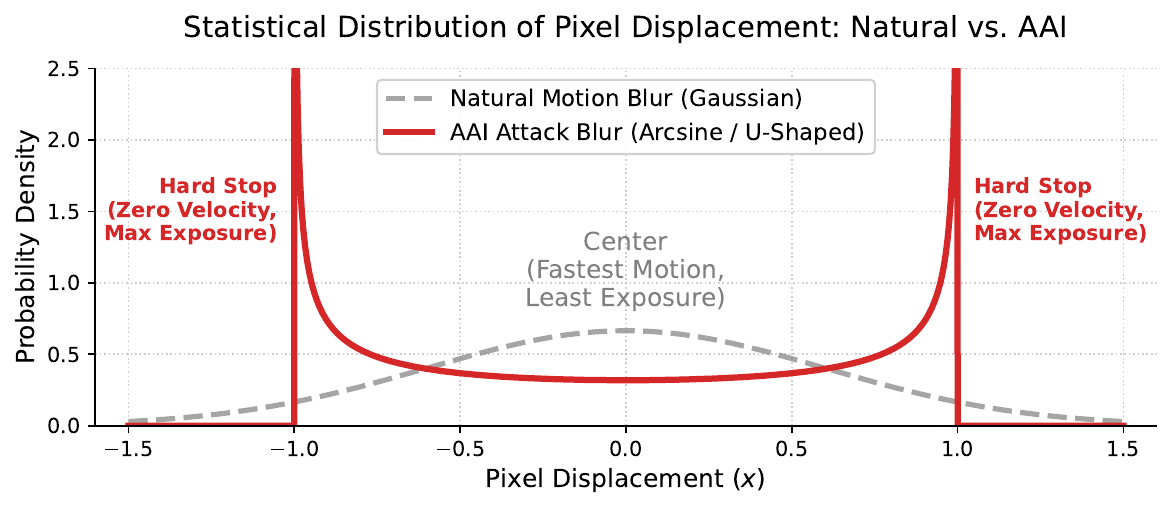}
\vspace{-0.3cm}
\caption{Natural Gaussian blur motion Vs. AAI-induced blur.}
\label{fig:blur_gaussian_compare}
\vspace{-0.3cm}
\end{figure}

Simulating all 866 cycles in a differentiable loop is computationally expensive. However, we observe that the \textit{Point Spread Function (PSF)}, the statistical distribution of the pixel displacement, converges rapidly. For a sinusoidal displacement $d(t) = A \sin(\omega t)$, the intensity distribution of the blur kernel follows a bounded \textit{U-shaped} distribution (the Arcsine distribution), as shown in Fig.\ref{fig:blur_gaussian_compare}, regardless of the frequency $\omega$, provided $\omega \gg 1/T_{\text{exp}}$. Therefore, we approximate the high-frequency integral using a \textit{single-period Monte Carlo approximation}. We model the blur kernel by sampling the displacement trajectory $d(t)$ at $\mathcal{D}$ discrete time steps evenly spaced over one phase cycle $[-\pi, \pi]$:
$$
F^{\text{adv}}_t \approx \frac{1}{\mathcal{D}} \sum_{k=1}^{\mathcal{D}} \mathcal{T}r(F_t, \bm{\delta}_k) \;\; \text{where} \;\; \bm{\delta}_k = (x, y) \cdot \sin\left(\frac{2\pi k}{\mathcal{D}}\right)
$$
where $\mathcal{T}r(\cdot)$ is the differentiable affine transformation operation (implemented via Spatial Transformer Networks). The full simulation process unfolds as follows:
\begin{enumerate}
    \item A set of discrete offsets, $\{\bm{\delta}_k\}_{k=1}^{\mathcal{D}}$, is generated based on the sinusoidal function controlled by parameters $x, y, \phi$.
    \item For each step $k$, a corresponding transformation grid $G_k$ is created. This grid encodes the spatial shift to be applied to the entire image.
    \item The original image $F_t$ is spatially transformed at each step $k$ using its respective grid $G_k$, resulting in a set of $\mathcal{D}$ perturbed image instances, $\{S_k\}_{k=1}^{\mathcal{D}}$.
    \item The final motion-blurred frame $F^{adv}_t$ is obtained by averaging these transformed instances:
    $$F^{adv}_t = \frac{1}{\mathcal{D}} \sum_{k=1}^{\mathcal{D}} S_k$$
\end{enumerate}
In our experiments, we found that $\mathcal{D}=10$ samples provides a sufficient approximation of the Optical Image Stabilization (OIS) blur kernel while maintaining high efficiency. This pipeline is fully differentiable, allowing the adversarial loss to back-propagate through the physical simulation and directly optimize $\theta_{AAI}$.

\subsubsection{AAI Motion Blur vs. Gaussian Blur}
As illustrated in Fig.~\ref{fig:blur_gaussian_compare}, the simulated blur kernel differs significantly from standard Gaussian or defocus blur. Because the AAI attack induces a harmonic vibration $d(t) = \alpha \sin(\omega t)$, the camera sensor spends the majority of the exposure time near the physical limits of the oscillation (the peaks and troughs), where the instantaneous velocity of the sensor approaches zero ($v(t) \to 0$). 
Mathematically, the probability density function of this displacement follows an Arcsine distribution:
$$ p(x) = \frac{1}{\pi \sqrt{\alpha^2 - x^2}}, \quad x \in (-\alpha, \alpha) $$
which approaches infinity at the boundaries $x \to \pm \alpha$. Visually, this does not merely \textit{smear} the object; it creates a distinct \textit{double-edge} or \textit{ghosting} effect. 

\subsubsection{Physical Parameters Calibration and Bounds}
\label{sec:calibration}
To ensure our FADE-AAI attack remains physically realizable, we constrain our learnable parameters based on the hardware limitations and experimental calibration data established in the Poltergeist framework~\cite{ji2021poltergeist}.

\noindent \textbf{(\textit{i}) Resonance Frequency Bounds.}
The effectiveness of acoustic injection is limited to the resonant bandwidth of the MEMS sensor in the targeted camera. Outside this narrow band, the induced drift is negligible.
\begin{itemize}
    \item \textbf{Calibration Method:} The resonant frequency is identified via a frequency sweep (0Hz--30kHz) while monitoring the MEMS sensor raw output.
    \item \textbf{Target Hardware:} For example, for the Samsung Galaxy S20 (SM-G981U) utilized in our AAI simulation model, the STMicroelectronics LSM6DSO MEMS sensor exhibits a sharp resonance peak at $f_{\text{res}} \approx 26.3 \text{kHz}$.
    \item \textbf{Constraint:} In our simulation, $f_{\text{res}}$ is treated as a fixed environmental constant, set to $26.3 \text{kHz}$.
\end{itemize}

\noindent \textbf{(\textit{ii}) Resonance Amplitude Bounds.}
The learnable displacement parameters $(x, y)$ correspond to the amplitude of the induced oscillatory drift. This amplitude is physically constrained by the maximum Sound Pressure Level (SPL) and the device's acoustic transfer function. We bound the magnitude $\|\bm{\alpha_m}\|$ using the empirical log-linear relationship derived in~\cite{ji2021poltergeist}:
$$ \alpha_{\text{drift}} \propto 10^{\frac{\text{SPL}_{dB}}{20}} $$
\begin{itemize}
    \item \textbf{Constraint:} We clamp $\bm{\alpha_m}$ such that the induced angular velocity does not exceed $\pm 2.0 \text{rad/s}$. This corresponds to an acoustic injection of approx. $110 \text{dB}$ at the sensor, the upper limit of standard ultrasonic arrays before non-linear distortion dominates.
\end{itemize}

\noindent \textbf{(\textit{iii}) OIS Physical Saturation Limits ($D_{\text{max}}$).}
The Optical Image Stabilization (OIS) system compensates for drift by physically shifting the lens. This movement is bounded by the Voice Coil Motor (VCM) travel range.
\begin{itemize}
    \item \textbf{Calibration:} Experimental results show that OIS systems typically saturate at displacements equivalent to $1\% \sim 3\%$ of the image width.
    \item \textbf{Hard Constraint:} We apply a differentiable clamping function to the computed displacement trajectory $d(t)$:
    $$ \|d(t)\|_2 \leq D_{\text{max}} $$
    where $D_{\text{max}} = 0.03 \times W$ (approx. 50 pixels for a 1080p stream). Any optimization attempting to push the lens beyond this limit yields diminishing returns, simulating the physical \textit{hard stop} of the OIS mechanism.
\end{itemize}

\begin{figure}[h]
\centering
\includegraphics[width=.48\textwidth]{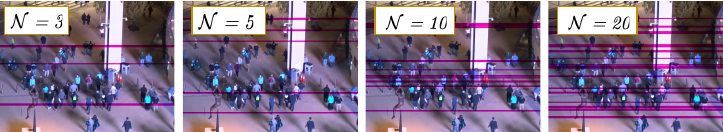}
\vspace{-0.4cm}
\caption{Illustration of the adversarial frame $F^{adv}_t$ from the $\mathcal{P}_{EAI}$ transformation. The top row when applying random color stripes while varying $\mathcal{N}$ from 3 (low) to 20 (high).}
\label{fig:emi_transofrm}
\vspace{-0.4cm}
\end{figure}

\subsection{  Electromagnetic Adversarial Injection (EAI)}
\label{sec:supp_eai}
The EAI attack simulates the effects of electromagnetic interference (EMI) that directly corrupts a camera's electronic signal transmission. A primary target is the Mobile Industry Processor Interface (MIPI) CSI-2 lanes or the Analog-to-Digital Converter (ADC) path, responsible for transmitting the raw sensor data. Such corruption introduces structured noise patterns and color artifacts into the raw image data, which are then propagated through the camera's internal Image Signal Processor (ISP) pipeline~\cite{jiang2023glitchhiker, zhang2024understanding, ren2025ghostshot}.

To simulate this physically-plausible phenomenon in a differentiable manner, we introduce a simulator function $\mathcal{P}_{EAI}(\cdot; \theta_{EAI})$ that mimics the stages of a camera's image acquisition and processing.
The perturbed frame $F^{adv}_t$ is generated as:
$$F^{adv}_t = \mathcal{P}_{EAI}(F_t; \theta_{EAI})$$
where the attack's learnable physical parameters are defined as $\theta_{EAI} = ((\mathbf{r}, \mathbf{w}) \in \mathbb{M})$. Here, $\mathbf{r} \in \mathbb{R}^\mathbb{N}$ represents the vertical row indices (timing delays) of the interference, and $\mathbf{w} \in \mathbb{R}^\mathbb{N}$ represents the width (duration) of each corrupted stripe. The number of stripes, $\mathbb{N}$, is a fixed hyperparameter representing the duty cycle limit of the EMI injector.

\subsubsection{Differentiable Approximation of Corruption}
\label{sec:eai_implementation}
Unlike standard pixel-noise attacks, EAI corruption occurs in the raw signal domain \textit{before} image processing. We model this by inverting the ISP pipeline. Our models is based on the GlitchHiker pipeline and implementation~\cite{jiang2023glitchhiker}. The simulation process unfolds as follows:

\noindent \textbf{(\textit{i}) Raw Sensor Simulation (Inverse ISP):} The clean input image $F_t$ is first converted into a simulated RGGB Bayer pattern $F_{\text{Bayer}}$ using a differentiable masking function that mimics the initial sensor readout.
$$ F_{\text{Bayer}} = F_t \odot \mathbb{M}_{\text{CFA}} $$
    
\noindent \textbf{(\textit{ii}) Sensor-Level Corruption:} We simulate the \textit{Glitch} attack mode identified in~\cite{jiang2023glitchhiker} by selectively zeroing out specific color channels within the Bayer pattern. This mimics the suppression of the differential voltage signal during the EMI pulse.
$$ F_{\text{Corrupted\_Raw}} = F_{\text{Bayer}} \odot \mathbb{M}_{\text{drop}} $$
where $\mathbb{M}_{\text{drop}}$ here zeros out the Green channel (the luminance carrier), which is a characteristic signature of MIPI synchronization failures.
    
\noindent \textbf{(\textit{iii}) Demosaicing Reconstruction:} The corrupted Bayer pattern is reconstructed into a full RGB image via a differentiable bilinear demosaicing algorithm $\mathcal{D}_{demosaic}$. This step generates the non-linear \textit{zippering} and false-color artifacts (e.g., purple stripes) that are distinct from additive noise.
    $$ F_{\text{Artifact}} = \mathcal{D}_{demosaic}(F_{\text{Corrupted\_Raw}}) $$
    
\noindent \textbf{(\textit{iv}) Differentiable Stripe Masking:} To optimize the location of the glitches, a differentiable soft mask $\mathbb{M}_{\text{Soft}}$ is constructed from the learnable parameters $\mathbf{r}$ and $\mathbf{w}$ using a sigmoid product function:
    $$ \mathbb{M}_{\text{Soft}}(y) = \max_{k=1}^N \left[ \sigma(s(y - r_k)) \cdot \sigma(s(r_k + w_k - y)) \right] $$
    This mask allows gradients to flow into the row coordinates, enabling the attacker to learn \textit{where} to inject the pulse. The final adversarial frame is a blend:
    $$F^{adv}_t = F_t \odot (1 - \mathbb{M}_{\text{Soft}}) + F_{\text{Artifact}} \odot \mathbb{M}_{\text{Soft}} \odot \lambda$$
    where $\lambda$ scales the intensity of the corruption.

This entire pipeline is fully differentiable, allowing the adversarial loss to back-propagate through the demosaicing algorithm and directly optimize $\theta_{EAI}$.

\subsubsection{Physical Parameters Calibration and Bounds}
\label{sec:eai_calibration}
To ensure the optimized parameters correspond to a realizable EMI attack, we constrain $\theta_{EAI}$ based on the hardware limitations of standard EMI injection devices (e.g., spark-gap generators or amplified SDRs).

\noindent \textbf{(\textit{i}) Pulse Width Constraints ($\mathbf{w}$).}
The width of the glitch $w_k$ corresponds to the duration of the EMI pulse. Hardware capacitors limit the minimum and maximum pulse duration.
\begin{itemize}
    \item \textbf{Constraint:} $w_{\text{min}} \leq w_k \leq w_{\text{max}}$. We set $w_{\text{min}} \approx 5$ rows (approx. $100\mu s$) and $w_{\text{max}} \approx 50$ rows, reflecting the discharge limits of portable EMI injectors.
\end{itemize}

\noindent \textbf{(\textit{ii}) Pulse Frequency / Count ($\mathbb{N}$).}
High-voltage EMI sources have a recharge delay (duty cycle limit). An attacker cannot jam every single row in a frame.
\begin{itemize}
    \item \textbf{Constraint:} We fix the maximum number of glitches $\mathbb{N}$ (e.g., $\mathbb{N}=20$) to model a realistic recharge cycle, forcing the optimizer to select only the most impactful temporal windows (image rows) for injection.
\end{itemize}

\noindent \textbf{(\textit{iii}) VSYNC Synchronization ($\mathbf{r}$).}
The row index $r_k$ is physically controlled by the time delay $\Delta t$ of the pulse injection relative to the camera's VSYNC interrupt signal.
\begin{itemize}
    \item \textbf{Constraint:} $0 \leq r_k \leq H_{\text{image}}$. The position is fully continuous within the frame readout time, allowing precise targeting of specific object features.
\end{itemize}

\section{  Differentiable Attacks Optimization}
\subsection{  Digital Projected Gradient Descent}
We detail the digital PGD attack. This method generates pixel-level adversarial perturbations by directly optimizing the pixel values of the input image. Unlike physical attacks, which are constrained by the plausibility of their underlying parameters, digital attacks are constrained by the magnitude of the perturbation in the pixel space.
The attack is implemented using a PGD-style iterative optimization loop, which aims to maximize a specific attack loss by taking small, controlled steps in the direction of the loss gradient.

\noindent \textbf{Configuration.}
The PGD attacker is configured with the following key parameters:
\begin{itemize}
\item Maximum Perturbation $(\epsilon)$: Defines the maximum allowed magnitude of the total perturbation in the pixel space. The attack generates perturbations such that they are constrained to an $L_{\infty}$ norm ball, i.e., $||\delta||_{\infty} \le \epsilon$.  We set $\epsilon$ to 8/255.
\item Step Size $(\alpha)$: The size of each step taken during the iterative optimization. It determines how quickly the perturbation is updated. A small $\alpha$ ensures small, controlled updates. We set $\alpha$ to 1/255.
\item Number of Steps $(\mathcal{K})$: The number of iterations for the optimization loop. The attack runs for $\mathcal{K}=50$ steps.
\end{itemize}

\noindent \textbf{Mechanism.} The digital PGD attack is an iterative process that calculates the gradient of the attack loss with respect to the input image and updates the perturbation accordingly. The process, outlined in Algorithm \ref{alg:digital_pgd}, begins by initializing the perturbation $\delta$ to a small value (often zero). In each iteration, a temporary adversarial image $F^{adv}_t = F_t + \delta$ is generated. Then, a forward pass of this perturbed image is performed on the tracker $f_{\text{MOT}}$, and the attack loss $\mathcal{L}_{\textit{adv}}$ is computed. The gradient $\nabla_{\delta} \mathcal{L}_{\text{adv}}$ is back-propagated to determine the direction of the highest loss increase. The perturbation $\delta$ is then updated by taking a step in this direction, and is subsequently clipped to ensure it remains within the predefined $L_{\infty}$ constraint $(\epsilon)$. The final perturbed image is then produced after $\mathcal{K}$ iterations. In our implementation we freeze the model's weights, ensuring that gradients are computed only with respect to the input perturbation, and preventing any changes to the model's parameters. 

\vspace{-.5cm}
\begin{figure}[h]
\centering
\begin{minipage}{.45\textwidth}
\begin{algorithm}[H]
\caption{Digital PGD Attack Algorithm}
\label{alg:digital_pgd}
\textbf{Input:} MOT Tracker $f_{\text{MOT}}$, Frame $F_t$, Attack Parameters $(\text{Perturbation size: }\epsilon, \text{Step size: }\alpha, \text{Iterations: }T)$ \\
\textbf{Output:} $F^{adv}_t$
\begin{algorithmic}[1]
\STATE Initialize perturbation $\delta \leftarrow \mathbf{0}$
\FOR{k = 1 to T}
\STATE $F^{adv}_t \leftarrow \text{Clamp}(F_t + \delta, \text{min}, \text{max})$
\STATE Get loss $\mathcal{L}_{\text{adv}}(f_{\text{MOT}}(F_t^{adv}(\omega)), \mathcal{G}_t)$
\STATE Compute gradient $\nabla_{\delta} \mathcal{L}_{\text{adv}}$
\STATE $\delta \leftarrow \delta + \alpha \cdot \text{sign}(\nabla_{\delta} \mathcal{L}_{\text{adv}})$
\STATE $\delta \leftarrow \text{Clamp}(\delta, -\epsilon, \epsilon)$
\ENDFOR
\RETURN $F^{adv}_t \leftarrow \text{Clamp}(F_t + \delta, \text{min}, \text{max})$
\end{algorithmic}
\end{algorithm}
\end{minipage}
\end{figure}
\vspace{-0.2cm}

\subsection{  Physical Projected Gradient Descent}
The key distinction from the digital attack is that PGD is applied to the physical parameters $(\theta)$ rather than the pixel values $(\delta)$. Unlike the digital PGD attacker, which directly optimizes perturbations in the pixel space, the physical PGD attacker optimizes the parameters of a differentiable physical simulation model. The core principle remains the same: iteratively taking steps in the direction of the loss gradient to maximize an attack objective $\mathcal{L}_{\text{adv}}$. However, in this case, the gradient is computed with respect to the physical parameters $\theta_{(AAI||EAI)}$, and the projection step ensures these parameters remain within their plausible physical ranges, rather than constraining the pixel-space perturbation.

\subsubsection{PGD for Acoustic Attack (AAI).}
The PGD for the AAI attack optimizes the parameters of the motion blur simulation. The algorithm iteratively refines the physical parameters of the simulated camera motion to maximize the adversarial loss $\mathcal{L}_{\text{TMC}}$.

\noindent \textbf{Configuration.}
The PGD loop for AAI operates on the physical parameters $\theta_{AAI} = (x, y, \phi)$, which represent the amplitudes of horizontal and vertical sinusoidal oscillations, and a phase offset, respectively. These parameters are bounded by a predefined plausible range ($[min_x, max_x]$, $[min_y, max_y]$, $[min_\phi, max_\phi]$). The number of blur samples, $\mathcal{D}$, is a fixed hyperparameter for a given attack configuration. The step size, $\alpha$, is applied to each physical parameter independently during optimization. The attack runs for $\mathcal{K}=150$ steps.

\begin{figure}[ht]
\centering
\begin{minipage}{.45\textwidth}
\begin{algorithm}[H]
\caption{PGD for Acoustic Attack (FADE-AAI)}
\label{alg:aai_pgd}
\textbf{Input:} Tracker $f_{\text{MOT}}$, Frame $F_t$, AAI Attack Parameters ($\theta_{AAI}=(x, y, \phi, \mathbb{D}), \text{Step size: }\alpha, \text{Iterations: }T)$ \\
\textbf{Output:} $F^{adv}_t$
\begin{algorithmic}[1]
\STATE Initialize physical parameters $x, y, \phi \in \theta_{AAI}$
\FOR{k = 1 to $\mathcal{K}$}
\STATE $F^{adv}_t \leftarrow \mathcal{P}_{AAI}(F_t; x, y, \phi)$
\STATE Get loss $\mathcal{L}_{\text{adv}}(f_{\text{MOT}}(F_t^{adv}(\omega)), \mathcal{G}_t)$
\STATE Compute gradients $\nabla_x, \nabla_y, \nabla_{\phi} \mathcal{L}_{\text{adv}}$
\STATE $x \leftarrow x + \alpha \cdot \text{sign}(\nabla_x)$
\STATE $y \leftarrow y + \alpha \cdot \text{sign}(\nabla_y)$
\STATE $\phi \leftarrow \phi + \alpha \cdot \text{sign}(\nabla_{\phi})$
\STATE $\text{Clamp}_x(x), \text{Clamp}_y(y), \text{Clamp}_{\phi}(\phi)$
\ENDFOR
\RETURN $F^{adv}_t \leftarrow \mathcal{P}_{AAI}(F_t; x, y, \phi)$
\end{algorithmic}
\end{algorithm}
\end{minipage}
\end{figure}

\noindent \textbf{Mechanism:}
The PGD for AAI, outlined in Algorithm \ref{alg:aai_pgd}, initializes the parameters $x, y, \phi$ to their minimum values. In each iteration, a perturbed image $F^{adv}_t$ is generated by passing the current image and the current parameters to the differentiable AAI simulation model $\mathcal{P}_{AAI}$. A forward pass on the target tracker $f_{\text{MOT}}$ is performed with $F^{adv}_t$, and the adversarial loss $\mathcal{L}_{TMC}$ is computed. The gradients of this loss with respect to $x, y, \phi$ are then calculated. The parameters are updated using a signed gradient step and are subsequently clamped back into their predefined physical ranges, which serves as the projection step in this optimization.

\subsubsection{PGD for Electromagnetic Attack (EAI)}
The PGD for the EAI attack optimizes the spatial parameters that define the adversarial stripes. The algorithm iteratively refines the location and dimensions of these corrupted regions to maximize the adversarial loss $\mathcal{L}_{TMC}$.

\noindent \textbf{Configuration.}
PGD for EAI operates on the $\mathbb{M} \in \theta_{EAI}$, which is a 2D matrix where each row [$r$, $w$] defines a horizontal stripe's row ID and width. The maximum number of stripes, $\mathbb{N}$, is a fixed hyperparameter for a given attack. The step size, $\alpha$, is applied to all parameters within $\mathbb{M}$ simultaneously. The plausible range for these parameters (e.g., $r$ within image height, $w$ within valid bounds) is enforced as a projection constraint. The attack runs for $\mathcal{K}=150$.

\noindent \textbf{Mechanism.}
The PGD algorithm for EAI, outlined in Algorithm \ref{alg:eai_pgd}, initializes the $\mathbb{M}$ to a predefined configuration (e.g., uniform, random). In each iteration, a perturbed image $F^{adv}_t$ is generated by passing the current $\mathbb{M}$ to the differentiable EAI simulation model $\mathcal{P}_{EAI}$. A forward pass on the target tracker $f_{\text{MOT}}$ is performed with $F^{adv}_t$, and the adversarial loss $\mathcal{L}_{TMC}$ is computed. The gradients of this loss with respect to $\mathbb{M}$ are then calculated. The $\mathbb{M}$ tensor is updated using a signed gradient step. The projection back into the plausible parameter space is implicitly handled by clamping functions within the EAI simulation model.

\begin{figure}[ht]
\centering
\begin{minipage}{.45\textwidth}
\begin{algorithm}[H]
\caption{PGD for Electromagnetic Attack (EAI)}
\label{alg:eai_pgd}
\textbf{Input:} Tracker $f_{\text{MOT}}$, Frame $F_t$, AAI Attack Parameters ($\theta_{EAI}=(\mathbb{M}, \mathbb{N}), \text{Step size: }\alpha, \text{Iterations: }T)$ \\
\textbf{Output:} $F^{adv}_t$
\begin{algorithmic}[1]
\STATE Initialize $\mathbb{M}, \theta_{EAI} \in \theta_{EAI}$
\FOR{k = 1 to $\mathcal{K}$}
\STATE $F^{adv}_t \leftarrow \mathcal{P}_{EAI}(F_t; \theta_{EAI})$
\STATE Get loss $\mathcal{L}_{attack}(\mathcal{M}(F^{adv}_t), \mathcal{G}_t)$
\STATE Compute gradient $\nabla_{\theta_{EAI}} \mathcal{L}_{attack}$
\STATE $\theta_{EAI} \leftarrow \theta_{EAI} + \alpha \cdot \text{sign}(\nabla_{\theta_{EAI}} \mathcal{L}_{attack})$
\STATE $\theta_{EAI} \leftarrow \text{Clip}_{\Theta}(\theta_{EAI})$
\ENDFOR
\RETURN $F^{adv}_t \leftarrow \mathcal{P}_{EAI}(F_t; \theta_{EAI})$
\end{algorithmic}
\end{algorithm}
\end{minipage}
\end{figure}

\section{  Additional Evaluations}
We provide additional evaluations of FADE, including further comparison with baselines, ablation studies, transferability analysis, results of the TQF-guided simulated physical attacks, analysis of the long-term memory integrity in TBP trackers, and the efficacy of FADE against defenses. 

\vspace{-0.2cm}
\begin{table}[h]
\centering
\caption{\small FADE vs. Untargeted PGD and recent MOT attack.}
\vspace{-0.35cm}
\label{tab:results}
\scalebox{.48}{
\begin{tabular}{lc|c|c|c|c|c} 
\toprule
\textbf{Attack} & \textbf{Dataset} & \textbf{MOTR} & \textbf{MOTRv2} & \textbf{MeMOTR} & \textbf{Samba} & \textbf{CO-MOT} \\ 
\hline
Clean & MOT17 \textbar{} MOT20 & 58.63 \textbar{} 55.40 & 59.96 \textbar{} 56.70 & 67.35 \textbar{} 68.20 & 62.91 \textbar{} 64.50 & 58.16 \textbar{} 61.20 \\
\hline
Untargeted PGD & MOT17 \textbar{} MOT20 & 49.90 \textbar{} 40.10 & 57.41 \textbar{} 57.77 & 59.85 \textbar{} 60.74 & 58.38 \textbar{} 57.61 & 55.85 \textbar{} 60.95 \\
BBA\_Blind & MOT17 \textbar{} MOT20 & 48.51 \textbar{} 41.19 & 56.07 \textbar{} 59.36 & 59.10 \textbar{} 58.47 & 56.79 \textbar{} 54.27 & 53.02 \textbar{} 57.56 \\
BBA\_Blur & MOT17 \textbar{} MOT20 & 46.92 \textbar{} \textcolor{red}{35.85} & 57.13 \textbar{} 58.13 & 62.54 \textbar{} 64.67 & 56.52 \textbar{} 51.63 & 53.55 \textbar{} 58.97 \\
\hline
\textbf{$\text{FADE}_{\text{TMC}}$} & MOT17 \textbar{} MOT20 & \textcolor{red}{45.90} \textbar{} 42.63 & \textcolor{red}{39.29} \textbar{} \textcolor{red}{29.64} & \textcolor{red}{41.41} \textbar{} \textcolor{red}{57.67} & \textcolor{red}{45.53} \textbar{} \textcolor{red}{46.85} & \textcolor{red}{37.26} \textbar{} \textcolor{red}{33.37} \\
\textbf{$\text{FADE}_{\text{TQF}}$} & MOT17 \textbar{} MOT20 & \textcolor{red}{45.89} \textbar{} 40.38 & \textcolor{red}{46.76} \textbar{} \textcolor{red}{56.68} & \textcolor{red}{41.56} \textbar{} \textcolor{red}{37.70} & \textcolor{red}{48.04} \textbar{} \textcolor{red}{54.91} & \textcolor{red}{41.73} \textbar{} \textcolor{red}{49.28} \\
\bottomrule
\end{tabular}
}
\end{table}
\vspace{-0.3cm}

\subsection{  Comparison with Additional Baselines}
\label{sec:sm_baselines}

As FADE represents the first adversarial framework specifically targeting TBP trackers, native baselines designed for this architecture are currently absent in the literature. For a rigorous evaluation, we extend the cross-paradigm comparisons presented in the main paper by evaluating FADE against three additional baseline strategies: standard \textit{Untargeted PGD}, and two spatial-degradation attacks, \textit{BBA-Blind} and \textit{BBA-Blur}~\cite{pang2024blinding}. Several key insights emerge from this extended comparison. While spatial-degradation attacks can degrade performance in high-density scenes or on weaker detection backbones, they fail to induce the systemic identity collapse observed with FADE, which achieves disproportionately higher degradation on robust, memory-augmented models such as MeMOTR, Samba, and CO-MOT. Furthermore, the significant performance gap between FADE and Untargeted PGD verifies that TBP vulnerabilities are fundamentally architectural; standard gradient-based noise is insufficient to disrupt the recurrent state propagation that FADE specifically poisons. Ultimately, while standard MOT attacks are primarily designed for spatial disruption, FADE exploits the temporal dependencies of the query-update logic to trigger a non-transient \textit{forgetting} state that standard baselines cannot replicate.

\subsection{  Ground-Truth Label Independence}
\label{sec:sm_cost_mimicry}

To verify the practical feasibility of the TQF attack, we investigate the impact of replacing privileged Ground Truth (GT) information with online tracker predictions for the cost mimicry loss ($\mathcal{L}_{\text{Cost}}$). While the primary experiments in Section~4.7.1 utilize GT to set a performance upper-bound for the attack, the results summarized in Table~\ref{tab:gt_cost} confirm that FADE operates with near-identical efficacy using surrogate labels derived directly from the tracker's own predictions. We observe a negligible average HOTA delta of $\le 0.2$ points across all tested models when transitioning from GT to predicted labels, reinforcing that FADE does not require privileged information to remain robust. Furthermore, this ablation clarifies the specific role of the $\mathcal{L}_{\text{Cost}}$ component: while $\mathcal{L}_{\text{Flood}}$ drives the primary exhaustion of the query budget, $\mathcal{L}_{\text{Cost}}$ is necessary for bypassing the temporal consistency filters (e.g., State Space Models) inherent in memory-heavy architectures like Samba. Without cost mimicry, the attack's impact on Samba is significantly diminished, whereas the inclusion of $\mathcal{L}_{\text{Cost}}$, even when guided by noisy predictions, successfully deceives the temporal updater into siphoning legitimate track identities.

\begin{table}[h]
\centering
\caption{\small Ablation on Cost Mimicry in the TQF Loss (HOTA)}
\vspace{-0.35cm}
\label{tab:gt_cost}
\scalebox{.49}{
\begin{tabular}{lc|c|c|c|c|c} 
\toprule
\textbf{Attack} & \textbf{Dataset} & \textbf{MOTR} & \textbf{MOTRv2} & \textbf{MeMOTR} & \textbf{Samba} & \textbf{CO-MOT} \\ 
\hline
TQF w/ GT & MOT17 \textbar{} MOT20 & 45.90 \textbar{}~42.63 & 39.29 \textbar{}~29.64 & 41.41 \textbar{}~57.67 & 45.53 \textbar{}~46.85 & 37.26 \textbar{}~33.37 \\
TQF w/ Pred & MOT17 \textbar{} MOT20 & 45.91 \textbar{}~42.46 & 39.17 \textbar{}~29.69 & 42.26 \textbar{}~58.25 & 44.82 \textbar{}~46.31 & 37.11 \textbar{}~33.30 \\
TQF w/o $\mathcal{L}_{\text{Cost}}$ & MOT17 \textbar{} MOT20 & 45.84 \textbar{}~42.42 & 38.48 \textbar{}~29.67 & 42.53~\textbar{} 59.38 & 53.14 \textbar{}~55.62 & 36.44 \textbar{}~33.58 \\
\bottomrule
\end{tabular}
}
\end{table}

\subsection{  TQF Guided Physical Attacks}
\label{sec:tqf_phy_analysis}
We further evaluate the efficacy of the FADE's TQF attack when optimizing the physical AAI and EAI perturbations. Tables \ref{tab:phy_results_mot17_add} and \ref{tab:phy_results_mot20_add} breakdown the performances on the MOT17 and MOT20 benchmarks, respectively.

\begin{table}[h]
\centering
\scriptsize
\setlength{\tabcolsep}{3.5pt}
\renewcommand{\arraystretch}{1.0}
\caption{Simulated Physical FADE-TQF Attacks on MOT17.}  
\label{tab:phy_results_mot17_add}
\vspace{-0.3cm}
\scalebox{.83}{
\begin{tabular}{llccccccc}
\toprule
\rowcolor{HeaderGray}
\multicolumn{9}{c}{
\textbf{MOT17 Dataset – Average scene density $\sim$21~objects/frame.}
}\\[-0.2em]
\midrule
\textbf{Tracker} & \textbf{Attack\_Vector} & \textbf{HOTA}$\downarrow$ & \textbf{DetA}$\downarrow$ & \textbf{AssA}$\downarrow$ & \textbf{IDF1}$\downarrow$ & \textbf{IDR}$\downarrow$ & \textbf{IDP}$\downarrow$ & \textbf{IDSW}$\uparrow$ \\
\midrule
\multirow{3}{*}{\textbf{MOTR}}
 & {\cellcolor{CleanGreen}}Clean 
   & {\cellcolor{CleanGreen}}58.63 
   & {\cellcolor{CleanGreen}}49.90 
   & {\cellcolor{CleanGreen}}70.38 
   & {\cellcolor{CleanGreen}}69.35 
   & {\cellcolor{CleanGreen}}68.48 
   & {\cellcolor{CleanGreen}}71.40 
   & {\cellcolor{CleanGreen}}7.23 \\
 & TQF\_AAI 
   & \textbf{\textcolor{DeepRed}{52.66}} 
   & \textbf{\textcolor{DeepRed}{45.19}} 
   & \textbf{\textcolor{DeepRed}{62.80}} 
   & \textbf{\textcolor{DeepRed}{64.66}} 
   & \textbf{\textcolor{DeepRed}{61.06}} 
   & \textbf{\textcolor{DeepRed}{70.08}} 
   & \textbf{\textcolor{DeepRed}{7.70}} \\
 & TQF\_EAI 
   & \textbf{\textcolor{DeepRed}{42.74}} 
   & \textbf{\textcolor{DeepRed}{48.15}} 
   & \textbf{\textcolor{DeepRed}{39.04}} 
   & \textbf{\textcolor{DeepRed}{47.68}} 
   & \textbf{\textcolor{DeepRed}{42.49}} 
   & \textbf{\textcolor{DeepRed}{54.77}} 
   & \textbf{\textcolor{DeepRed}{8.87}} \\
\midrule
\multirow{3}{*}{\textbf{MOTRv2}}
 & {\cellcolor{CleanGreen}}Clean 
   & {\cellcolor{CleanGreen}}59.96 
   & {\cellcolor{CleanGreen}}49.15 
   & {\cellcolor{CleanGreen}}74.71 
   & {\cellcolor{CleanGreen}}71.99 
   & {\cellcolor{CleanGreen}}60.89 
   & {\cellcolor{CleanGreen}}89.68 
   & {\cellcolor{CleanGreen}}1.75 \\
 & TQF\_AAI 
   & \textbf{\textcolor{DeepRed}{52.50}} 
   & \textbf{\textcolor{DeepRed}{43.08}} 
   & \textbf{\textcolor{DeepRed}{65.32}} 
   & \textbf{\textcolor{DeepRed}{65.06}} 
   & \textbf{\textcolor{DeepRed}{52.48}} 
   & \textbf{\textcolor{DeepRed}{87.28}} 
   & \textbf{\textcolor{DeepRed}{2.64}} \\
 & TQF\_EAI 
   & \textbf{\textcolor{DeepRed}{53.79}} 
   & \textbf{\textcolor{DeepRed}{44.26}} 
   & \textbf{\textcolor{DeepRed}{66.68}} 
   & \textbf{\textcolor{DeepRed}{66.79}} 
   & \textbf{\textcolor{DeepRed}{53.87}} 
   & \textbf{\textcolor{DeepRed}{89.35}} 
   & \textbf{\textcolor{DeepRed}{1.94}} \\
\midrule
\multirow{3}{*}{\textbf{MeMOTR}}
 & {\cellcolor{CleanGreen}}Clean 
   & {\cellcolor{CleanGreen}}67.35 
   & {\cellcolor{CleanGreen}}57.87 
   & {\cellcolor{CleanGreen}}79.60 
   & {\cellcolor{CleanGreen}}80.83 
   & {\cellcolor{CleanGreen}}70.78 
   & {\cellcolor{CleanGreen}}94.84 
   & {\cellcolor{CleanGreen}}0.81 \\
 & TQF\_AAI 
   & \textbf{\textcolor{DeepRed}{56.55}} 
   & \textbf{\textcolor{DeepRed}{47.76}} 
   & \textbf{\textcolor{DeepRed}{68.03}} 
   & \textbf{\textcolor{DeepRed}{70.95}} 
   & \textbf{\textcolor{DeepRed}{57.35}} 
   & \textbf{\textcolor{DeepRed}{93.59}} 
   & \textbf{\textcolor{DeepRed}{1.29}} \\
 & TQF\_EAI 
   & \textbf{\textcolor{DeepRed}{59.80}} 
   & \textbf{\textcolor{DeepRed}{51.07}} 
   & \textbf{\textcolor{DeepRed}{71.09}} 
   & \textbf{\textcolor{DeepRed}{74.76}} 
   & \textbf{\textcolor{DeepRed}{62.02}} 
   & \textbf{\textcolor{DeepRed}{94.74}} 
   & \textbf{\textcolor{DeepRed}{1.01}} \\
\midrule
\multirow{3}{*}{\textbf{Samba}}
 & {\cellcolor{CleanGreen}}Clean 
   & {\cellcolor{CleanGreen}}62.91 
   & {\cellcolor{CleanGreen}}50.58 
   & {\cellcolor{CleanGreen}}79.37 
   & {\cellcolor{CleanGreen}}73.67 
   & {\cellcolor{CleanGreen}}60.30 
   & {\cellcolor{CleanGreen}}95.93 
   & {\cellcolor{CleanGreen}}1.02 \\
 & TQF\_AAI 
   & \textbf{\textcolor{DeepRed}{48.94}} 
   & \textbf{\textcolor{DeepRed}{37.95}} 
   & \textbf{\textcolor{DeepRed}{64.15}} 
   & \textbf{\textcolor{DeepRed}{59.15}} 
   & \textbf{\textcolor{DeepRed}{43.75}} 
   & \textbf{\textcolor{DeepRed}{93.01}} 
   & \textbf{\textcolor{DeepRed}{1.94}} \\
 & TQF\_EAI 
   & \textbf{\textcolor{DeepRed}{56.67}} 
   & \textbf{\textcolor{DeepRed}{45.88}} 
   & \textbf{\textcolor{DeepRed}{71.01}} 
   & \textbf{\textcolor{DeepRed}{69.16}} 
   & \textbf{\textcolor{DeepRed}{54.59}} 
   & \textbf{\textcolor{DeepRed}{95.37}} 
   & \textbf{\textcolor{DeepRed}{1.14}} \\
\midrule

\multirow{3}{*}{\textbf{CO-MOT}}
 & {\cellcolor{CleanGreen}}Clean 
   & {\cellcolor{CleanGreen}}58.16 
   & {\cellcolor{CleanGreen}}46.22 
   & {\cellcolor{CleanGreen}}74.87 
   & {\cellcolor{CleanGreen}}69.87 
   & {\cellcolor{CleanGreen}}57.21 
   & {\cellcolor{CleanGreen}}91.97 
   & {\cellcolor{CleanGreen}}1.83 \\
 & TQF\_AAI 
   & \textbf{\textcolor{DeepRed}{50.80}} 
   & \textbf{\textcolor{DeepRed}{40.02}} 
   & \textbf{\textcolor{DeepRed}{65.98}} 
   & \textbf{\textcolor{DeepRed}{63.15}} 
   & \textbf{\textcolor{DeepRed}{49.45}} 
   & \textbf{\textcolor{DeepRed}{89.66}} 
   & \textbf{\textcolor{DeepRed}{2.96}} \\
 & TQF\_EAI 
   & \textbf{\textcolor{DeepRed}{51.72}} 
   & \textbf{\textcolor{DeepRed}{41.22}} 
   & \textbf{\textcolor{DeepRed}{66.48}} 
   & \textbf{\textcolor{DeepRed}{64.64}} 
   & \textbf{\textcolor{DeepRed}{51.16}} 
   & \textbf{\textcolor{DeepRed}{90.46}} 
   & \textbf{\textcolor{DeepRed}{3.00}} \\
\bottomrule
\end{tabular}
}
\vspace{-0.2cm}
\end{table}

\begin{table}[h]
\centering
\scriptsize
\setlength{\tabcolsep}{3.5pt}
\renewcommand{\arraystretch}{1.0}
\caption{Simulated Physical FADE-TQF Attacks on MOT20.}  
\label{tab:phy_results_mot20_add}
\vspace{-0.3cm}
\scalebox{.8}{
\begin{tabular}{llccccccc}
\toprule
\rowcolor{HeaderGray}
\multicolumn{9}{c}{
\textbf{MOT20 Dataset – High Density: Avg. $\sim$150 objects/frame.}
}\\[-0.2em]
\midrule
\textbf{Tracker} & \textbf{Attack\_Vector} & \textbf{HOTA}$\downarrow$ & \textbf{DetA}$\downarrow$ & \textbf{AssA}$\downarrow$ & \textbf{IDF1}$\downarrow$ & \textbf{IDR}$\downarrow$ & \textbf{IDP}$\downarrow$ & \textbf{IDSW}$\uparrow$ \\
\midrule
\multirow{3}{*}{\textbf{MOTR}}
 & {\cellcolor{CleanGreen}}Clean 
   & {\cellcolor{CleanGreen}}55.06 
   & {\cellcolor{CleanGreen}}40.57 
   & {\cellcolor{CleanGreen}}75.89 
   & {\cellcolor{CleanGreen}}64.10 
   & {\cellcolor{CleanGreen}}57.65 
   & {\cellcolor{CleanGreen}}74.80 
   & {\cellcolor{CleanGreen}}5.14 \\
 & TQF\_AAI
   & \textbf{\textcolor{DeepRed}{50.47}}
   & \textbf{\textcolor{DeepRed}{37.94}}
   & \textbf{\textcolor{DeepRed}{68.31}}
   & \textbf{\textcolor{DeepRed}{61.04}}
   & \textbf{\textcolor{DeepRed}{52.36}}
   & \textbf{\textcolor{DeepRed}{75.36}}
   & \textbf{\textcolor{DeepRed}{5.45}} \\
 & TQF\_EAI
   & \textbf{\textcolor{DeepRed}{41.97}}
   & \textbf{\textcolor{DeepRed}{36.05}}
   & \textbf{\textcolor{DeepRed}{52.00}}
   & \textbf{\textcolor{DeepRed}{48.81}}
   & \textbf{\textcolor{DeepRed}{38.48}}
   & \textbf{\textcolor{DeepRed}{68.60}}
   & \textbf{\textcolor{DeepRed}{6.60}} \\
\midrule
\multirow{3}{*}{\textbf{MOTRv2}}
 & {\cellcolor{CleanGreen}}Clean 
   & {\cellcolor{CleanGreen}}59.56 
   & {\cellcolor{CleanGreen}}44.74 
   & {\cellcolor{CleanGreen}}80.71 
   & {\cellcolor{CleanGreen}}69.55 
   & {\cellcolor{CleanGreen}}54.90 
   & {\cellcolor{CleanGreen}}96.81 
   & {\cellcolor{CleanGreen}}0.73 \\
 & TQF\_AAI
   & \textbf{\textcolor{DeepRed}{53.75}}
   & \textbf{\textcolor{DeepRed}{40.34}}
   & \textbf{\textcolor{DeepRed}{72.93}}
   & \textbf{\textcolor{DeepRed}{64.76}}
   & \textbf{\textcolor{DeepRed}{49.18}}
   & \textbf{\textcolor{DeepRed}{97.02}}
   & \textbf{\textcolor{DeepRed}{0.80}} \\
 & TQF\_EAI
   & \textbf{\textcolor{DeepRed}{52.97}}
   & \textbf{\textcolor{DeepRed}{39.91}}
   & \textbf{\textcolor{DeepRed}{71.54}}
   & \textbf{\textcolor{DeepRed}{63.92}}
   & \textbf{\textcolor{DeepRed}{48.39}}
   & \textbf{\textcolor{DeepRed}{96.09}}
   & \textbf{\textcolor{DeepRed}{0.89}} \\
\midrule
\multirow{3}{*}{\textbf{MeMOTR}}
 & {\cellcolor{CleanGreen}}Clean 
   & {\cellcolor{CleanGreen}}69.61 
   & {\cellcolor{CleanGreen}}59.19 
   & {\cellcolor{CleanGreen}}83.17 
   & {\cellcolor{CleanGreen}}83.31 
   & {\cellcolor{CleanGreen}}72.84 
   & {\cellcolor{CleanGreen}}97.99 
   & {\cellcolor{CleanGreen}}0.46 \\
 & TQF\_AAI
   & \textbf{\textcolor{DeepRed}{59.24}}
   & \textbf{\textcolor{DeepRed}{49.89}}
   & \textbf{\textcolor{DeepRed}{71.45}}
   & \textbf{\textcolor{DeepRed}{73.99}}
   & \textbf{\textcolor{DeepRed}{60.36}}
   & \textbf{\textcolor{DeepRed}{96.58}}
   & \textbf{\textcolor{DeepRed}{0.86}} \\
 & TQF\_EAI
   & \textbf{\textcolor{DeepRed}{62.29}}
   & \textbf{\textcolor{DeepRed}{52.95}}
   & \textbf{\textcolor{DeepRed}{74.41}}
   & \textbf{\textcolor{DeepRed}{77.68}}
   & \textbf{\textcolor{DeepRed}{64.72}}
   & \textbf{\textcolor{DeepRed}{97.75}}
   & \textbf{\textcolor{DeepRed}{0.57}} \\
\midrule
\multirow{3}{*}{\textbf{Samba}}
 & {\cellcolor{CleanGreen}}Clean 
   & {\cellcolor{CleanGreen}}62.49 
   & {\cellcolor{CleanGreen}}47.83 
   & {\cellcolor{CleanGreen}}83.29 
   & {\cellcolor{CleanGreen}}72.47 
   & {\cellcolor{CleanGreen}}58.23 
   & {\cellcolor{CleanGreen}}98.27 
   & {\cellcolor{CleanGreen}}0.43 \\
 & TQF\_AAI
   & \textbf{\textcolor{DeepRed}{48.93}}
   & \textbf{\textcolor{DeepRed}{34.56}}
   & \textbf{\textcolor{DeepRed}{70.97}}
   & \textbf{\textcolor{DeepRed}{57.08}}
   & \textbf{\textcolor{DeepRed}{41.19}}
   & \textbf{\textcolor{DeepRed}{96.94}}
   & \textbf{\textcolor{DeepRed}{0.81}} \\
 & TQF\_EAI
   & \textbf{\textcolor{DeepRed}{55.29}}
   & \textbf{\textcolor{DeepRed}{41.71}}
   & \textbf{\textcolor{DeepRed}{74.79}}
   & \textbf{\textcolor{DeepRed}{66.22}}
   & \textbf{\textcolor{DeepRed}{50.62}}
   & \textbf{\textcolor{DeepRed}{98.27}}
   & \textbf{\textcolor{DeepRed}{0.51}} \\
\midrule

\multirow{3}{*}{\textbf{CO-MOT}}
 & {\cellcolor{CleanGreen}}Clean 
   & {\cellcolor{CleanGreen}}64.31 
   & {\cellcolor{CleanGreen}}52.30 
   & {\cellcolor{CleanGreen}}80.35 
   & {\cellcolor{CleanGreen}}78.01 
   & {\cellcolor{CleanGreen}}65.43 
   & {\cellcolor{CleanGreen}}97.78 
   & {\cellcolor{CleanGreen}}0.45 \\
 & TQF\_AAI
   & \textbf{\textcolor{DeepRed}{57.40}}
   & \textbf{\textcolor{DeepRed}{46.52}}
   & \textbf{\textcolor{DeepRed}{72.00}}
   & \textbf{\textcolor{DeepRed}{72.23}}
   & \textbf{\textcolor{DeepRed}{57.77}}
   & \textbf{\textcolor{DeepRed}{97.75}}
   & \textbf{\textcolor{DeepRed}{0.60}} \\
 & TQF\_EAI
   & \textbf{\textcolor{DeepRed}{56.28}}
   & \textbf{\textcolor{DeepRed}{45.29}}
   & \textbf{\textcolor{DeepRed}{71.14}}
   & \textbf{\textcolor{DeepRed}{70.67}}
   & \textbf{\textcolor{DeepRed}{56.06}}
   & \textbf{\textcolor{DeepRed}{97.20}}
   & \textbf{\textcolor{DeepRed}{0.66}} \\
\bottomrule
\end{tabular}
}
\vspace{-0.4cm}
\end{table}

\subsubsection{  Efficacy of TQF in Physical Constraints}
The TQF attack, designed to exhaust the tracker's query budget with spurious, high-confidence temporal tracks, proves effective even under simulated physical attacks. On MOT17, the TQF-EAI attack consistently degrades HOTA scores by 10-16 points across all trackers. For the purely query-based MOTR, TQF-EAI reduces HOTA from 58.63 to 42.74, a signifcant 27\% performance drop. A critical finding is the fragility of the Samba tracker to TQF-AAI. On MOT20, Samba's HOTA collapses from 62.49 to 48.93. The AAI-induced blur appears to interact destructively with the SSM's sequential scanning mechanism, allowing the flooding queries to easily overwhelm the state transitions, leading to massive detection loss (DetA drops from 47.83 to 34.56). In the high-density MOT20 dataset, the TQF attack is particularly powerful. Since the scene already contains $\sim$150 legitimate objects, the tracker's query budget is naturally near saturation. The TQF attack exploits this by pushing the system into failure with fewer adversarial perturbations than required in sparse scenes.

\begin{figure*}[t]
\centering
\includegraphics[width=1.0\textwidth]{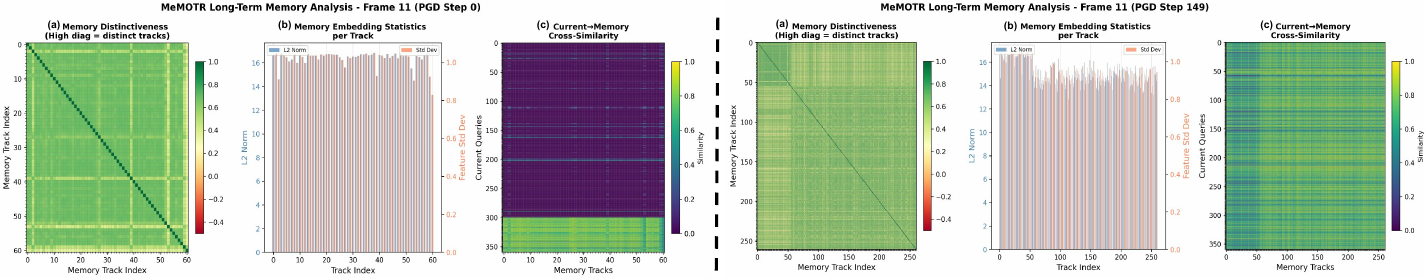}
\vspace{-0.4cm}
\caption{Analysis of Long-term Memory Integrity in TBP Trackers. The figure depicts the state of the MeMOTR~\cite{gao2023memotr}'s memory bank before (PGD Step 0) and after (PGD Step 149) the FADE attack, showing how FADE undermines the tracking identity memory mechanism.}
\label{fig:memory_state}
\vspace{-0.4cm}
\end{figure*}

\subsubsection{TMC vs. TQF in Simulated Physical Settings}
Comparing the two attack modalities (Tables \ref{tab:phy_results_mot17}/\ref{tab:phy_results_mot20} vs. Tables \ref{tab:phy_results_mot17_add}/\ref{tab:phy_results_mot20_add}) reveals distinct failure modes:

\noindent \textbf{(\textit{i}) Failure Mode Specificity (AssA vs. DetA):}
The TMC attack consistently causes larger drops in AssA than DetA. For example, on MOTR (MOT17), TMC-EAI drops AssA by $\sim$31 points, while DetA remains relatively stable. This confirms TMC breaks the \textit{temporal link}.
Conversely, TQF attacks often degrade DetA and Recall (IDR) more severely. On MOT20, TQF-AAI drops Samba's DetA by $\sim$13 points. This confirms TQF acts as a \textit{Denial-of-Service} attack, exhausting the query budget and forcing the tracker to discard and drop valid object detections entirely.

\noindent \textbf{(\textit{ii}) Robustness to Architecture:}
MeMOTR, with its dedicated long-term memory bank, shows higher resilience to TQF than TMC. Its explicit memory module seemingly helps it distinguish between \textit{flooding} noise and valid tracks better than it can handle the direct memory corruption of TMC. State-Space Models (Samba) appear uniquely vulnerable to TQF-AAI (Blur), suffering larger drops than purely Transformer-based models.

\noindent \textbf{(\textit{iii}) Attack Vector Synergy:}
The sharp, high-frequency artifacts of the EAI (Glitch) attack appear suited for the precise feature-cutting required by TMC. The smeared, broad artifacts of the AAI (Blur) attack seem effective at generating the wide-area confusion necessary for TQF flooding, especially in trackers with localized attention windows.

In summary, while both attacks are effective, they compromise TBP trackers through orthogonal mechanisms: TMC acts as a \textit{surgical strike} on identity integrity, while TQF acts as a \textit{capacity overload} on the detection pipeline.

\subsection{  Long-Term Memory State Analysis}
\noindent We analyze how the long-term tracking memory is compromised under FADE's TMC attack. The $\mathcal{L}_{\text{TMC}}$ adversarial loss is designed to fundamentally undermine the memory integrity in advanced TBP trackers, like MeMOTR~\cite{gao2023memotr}, used for temporal robustness and long-term identity association. In Fig.\ref{fig:memory_state}, by comparing the memory state before attack (PGD Step 0, clean) and at the final attack optimization step (PGD Step 149, fully attacked), we empirically validate the success of the attack in three critical dimensions.

\subsubsection{Memory Distinctiveness (Self-Similarity Matrix)}
\label{sec:mem_distinctiveness}
This analysis (Fig.\ref{fig:memory_state}.a plots) measures the self-similarity matrix ($\mathcal{M} \cdot \mathcal{M}^T$) of the stored track embeddings.
\begin{itemize}
    \item \textbf{PGD Step 0 (Clean):} The strong, high-valued diagonal stripe (similarity $\approx 1.0$) against orthogonal off-diagonal values confirms high identity integrity.
    \item \textbf{PGD Step 149 (Attacked State):} The diagonal feature is completely eliminated, vanishing into a uniform background of yellow-green noise. This demonstrates successful identity collapse, as all tracks are now nearly equally similar to one another.
\end{itemize}

\subsubsection{Memory Embedding Statistics (Feature Stability)}
\label{sec:mem_statistics}
This analysis (Fig.\ref{fig:memory_state}.b plots) tracks the magnitude (L2 Norm) and stability (Std Dev) of the embeddings.
\begin{itemize}
    \item \textbf{PGD Step 0 (Clean):} The L2 Norm (blue bars) is high and consistent, indicating robust, high-energy features across all stable tracks.
    \item \textbf{PGD Step 149 (Attacked State):} The L2 Norm remains relatively high but exhibits massive variability across the memory index. This visual chaos confirms the introduction of significant \textit{Feature Distortion} and instability, which destroys the reliable structure needed for smooth temporal propagation.
\end{itemize}

\subsubsection{Current-Memory Cross-Similarity}
\label{sec:mem_cross_similarity}
This analysis (Fig.\ref{fig:memory_state}.c plots) is the ultimate measure of tracking failure, showing how current observations match against the historical memory bank.
\begin{itemize}
    \item \textbf{PGD Step 0 (Baseline):} The presence of isolated, bright horizontal stripes indicates high confidence matching. Current frame queries find unambiguous, strong similarities against the relevant historical tracks.
    \item \textbf{PGD Step 149 (Attacked State):} The matrix is saturated with a diffuse, chaotic pattern. Distinct, high-confidence matches are eliminated. This demonstrates successful \textit{Match Ambiguity}, as current queries now find low-to-medium similarity with a large number of corrupted memory tracks, forcing the tracker's assignment algorithm to fail (leading to ID switches and track terminations).
\end{itemize}

\subsection{  Analysis of Black-box Transferability}
\label{sec:sm_transferability}

To evaluate the practical threat boundaries of FADE, we conduct a transferability analysis in a black-box setting. Following established protocols in the field~\cite{jia2020fooling, zhou2023f}, we employ a three-frame attack window to standardize the adversarial budget. This setup allows us to move beyond the minimalist efficiency baseline of the single-frame white-box attacks presented in the main paper and investigate the \textit{architectural persistence} of vulnerabilities across unseen targets.

\begin{table}[h]
\centering
\caption{\small Black-box Transferability Check (HOTA) on MOT17.}
\vspace{-0.35cm}
\label{tab:transfer}
\scalebox{.51}{
\begin{tabular}{lc|c|c|c|c|c} 
\toprule
\multirow{2}{*}{\textbf{Surrogate}} & \textbf{Target $\rightarrow$} & \multirow{2}{*}{\textbf{MOTR}} & \multirow{2}{*}{\textbf{MOTRv2}} & \multirow{2}{*}{\textbf{MeMOTR}} & \multirow{2}{*}{\textbf{Samba}} & \multirow{2}{*}{\textbf{CO-MOT}} \\ 
& \textbf{Clean HOTA} & & & & & \\
\hline
\multicolumn{2}{r|}{\textit{None (Baseline)}} & 58.63 & 59.96 & 67.35 & 62.91 & 58.16 \\ 
\hline
\textbf{MOTR} & TMC \textbar{} TQF & \cellcolor{gray!20}42.74 \textbar{} 42.72 & 51.89 \textbar{} 52.04 & 60.31 \textbar{} 60.31 & 56.12 \textbar{} 55.53 & 48.97 \textbar{} 49.51 \\
\textbf{MOTRv2} & TMC \textbar{} TQF & \textcolor{red}{42.76} \textbar{} \textcolor{red}{42.79} & \cellcolor{gray!20}37.08 \textbar{} 27.98 & 60.65 \textbar{} 60.11 & 55.94 \textbar{} 56.56 & \textcolor{red}{48.09} \textbar{} \textcolor{red}{44.78} \\
\textbf{MeMOTR} & TMC \textbar{} TQF & 47.35 \textbar{} 46.26 & 54.11 \textbar{} 53.76 & \cellcolor{gray!20}46.01 \textbar{} 29.10 & \textcolor{red}{52.10} \textbar{} \textcolor{red}{52.43} & 52.16 \textbar{} 52.44 \\
\textbf{Samba} & TMC \textbar{} TQF & 47.60 \textbar{} 47.72 & 54.24 \textbar{} 53.88 & \textcolor{red}{55.20} \textbar{} \textcolor{red}{56.45} & \cellcolor{gray!20}43.22 \textbar{} 36.30 & 52.68 \textbar{} 52.13 \\
\textbf{CO-MOT} & TMC \textbar{} TQF & 42.76 \textbar{} 42.74 & \textcolor{red}{51.84} \textbar{} \textcolor{red}{49.62} & 60.53 \textbar{} 60.32 & 55.90 \textbar{} 56.29 & \cellcolor{gray!20}32.48 \textbar{} 27.03 \\
\bottomrule
\end{tabular}
}
\end{table}

As shown in Table~\ref{tab:transfer}, we observe significant transferability when the surrogate and target trackers share a common \textit{architectural lineage}. Our results reveal three primary transferability clusters:
\begin{itemize}
    \item \textit{Query-Update Lineage:} Strong transferability is observed between MOTR and MOTRv2, which share similar recurrent query-update logic.
    \item \textit{Detector Priors:} MOTRv2 and CO-MOT exhibit high mutual transferability due to their reliance on similar external detection priors.
    \item \textit{Memory Modules:} For specialized memory-augmented architectures like MeMOTR and Samba, using a surrogate with a comparable temporal memory module maximizes the HOTA degradation.
\end{itemize}

\noindent Notably, MOTRv2 emerges as the most effective general-purpose surrogate for black-box deployment, while memory-specific surrogates are required to effectively disrupt the most complex TBP architectures. These findings proves that FADE exploits fundamental structural dependencies inherent to the TBP paradigm rather than model-specific overfitting.

\subsection{  FADE under Common Defenses}
\label{sec:defense}
To evaluate the robustness of the FADE attack strategies against standard input transformation defenses, we employ three input transformation based defense techniques: Color Jittering (CJ), Spatial Smoothing (SS), and Gaussian Noise (GN). These methods aim to disrupt the specific adversarial perturbations generated by the attack before the input reaches the MOT tracker.

\noindent \textbf{Defense 1: Color Jittering (CJ)}
Adversarial perturbations often rely on precise pixel values to mislead the model's gradient. Color Jittering disrupts these dependencies by randomly altering the photometric properties of the input frames. We apply random transformations to the brightness, contrast, saturation, and hue of the adversarial images $x_{adv}$. The transformation can be formulated as a stochastic function $\mathcal{T}_{CJ}(\cdot)$:
$$
x_{def} = \mathcal{T}_{CJ}(x_{adv}; \lambda_{b}, \lambda_{c}, \lambda_{s}, \lambda_{h})
$$
\noindent where $\lambda$ represents the jitter factors. In our experiments, we randomly sample brightness, contrast, and saturation factors from $[1-\delta, 1+\delta]$ and hue from $[-\delta, \delta]$, setting $\delta = 0.2$. This forces the tracker to rely on structural features rather than specific pixel intensities that may have been compromised by the attack.

\begin{table}[t]
\centering
\scriptsize
\setlength{\tabcolsep}{3.5pt} 
\renewcommand{\arraystretch}{1.1} 

\caption{FADE-TMC Performances under Defenses on MOT17.}
\label{tab:defense_results_tmc}
\vspace{-0.3cm}
\scalebox{0.83}{
\begin{tabular}{l l l *{7}{c}} 
\toprule
\rowcolor{HeaderGray}
\multicolumn{10}{c}{\textbf{MOT17 Dataset – Average scene density $\sim$21 objects/frame.}} \\
\midrule

\textbf{Tracker} & \textbf{Attack\_Vector} & \textbf{Defense} 
& \textbf{HOTA} & \textbf{DetA} & \textbf{AssA} 
& \textbf{IDF1} & \textbf{IDR} & \textbf{IDP} & \textbf{IDSW} \\
\midrule 

\multirow{5}{*}{\textbf{MOTR}}
 & {\cellcolor{CleanGreen}Clean} & {\cellcolor{CleanGreen}None} 
   & {\cellcolor{CleanGreen}58.63} & {\cellcolor{CleanGreen}49.90} & {\cellcolor{CleanGreen}70.38} 
   & {\cellcolor{CleanGreen}69.35} & {\cellcolor{CleanGreen}68.48} & {\cellcolor{CleanGreen}71.40} & {\cellcolor{CleanGreen}7.23} \\

 & $\text{FADE}_{\text{TMC}}$ & \textbf{\textcolor{DefenseOrange}{CJ}} 
   & \textbf{\textcolor{DefenseOrange}{46.00}} & \textbf{\textcolor{DefenseOrange}{47.45}} 
   & \textbf{\textcolor{DefenseOrange}{45.88}} & \textbf{\textcolor{DefenseOrange}{51.67}} 
   & \textbf{\textcolor{DefenseOrange}{46.68}} & \textbf{\textcolor{DefenseOrange}{58.57}} 
   & \textbf{\textcolor{DefenseOrange}{9.67}} \\

 & $\text{FADE}_{\text{TMC}}$ & \textbf{\textcolor{DefenseOrange}{GN}} 
   & \textbf{\textcolor{DefenseOrange}{46.38}} & \textbf{\textcolor{DefenseOrange}{47.90}} 
   & \textbf{\textcolor{DefenseOrange}{46.22}} & \textbf{\textcolor{DefenseOrange}{51.87}} 
   & \textbf{\textcolor{DefenseOrange}{46.63}} & \textbf{\textcolor{DefenseOrange}{59.05}} 
   & \textbf{\textcolor{DefenseOrange}{9.51}} \\

 & $\text{FADE}_{\text{TMC}}$ & \textbf{\textcolor{DefenseOrange}{SS}} 
   & \textbf{\textcolor{DefenseOrange}{46.02}} & \textbf{\textcolor{DefenseOrange}{47.88}} 
   & \textbf{\textcolor{DefenseOrange}{45.54}} & \textbf{\textcolor{DefenseOrange}{51.48}} 
   & \textbf{\textcolor{DefenseOrange}{46.13}} & \textbf{\textcolor{DefenseOrange}{58.82}} 
   & \textbf{\textcolor{DefenseOrange}{9.56}} \\

 & $\text{FADE}_{\text{TMC}}$ & \textbf{\textcolor{DeepRed}{None}} 
   & \textbf{\textcolor{DeepRed}{45.89}} & \textbf{\textcolor{DeepRed}{51.36}} 
   & \textbf{\textcolor{DeepRed}{42.18}} & \textbf{\textcolor{DeepRed}{50.45}} 
   & \textbf{\textcolor{DeepRed}{46.35}} & \textbf{\textcolor{DeepRed}{55.79}} 
   & \textbf{\textcolor{DeepRed}{8.77}} \\
\hline 

\multirow{5}{*}{\textbf{MOTRv2}}
 & {\cellcolor{CleanGreen}Clean} & {\cellcolor{CleanGreen}None} 
   & {\cellcolor{CleanGreen}59.96} & {\cellcolor{CleanGreen}49.15} & {\cellcolor{CleanGreen}74.71} 
   & {\cellcolor{CleanGreen}71.99} & {\cellcolor{CleanGreen}60.89} & {\cellcolor{CleanGreen}89.68} 
   & {\cellcolor{CleanGreen}1.75} \\

 & $\text{FADE}_{\text{TMC}}$ & \textbf{\textcolor{DefenseOrange}{CJ}} 
   & \textbf{\textcolor{DefenseOrange}{53.35}} & \textbf{\textcolor{DefenseOrange}{42.30}} 
   & \textbf{\textcolor{DefenseOrange}{69.14}} & \textbf{\textcolor{DefenseOrange}{63.96}} 
   & \textbf{\textcolor{DefenseOrange}{52.95}} & \textbf{\textcolor{DefenseOrange}{83.62}} 
   & \textbf{\textcolor{DefenseOrange}{3.25}} \\

 & $\text{FADE}_{\text{TMC}}$ & \textbf{\textcolor{DefenseOrange}{GN}} 
   & \textbf{\textcolor{DefenseOrange}{52.61}} & \textbf{\textcolor{DefenseOrange}{41.30}} 
   & \textbf{\textcolor{DefenseOrange}{68.87}} & \textbf{\textcolor{DefenseOrange}{63.15}} 
   & \textbf{\textcolor{DefenseOrange}{51.72}} & \textbf{\textcolor{DefenseOrange}{83.65}} 
   & \textbf{\textcolor{DefenseOrange}{3.56}} \\

 & $\text{FADE}_{\text{TMC}}$ & \textbf{\textcolor{DefenseOrange}{SS}} 
   & \textbf{\textcolor{DefenseOrange}{52.95}} & \textbf{\textcolor{DefenseOrange}{41.95}} 
   & \textbf{\textcolor{DefenseOrange}{68.82}} & \textbf{\textcolor{DefenseOrange}{63.58}} 
   & \textbf{\textcolor{DefenseOrange}{52.38}} & \textbf{\textcolor{DefenseOrange}{83.97}} 
   & \textbf{\textcolor{DefenseOrange}{3.31}} \\

 & $\text{FADE}_{\text{TMC}}$ & \textbf{\textcolor{DeepRed}{None}} 
   & \textbf{\textcolor{DeepRed}{46.76}} & \textbf{\textcolor{DeepRed}{37.63}} 
   & \textbf{\textcolor{DeepRed}{59.44}} & \textbf{\textcolor{DeepRed}{56.22}} 
   & \textbf{\textcolor{DeepRed}{42.80}} & \textbf{\textcolor{DeepRed}{83.89}} 
   & \textbf{\textcolor{DeepRed}{3.83}} \\
\hline 

\multirow{5}{*}{\textbf{MeMOTR}}
 & {\cellcolor{CleanGreen}Clean} & {\cellcolor{CleanGreen}None} 
   & {\cellcolor{CleanGreen}67.35} & {\cellcolor{CleanGreen}57.87} & {\cellcolor{CleanGreen}79.60} 
   & {\cellcolor{CleanGreen}80.83} & {\cellcolor{CleanGreen}70.78} & {\cellcolor{CleanGreen}94.84} 
   & {\cellcolor{CleanGreen}0.81} \\

 & $\text{FADE}_{\text{TMC}}$ & \textbf{\textcolor{DefenseOrange}{CJ}} 
   & \textbf{\textcolor{DefenseOrange}{57.13}} & \textbf{\textcolor{DefenseOrange}{53.79}} 
   & \textbf{\textcolor{DefenseOrange}{61.78}} & \textbf{\textcolor{DefenseOrange}{66.96}} 
   & \textbf{\textcolor{DefenseOrange}{56.44}} & \textbf{\textcolor{DefenseOrange}{82.98}} 
   & \textbf{\textcolor{DefenseOrange}{2.80}} \\

 & $\text{FADE}_{\text{TMC}}$ & \textbf{\textcolor{DefenseOrange}{GN}} 
   & \textbf{\textcolor{DefenseOrange}{57.03}} & \textbf{\textcolor{DefenseOrange}{53.26}} 
   & \textbf{\textcolor{DefenseOrange}{62.25}} & \textbf{\textcolor{DefenseOrange}{66.78}} 
   & \textbf{\textcolor{DefenseOrange}{56.17}} & \textbf{\textcolor{DefenseOrange}{83.11}} 
   & \textbf{\textcolor{DefenseOrange}{2.86}} \\

 & $\text{FADE}_{\text{TMC}}$ & \textbf{\textcolor{DefenseOrange}{SS}} 
   & \textbf{\textcolor{DefenseOrange}{57.12}} & \textbf{\textcolor{DefenseOrange}{53.61}} 
   & \textbf{\textcolor{DefenseOrange}{62.01}} & \textbf{\textcolor{DefenseOrange}{66.67}} 
   & \textbf{\textcolor{DefenseOrange}{56.17}} & \textbf{\textcolor{DefenseOrange}{82.60}} 
   & \textbf{\textcolor{DefenseOrange}{2.81}} \\

 & $\text{FADE}_{\text{TMC}}$ & \textbf{\textcolor{DeepRed}{None}} 
   & \textbf{\textcolor{DeepRed}{41.56}} & \textbf{\textcolor{DeepRed}{35.74}} 
   & \textbf{\textcolor{DeepRed}{49.18}} & \textbf{\textcolor{DeepRed}{51.60}} 
   & \textbf{\textcolor{DeepRed}{37.61}} & \textbf{\textcolor{DeepRed}{84.07}} 
   & \textbf{\textcolor{DeepRed}{4.63}} \\
\hline

\multirow{5}{*}{\textbf{Samba}}
 & {\cellcolor{CleanGreen}Clean} & {\cellcolor{CleanGreen}None} 
   & {\cellcolor{CleanGreen}62.91} & {\cellcolor{CleanGreen}50.58} & {\cellcolor{CleanGreen}79.37} 
   & {\cellcolor{CleanGreen}73.67} & {\cellcolor{CleanGreen}60.30} & {\cellcolor{CleanGreen}95.93} 
   & {\cellcolor{CleanGreen}1.02} \\

 & $\text{FADE}_{\text{TMC}}$ & \textbf{\textcolor{DefenseOrange}{CJ}} 
   & \textbf{\textcolor{DefenseOrange}{50.55}} & \textbf{\textcolor{DefenseOrange}{46.44}} 
   & \textbf{\textcolor{DefenseOrange}{56.02}} & \textbf{\textcolor{DefenseOrange}{58.11}} 
   & \textbf{\textcolor{DefenseOrange}{46.06}} & \textbf{\textcolor{DefenseOrange}{80.74}} 
   & \textbf{\textcolor{DefenseOrange}{3.19}} \\

 & $\text{FADE}_{\text{TMC}}$ & \textbf{\textcolor{DefenseOrange}{GN}} 
   & \textbf{\textcolor{DefenseOrange}{50.97}} & \textbf{\textcolor{DefenseOrange}{46.95}} 
   & \textbf{\textcolor{DefenseOrange}{56.51}} & \textbf{\textcolor{DefenseOrange}{58.31}} 
   & \textbf{\textcolor{DefenseOrange}{46.44}} & \textbf{\textcolor{DefenseOrange}{80.72}} 
   & \textbf{\textcolor{DefenseOrange}{3.24}} \\

 & $\text{FADE}_{\text{TMC}}$ & \textbf{\textcolor{DefenseOrange}{SS}} 
   & \textbf{\textcolor{DefenseOrange}{50.43}} & \textbf{\textcolor{DefenseOrange}{46.61}} 
   & \textbf{\textcolor{DefenseOrange}{55.66}} & \textbf{\textcolor{DefenseOrange}{57.70}} 
   & \textbf{\textcolor{DefenseOrange}{45.89}} & \textbf{\textcolor{DefenseOrange}{79.90}} 
   & \textbf{\textcolor{DefenseOrange}{3.31}} \\

 & $\text{FADE}_{\text{TMC}}$ & \textbf{\textcolor{DeepRed}{None}} 
   & \textbf{\textcolor{DeepRed}{48.04}} & \textbf{\textcolor{DeepRed}{45.19}} 
   & \textbf{\textcolor{DeepRed}{51.93}} & \textbf{\textcolor{DeepRed}{56.01}} 
   & \textbf{\textcolor{DeepRed}{44.01}} & \textbf{\textcolor{DeepRed}{78.74}} 
   & \textbf{\textcolor{DeepRed}{3.63}} \\
\hline

\multirow{5}{*}{\textbf{CO-MOT}}
 & {\cellcolor{CleanGreen}Clean} & {\cellcolor{CleanGreen}None} 
   & {\cellcolor{CleanGreen}58.16} & {\cellcolor{CleanGreen}46.22} & {\cellcolor{CleanGreen}74.87} 
   & {\cellcolor{CleanGreen}69.87} & {\cellcolor{CleanGreen}57.21} & {\cellcolor{CleanGreen}91.97} 
   & {\cellcolor{CleanGreen}1.83} \\

 & $\text{FADE}_{\text{TMC}}$ & \textbf{\textcolor{DefenseOrange}{CJ}} 
   & \textbf{\textcolor{DefenseOrange}{52.61}} & \textbf{\textcolor{DefenseOrange}{40.54}} 
   & \textbf{\textcolor{DefenseOrange}{70.57}} & \textbf{\textcolor{DefenseOrange}{62.51}} 
   & \textbf{\textcolor{DefenseOrange}{49.77}} & \textbf{\textcolor{DefenseOrange}{88.24}} 
   & \textbf{\textcolor{DefenseOrange}{2.47}} \\

 & $\text{FADE}_{\text{TMC}}$ & \textbf{\textcolor{DefenseOrange}{GN}} 
   & \textbf{\textcolor{DefenseOrange}{53.54}} & \textbf{\textcolor{DefenseOrange}{41.69}} 
   & \textbf{\textcolor{DefenseOrange}{71.23}} & \textbf{\textcolor{DefenseOrange}{63.81}} 
   & \textbf{\textcolor{DefenseOrange}{51.07}} & \textbf{\textcolor{DefenseOrange}{89.64}} 
   & \textbf{\textcolor{DefenseOrange}{1.91}} \\

 & $\text{FADE}_{\text{TMC}}$ & \textbf{\textcolor{DefenseOrange}{SS}} 
   & \textbf{\textcolor{DefenseOrange}{52.86}} & \textbf{\textcolor{DefenseOrange}{40.75}} 
   & \textbf{\textcolor{DefenseOrange}{70.70}} & \textbf{\textcolor{DefenseOrange}{63.09}} 
   & \textbf{\textcolor{DefenseOrange}{50.82}} & \textbf{\textcolor{DefenseOrange}{87.54}} 
   & \textbf{\textcolor{DefenseOrange}{2.68}} \\

 & $\text{FADE}_{\text{TMC}}$ & \textbf{\textcolor{DeepRed}{None}} 
   & \textbf{\textcolor{DeepRed}{41.73}} & \textbf{\textcolor{DeepRed}{31.82}} 
   & \textbf{\textcolor{DeepRed}{55.89}} & \textbf{\textcolor{DeepRed}{50.34}} 
   & \textbf{\textcolor{DeepRed}{39.23}} & \textbf{\textcolor{DeepRed}{72.34}} 
   & \textbf{\textcolor{DeepRed}{10.94}} \\

\bottomrule
\end{tabular}
}
\end{table}

\begin{table}[t]
\centering
\scriptsize
\setlength{\tabcolsep}{3.5pt} 
\renewcommand{\arraystretch}{1.1} 

\caption{FADE-TQF Performances under Defenses on MOT17.}
\label{tab:defense_results_tqf}
\vspace{-0.3cm}
\scalebox{0.83}{
\begin{tabular}{l l l *{7}{c}} 
\toprule
\rowcolor{HeaderGray}
\multicolumn{10}{c}{\textbf{MOT17 Dataset – Average scene density $\sim$21 objects/frame.}} \\[-0.2em]
\midrule

\textbf{Tracker} & \textbf{Attack\_Vector} & \textbf{Defense} 
& \textbf{HOTA} & \textbf{DetA} & \textbf{AssA} 
& \textbf{IDF1} & \textbf{IDR} & \textbf{IDP} & \textbf{IDSW} \\
\midrule

\multirow{5}{*}{\textbf{MOTR}}
 & {\cellcolor{CleanGreen}Clean} & {\cellcolor{CleanGreen}None} 
   & {\cellcolor{CleanGreen}58.63} & {\cellcolor{CleanGreen}49.90} & {\cellcolor{CleanGreen}70.38} 
   & {\cellcolor{CleanGreen}69.35} & {\cellcolor{CleanGreen}68.48} & {\cellcolor{CleanGreen}71.40} & {\cellcolor{CleanGreen}7.23} \\

 & $\text{FADE}_{\text{TQF}}$ & \textbf{\textcolor{DefenseOrange}{CJ}} 
   & \textbf{\textcolor{DefenseOrange}{47.00}} & \textbf{\textcolor{DefenseOrange}{47.47}} 
   & \textbf{\textcolor{DefenseOrange}{47.87}} & \textbf{\textcolor{DefenseOrange}{53.09}} 
   & \textbf{\textcolor{DefenseOrange}{47.88}} & \textbf{\textcolor{DefenseOrange}{60.24}} 
   & \textbf{\textcolor{DefenseOrange}{9.17}} \\

 & $\text{FADE}_{\text{TQF}}$ & \textbf{\textcolor{DefenseOrange}{GN}} 
   & \textbf{\textcolor{DefenseOrange}{47.03}} & \textbf{\textcolor{DefenseOrange}{47.92}} 
   & \textbf{\textcolor{DefenseOrange}{47.51}} & \textbf{\textcolor{DefenseOrange}{53.19}} 
   & \textbf{\textcolor{DefenseOrange}{48.23}} & \textbf{\textcolor{DefenseOrange}{60.04}} 
   & \textbf{\textcolor{DefenseOrange}{9.69}} \\

 & $\text{FADE}_{\text{TQF}}$ & \textbf{\textcolor{DefenseOrange}{SS}} 
   & \textbf{\textcolor{DefenseOrange}{47.18}} & \textbf{\textcolor{DefenseOrange}{47.26}} 
   & \textbf{\textcolor{DefenseOrange}{48.43}} & \textbf{\textcolor{DefenseOrange}{53.21}} 
   & \textbf{\textcolor{DefenseOrange}{48.15}} & \textbf{\textcolor{DefenseOrange}{60.08}} 
   & \textbf{\textcolor{DefenseOrange}{9.47}} \\

 & $\text{FADE}_{\text{TQF}}$ & \textbf{\textcolor{DeepRed}{None}} 
   & \textbf{\textcolor{DeepRed}{45.90}} & \textbf{\textcolor{DeepRed}{51.40}} 
   & \textbf{\textcolor{DeepRed}{42.17}} & \textbf{\textcolor{DeepRed}{50.51}} 
   & \textbf{\textcolor{DeepRed}{46.40}} & \textbf{\textcolor{DeepRed}{55.86}} 
   & \textbf{\textcolor{DeepRed}{8.73}} \\
\hline

\multirow{5}{*}{\textbf{MOTRv2}}
 & {\cellcolor{CleanGreen}Clean} & {\cellcolor{CleanGreen}None} 
   & {\cellcolor{CleanGreen}59.96} & {\cellcolor{CleanGreen}49.15} & {\cellcolor{CleanGreen}74.71} 
   & {\cellcolor{CleanGreen}71.99} & {\cellcolor{CleanGreen}60.89} & {\cellcolor{CleanGreen}89.68} 
   & {\cellcolor{CleanGreen}1.75} \\

 & $\text{FADE}_{\text{TQF}}$ & \textbf{\textcolor{DefenseOrange}{CJ}} 
   & \textbf{\textcolor{DefenseOrange}{53.19}} & \textbf{\textcolor{DefenseOrange}{42.42}} 
   & \textbf{\textcolor{DefenseOrange}{68.55}} & \textbf{\textcolor{DefenseOrange}{63.96}} 
   & \textbf{\textcolor{DefenseOrange}{52.70}} & \textbf{\textcolor{DefenseOrange}{84.00}} 
   & \textbf{\textcolor{DefenseOrange}{3.02}} \\

 & $\text{FADE}_{\text{TQF}}$ & \textbf{\textcolor{DefenseOrange}{GN}} 
   & \textbf{\textcolor{DefenseOrange}{53.18}} & \textbf{\textcolor{DefenseOrange}{42.09}} 
   & \textbf{\textcolor{DefenseOrange}{69.06}} & \textbf{\textcolor{DefenseOrange}{63.68}} 
   & \textbf{\textcolor{DefenseOrange}{52.51}} & \textbf{\textcolor{DefenseOrange}{84.13}} 
   & \textbf{\textcolor{DefenseOrange}{3.31}} \\

 & $\text{FADE}_{\text{TQF}}$ & \textbf{\textcolor{DefenseOrange}{SS}} 
   & \textbf{\textcolor{DefenseOrange}{52.77}} & \textbf{\textcolor{DefenseOrange}{42.14}} 
   & \textbf{\textcolor{DefenseOrange}{67.93}} & \textbf{\textcolor{DefenseOrange}{63.44}} 
   & \textbf{\textcolor{DefenseOrange}{52.37}} & \textbf{\textcolor{DefenseOrange}{83.19}} 
   & \textbf{\textcolor{DefenseOrange}{3.89}} \\

 & $\text{FADE}_{\text{TQF}}$ & \textbf{\textcolor{DeepRed}{None}} 
   & \textbf{\textcolor{DeepRed}{39.29}} & \textbf{\textcolor{DeepRed}{31.68}} 
   & \textbf{\textcolor{DeepRed}{49.96}} & \textbf{\textcolor{DeepRed}{49.02}} 
   & \textbf{\textcolor{DeepRed}{35.87}} & \textbf{\textcolor{DeepRed}{78.77}} 
   & \textbf{\textcolor{DeepRed}{5.65}} \\
\hline

\multirow{5}{*}{\textbf{MeMOTR}}
 & {\cellcolor{CleanGreen}Clean} & {\cellcolor{CleanGreen}None} 
   & {\cellcolor{CleanGreen}67.35} & {\cellcolor{CleanGreen}57.87} & {\cellcolor{CleanGreen}79.60} 
   & {\cellcolor{CleanGreen}80.83} & {\cellcolor{CleanGreen}70.78} & {\cellcolor{CleanGreen}94.84} 
   & {\cellcolor{CleanGreen}0.81} \\

 & $\text{FADE}_{\text{TQF}}$ & \textbf{\textcolor{DefenseOrange}{CJ}} 
   & \textbf{\textcolor{DefenseOrange}{62.97}} & \textbf{\textcolor{DefenseOrange}{54.47}} 
   & \textbf{\textcolor{DefenseOrange}{73.96}} & \textbf{\textcolor{DefenseOrange}{75.44}} 
   & \textbf{\textcolor{DefenseOrange}{63.93}} & \textbf{\textcolor{DefenseOrange}{92.69}} 
   & \textbf{\textcolor{DefenseOrange}{1.43}} \\

 & $\text{FADE}_{\text{TQF}}$ & \textbf{\textcolor{DefenseOrange}{GN}} 
   & \textbf{\textcolor{DefenseOrange}{63.39}} & \textbf{\textcolor{DefenseOrange}{54.74}} 
   & \textbf{\textcolor{DefenseOrange}{74.53}} & \textbf{\textcolor{DefenseOrange}{76.10}} 
   & \textbf{\textcolor{DefenseOrange}{64.67}} & \textbf{\textcolor{DefenseOrange}{93.08}} 
   & \textbf{\textcolor{DefenseOrange}{1.35}} \\

 & $\text{FADE}_{\text{TQF}}$ & \textbf{\textcolor{DefenseOrange}{SS}} 
   & \textbf{\textcolor{DefenseOrange}{62.91}} & \textbf{\textcolor{DefenseOrange}{54.43}} 
   & \textbf{\textcolor{DefenseOrange}{73.88}} & \textbf{\textcolor{DefenseOrange}{75.25}} 
   & \textbf{\textcolor{DefenseOrange}{63.72}} & \textbf{\textcolor{DefenseOrange}{92.61}} 
   & \textbf{\textcolor{DefenseOrange}{1.43}} \\

 & $\text{FADE}_{\text{TQF}}$ & \textbf{\textcolor{DeepRed}{None}} 
   & \textbf{\textcolor{DeepRed}{41.41}} & \textbf{\textcolor{DeepRed}{34.83}} 
   & \textbf{\textcolor{DeepRed}{50.03}} & \textbf{\textcolor{DeepRed}{51.41}} 
   & \textbf{\textcolor{DeepRed}{37.07}} & \textbf{\textcolor{DeepRed}{85.38}} 
   & \textbf{\textcolor{DeepRed}{4.31}} \\
\hline

\multirow{5}{*}{\textbf{Samba}}
 & {\cellcolor{CleanGreen}Clean} & {\cellcolor{CleanGreen}None} 
   & {\cellcolor{CleanGreen}62.91} & {\cellcolor{CleanGreen}50.58} & {\cellcolor{CleanGreen}79.37} 
   & {\cellcolor{CleanGreen}73.67} & {\cellcolor{CleanGreen}60.30} & {\cellcolor{CleanGreen}95.93} 
   & {\cellcolor{CleanGreen}1.02} \\

 & $\text{FADE}_{\text{TQF}}$ & \textbf{\textcolor{DefenseOrange}{CJ}} 
   & \textbf{\textcolor{DefenseOrange}{55.31}} & \textbf{\textcolor{DefenseOrange}{45.90}} 
   & \textbf{\textcolor{DefenseOrange}{68.17}} & \textbf{\textcolor{DefenseOrange}{64.25}} 
   & \textbf{\textcolor{DefenseOrange}{50.79}} & \textbf{\textcolor{DefenseOrange}{90.41}} 
   & \textbf{\textcolor{DefenseOrange}{1.93}} \\

 & $\text{FADE}_{\text{TQF}}$ & \textbf{\textcolor{DefenseOrange}{GN}} 
   & \textbf{\textcolor{DefenseOrange}{56.16}} & \textbf{\textcolor{DefenseOrange}{46.60}} 
   & \textbf{\textcolor{DefenseOrange}{69.16}} & \textbf{\textcolor{DefenseOrange}{65.46}} 
   & \textbf{\textcolor{DefenseOrange}{52.14}} & \textbf{\textcolor{DefenseOrange}{90.87}} 
   & \textbf{\textcolor{DefenseOrange}{1.78}} \\

 & $\text{FADE}_{\text{TQF}}$ & \textbf{\textcolor{DefenseOrange}{SS}} 
   & \textbf{\textcolor{DefenseOrange}{54.95}} & \textbf{\textcolor{DefenseOrange}{45.78}} 
   & \textbf{\textcolor{DefenseOrange}{67.60}} & \textbf{\textcolor{DefenseOrange}{63.99}} 
   & \textbf{\textcolor{DefenseOrange}{50.71}} & \textbf{\textcolor{DefenseOrange}{90.09}} 
   & \textbf{\textcolor{DefenseOrange}{1.93}} \\

 & $\text{FADE}_{\text{TQF}}$ & \textbf{\textcolor{DeepRed}{None}} 
   & \textbf{\textcolor{DeepRed}{45.53}} & \textbf{\textcolor{DeepRed}{32.71}} 
   & \textbf{\textcolor{DeepRed}{64.37}} & \textbf{\textcolor{DeepRed}{52.71}} 
   & \textbf{\textcolor{DeepRed}{37.62}} & \textbf{\textcolor{DeepRed}{91.33}} 
   & \textbf{\textcolor{DeepRed}{2.24}} \\
\hline

\multirow{5}{*}{\textbf{CO-MOT}}
 & {\cellcolor{CleanGreen}Clean} & {\cellcolor{CleanGreen}None} 
   & {\cellcolor{CleanGreen}58.16} & {\cellcolor{CleanGreen}46.22} & {\cellcolor{CleanGreen}74.87} 
   & {\cellcolor{CleanGreen}69.87} & {\cellcolor{CleanGreen}57.21} & {\cellcolor{CleanGreen}91.97} 
   & {\cellcolor{CleanGreen}1.83} \\

 & $\text{FADE}_{\text{TQF}}$ & \textbf{\textcolor{DefenseOrange}{CJ}} 
   & \textbf{\textcolor{DefenseOrange}{52.65}} & \textbf{\textcolor{DefenseOrange}{40.54}} 
   & \textbf{\textcolor{DefenseOrange}{70.50}} & \textbf{\textcolor{DefenseOrange}{62.88}} 
   & \textbf{\textcolor{DefenseOrange}{50.72}} & \textbf{\textcolor{DefenseOrange}{87.19}} 
   & \textbf{\textcolor{DefenseOrange}{2.99}} \\

 & $\text{FADE}_{\text{TQF}}$ & \textbf{\textcolor{DefenseOrange}{GN}} 
   & \textbf{\textcolor{DefenseOrange}{53.42}} & \textbf{\textcolor{DefenseOrange}{41.29}} 
   & \textbf{\textcolor{DefenseOrange}{71.05}} & \textbf{\textcolor{DefenseOrange}{63.81}} 
   & \textbf{\textcolor{DefenseOrange}{50.82}} & \textbf{\textcolor{DefenseOrange}{89.67}} 
   & \textbf{\textcolor{DefenseOrange}{1.91}} \\

 & $\text{FADE}_{\text{TQF}}$ & \textbf{\textcolor{DefenseOrange}{SS}} 
   & \textbf{\textcolor{DefenseOrange}{52.44}} & \textbf{\textcolor{DefenseOrange}{40.13}} 
   & \textbf{\textcolor{DefenseOrange}{70.71}} & \textbf{\textcolor{DefenseOrange}{62.15}} 
   & \textbf{\textcolor{DefenseOrange}{49.84}} & \textbf{\textcolor{DefenseOrange}{87.51}} 
   & \textbf{\textcolor{DefenseOrange}{2.50}} \\

 & $\text{FADE}_{\text{TQF}}$ & \textbf{\textcolor{DeepRed}{None}} 
   & \textbf{\textcolor{DeepRed}{37.26}} & \textbf{\textcolor{DeepRed}{27.43}} 
   & \textbf{\textcolor{DeepRed}{51.93}} & \textbf{\textcolor{DeepRed}{44.84}} 
   & \textbf{\textcolor{DeepRed}{32.88}} & \textbf{\textcolor{DeepRed}{74.66}} 
   & \textbf{\textcolor{DeepRed}{9.50}} \\
\bottomrule
\end{tabular}
}
\end{table}

\noindent \textbf{Defense 2: Spatial Smoothing (SS)}
Since adversarial noise typically manifests as high-frequency fluctuations imperceptible to the human eye, Spatial Smoothing acts as a low-pass filter to mitigate these artifacts. We utilize a Gaussian blur kernel to convolve the input image, effectively smoothing out the sharp gradients introduced by the attack optimization process. The smoothed image $x_{SS}$ is obtained via:
$$
x_{SS} = x_{adv} * G_{k, \sigma}
$$
\noindent where $G$ is a Gaussian kernel of size $k \times k$ with standard deviation $\sigma$. We evaluate robustness using a kernel size of $k=3$ and $\sigma=0.5$. This defense tests the attack's ability to survive the removal of its high-frequency components.

\noindent \textbf{Defense 3: Gaussian Noise (GN)}
The Gaussian Noise defense introduces random stochasticity to the input, aiming to disrupt the precise alignment of the adversarial perturbation. While adding noise seems counter-intuitive, it is a foundational concept in \textit{Randomized Smoothing}, where the classifier is forced to be robust within a local neighborhood of the input. We inject additive white Gaussian noise $\eta$ into the adversarial input:
$$
x_{GN} = x_{adv} + \eta, \quad \text{where } \eta \sim \mathcal{N}(0, \sigma^2_{gn})
$$
\noindent We set the noise standard deviation $\sigma_{gn} = 0.1$. This defense effectively drowns out small-magnitude adversarial perturbations, testing whether the attack features are distinct enough to persist through random interference.

\subsubsection{Analysis of Robustness and Defense Efficacy}
To assess the robustness of the FADE framework, we evaluate three standard input transformation defenses: Color Jitter (CJ), Gaussian Noise (GN), and Spatial Smoothing (SS). Tables \ref{tab:defense_results_tmc} and \ref{tab:defense_results_tqf} detail the performance of five TBP-based trackers under these defenses against FADE-TMC and FADE-TQF, respectively. We consider the same attack implementation details for one attacked frame only.

\noindent \textbf{(\textit{i}) The Recovery Gap.}
A consistent pattern across all experiments is the existence of a \textit{Recovery Gap}: The performance difference between the Defended state and the Clean baseline.
On the MOTR tracker, defenses provide negligible benefit against the TMC attack. As shown in Table \ref{tab:defense_results_tmc}, the undefended FADE-TMC attack reduces HOTA to 45.89. Applying Gaussian Noise (GN) only raises this to 46.38, and Spatial Smoothing (SS) to 46.02. This indicates that for pure query-propagation architectures, the adversarial perturbation is deeply embedded in the feature representation and cannot be removed by simple pre-processing.
In contrast, the MeMOTR tracker demonstrates some recoverability, particularly against TQF. In Table \ref{tab:defense_results_tqf}, the TQF attack drops MeMOTR's HOTA to 41.41. However, defenses are able to restore this to $\sim$63.0 (e.g., GN yields 63.39), bringing it quite closer to the clean baseline of 67.35. This suggests that MeMOTR's long-term memory bank provides a \textit{stabilizing anchor} that, when aided by input sanitization, can filter out transient adversarial noise.

\noindent \textbf{(\textit{ii}) Defense Invariance.}
An analysis of the specific defense modalities reveals a high degree of performance invariance. Across almost all trackers and attack vectors, the variation in HOTA scores between CJ, GN, and SS is statistically insignificant (typically $< 0.5$ points).
For example, on MOTRv2 under FADE-TQF (Table \ref{tab:defense_results_tqf}), the HOTA scores for CJ, GN, and SS are 53.19, 53.18, and 52.77, respectively.
This invariance implies that the FADE attack does not rely on a single fragile feature modality (such as high-frequency noise alone or specific color triggers). Instead, the adversarial optimization converges to a robust perturbation that permeates multiple feature domains, making it resilient to any single type of input transformation.

\noindent \textbf{(\textit{iii}) Vulnerability of State-Space Models.}
The results highlight a fragility in the Samba tracker regarding defense efficacy. While MeMOTR recovers $\sim$83\% of its lost performance under defenses against TQF, Samba shows significantly lower elasticity.
Under FADE-TMC (Table \ref{tab:defense_results_tmc}), Samba drops to 48.04 HOTA. The best-performing defense (GN) only restores this to 50.97, leaving a massive gap from the clean baseline of 62.91.
This suggests that for State-Space Models, the adversarial corruption interferes with the sequential state dynamics in a way that frame-level pre-processing cannot correct. Once the state transition is effectively jammed by the attack, cleaning the input pixels is insufficient to reset the tracking logic.

\noindent \textbf{(\textit{iv}) Divergent Defense Efficacy: TMC vs. TQF.}
A cross-comparison of Tables \ref{tab:defense_results_tmc} and \ref{tab:defense_results_tqf} reveals a fundamental divergence in how the two attacks respond to input sanitization. 
The TQF attack operates by generating fake queries, spurious high-confidence detections, designed to exhaust the tracker's budget. The results show that these fake queries are relatively fragile.
On the MeMOTR tracker, defenses recover the HOTA score from 41.41 (undefended) to $\sim$63.0. This represents a recovery of nearly 83\% of the performance drop. 
Spurious objects are often triggered by high-frequency adversarial noise patterns that mimic object textures. Standard defenses like Spatial Smoothing and Gaussian Noise can disrupt these high-frequency triggers, causing the detector to suppress the fake queries. Once the fakes are removed, the query budget is relieved, and the tracker recovers.
In contrast, the TMC attack proves exceptionally resistant to defenses.
On the same MeMOTR tracker, defenses against TMC only recover the HOTA score from 41.56 to $\sim$57.03. Recovery saturates at just 57\%, leaving a permanent performance gap. On MOTR, the recovery is negligible ($<1\%$). 
TMC does not rely on creating new, fragile objects. Instead, it subtly shifts the embeddings of \textit{existing, valid} objects within the feature manifold to break their temporal links. This perturbation is semantically aligned with the track identity features (often lower-frequency patterns) rather than being mere surface noise. Consequently, simple input filters (CJ, GN, SS) cannot remove the adversarial signal without also degrading the legitimate object features, rendering the defense ineffective.
This divergence confirms that while TQF acts as a \textit{surface-level} capacity attack (mitigatable via pre-processing), TMC acts as a \textit{deep-feature} identity attack that requires architectural hardening to defeat.

\subsection{  Practicality, Realism, and Future Work}
\label{sec:sm_discussion}

The practical viability of FADE is supported by its computational efficiency and its grounding in physically calibrated sensor models. On modern hardware, the PGD optimization process requires approximately $3$~s on an NVIDIA RTX 3090 and reaches $\sim 1$~s on an A100 per frame. While per-frame optimization is useful for characterizing vulnerability bounds, the high transferability observed in Table~\ref{tab:transfer} suggests that pre-computed perturbations generated on a surrogate model can enable \textit{zero-latency online deployment}. By leveraging these pre-computed adversarial perturbations, an adversary can bypass the need for real-time optimization entirely, facilitating immediate state-poisoning of the target tracker. Regarding realism, our experiments utilize sensor-level simulations (AAI and EAI) that are directly calibrated to real-world camera profiles \cite{zhu2023tpatch, ji2021poltergeist, liu2025magshadow, jiang2023glitchhiker}, as detailed in Section~\ref{sec:aai_eai_details} of the Appendix. While we acknowledge that closed-loop hardware-in-the-loop validation remains a distinct engineering challenge involving variable environmental factors, FADE formulates the foundational algorithmic feasibility and the prerequisite failure physics necessary for such future benchmarks. To further enhance the robustness of these threats, future work can explore \textit{ensemble learning techniques} \cite{yao2024understanding, tang2024ensemble} to generate universal adversarial perturbations. By optimizing across a diverse pool of TBP architectures simultaneously, we posit that it may be possible to overcome current architectural lineage constraints and develop truly cross-paradigm adversarial vectors that generalize to entirely unknown multi-object tracking systems.

\end{document}